\documentclass{article}
\usepackage{jfrExamplee}
\usepackage{setspace}
\usepackage{algorithm,algcompatible,amsmath}

\newcommand{\change}[1]{{\color{black} #1}}
\algnewcommand\INPUT{\item[\textbf{Input:}]}%
\algnewcommand\OUTPUT{\item[\textbf{Output:}]}%

\title{Zeus: A System Description of the Two-Time Winner of the Collegiate SAE AutoDrive Competition}

\author{Keenan Burnett, Jingxing Qian, Xintong Du, Linqiao Liu, David J. Yoon, \\ Tianchang Shen, Susan Sun, Sepehr Samavi, Michael J. Sorocky, Mollie Bianchi, Kaicheng Zhang, Arkady Arkhangorodsky, Quinlan Sykora, Shichen Lu, Yizhou Huang, Angela P. Schoellig, Timothy D. Barfoot}

\author{
Keenan Burnett\thanks{All authors are affiliated with the University of Toronto. Questions and comments can be sent to keenan.burnett@autodrive.utoronto.ca}
\And
Jingxing Qian
\And
Xintong Du
\And
Linqiao Liu
\And
David J. Yoon
\AND
Tianchang Shen \\
\And
Susan Sun \\
\And
Sepehr Samavi \\
\And
Michael J. Sorocky \\
\AND
Mollie Bianchi \\
\And
Kaicheng Zhang \\
\And
Arkady Arkhangorodsky \\
\And
Quinlan Sykora \\
\AND
Shichen Lu \\
\And
Yizhou Huang \\
\And
Angela P. Schoellig \\
\And
Timothy D. Barfoot \\
}

\usepackage[pdftex]{graphicx}
\usepackage{subfigure}
\usepackage{url}
\usepackage{float}
\usepackage{dblfloatfix}
\usepackage{xcolor}
\usepackage[round,sort,comma,numbers]{natbib}
\usepackage[utf8]{inputenc}
\usepackage[short]{optidef}
\setcitestyle{authoryear}

\begin{document}

\maketitle
\thispagestyle{empty}
\pagestyle{empty}

\begin{abstract}
The SAE AutoDrive Challenge is a three-year collegiate competition to develop a self-driving car by 2020. The second year of the competition was held in June 2019 at MCity, a mock town built for self-driving car testing at the University of Michigan. Teams were required to autonomously navigate a series of intersections while handling pedestrians, traffic lights, and traffic signs. Zeus is aUToronto's winning entry in the AutoDrive Challenge. This article describes the system design and development of Zeus as well as many of the lessons learned along the way. This includes details on the team's organizational structure, sensor suite, software components, and performance at the Year 2 competition. With a team of mostly undergraduates and minimal resources, aUToronto has made progress towards a \change{functioning self-driving vehicle}, in just two years. This article may prove valuable to researchers looking to develop their own self-driving platform.
\end{abstract}

\section{Introduction}

aUToronto is a team of undergraduate and graduate students at the University of Toronto. In just two years, the team has built \textit{Zeus}, a self-driving car \change{that takes a step towards Level 4 autonomy on a closed course}. This work has been centered around competing in the SAE AutoDrive Challenge, a collegiate competition to develop a self-driving car, in only three years. Similar to the DARPA Urban Challenge from 2008, this competition required teams to autonomously sequence several intersections while handling a realistic urban driving environment \citep{bossurbanchallenge}.

The second year of the competition simulated a small town with tasks requiring a high degree of autonomy. Only one hour was given at the start of the competition to perform minor adjustments and prior access was prohibited. Furthermore, only two attempts were allowed for each course.

With this in mind, competing systems needed to be exceptionally reliable. Developing such a system with minimal resources in a constrained time frame required rapid development and simple designs. This article attempts to provide a complete picture of the design and development of Zeus with topics including team organization, sensor configuration, deep learning acceleration, and \change{data collection}. Further, this article will cover the design of each major software component. This article concludes with a detailed analysis of our Year 2 competition performance and lessons learned. As a lot of current self-driving development occurs in secret industrial labs, our hope is that this comprehensive summary of aUToronto's work may prove useful to the wider robotics community.

\begin{figure} [t]
    \includegraphics[width=0.7\columnwidth]{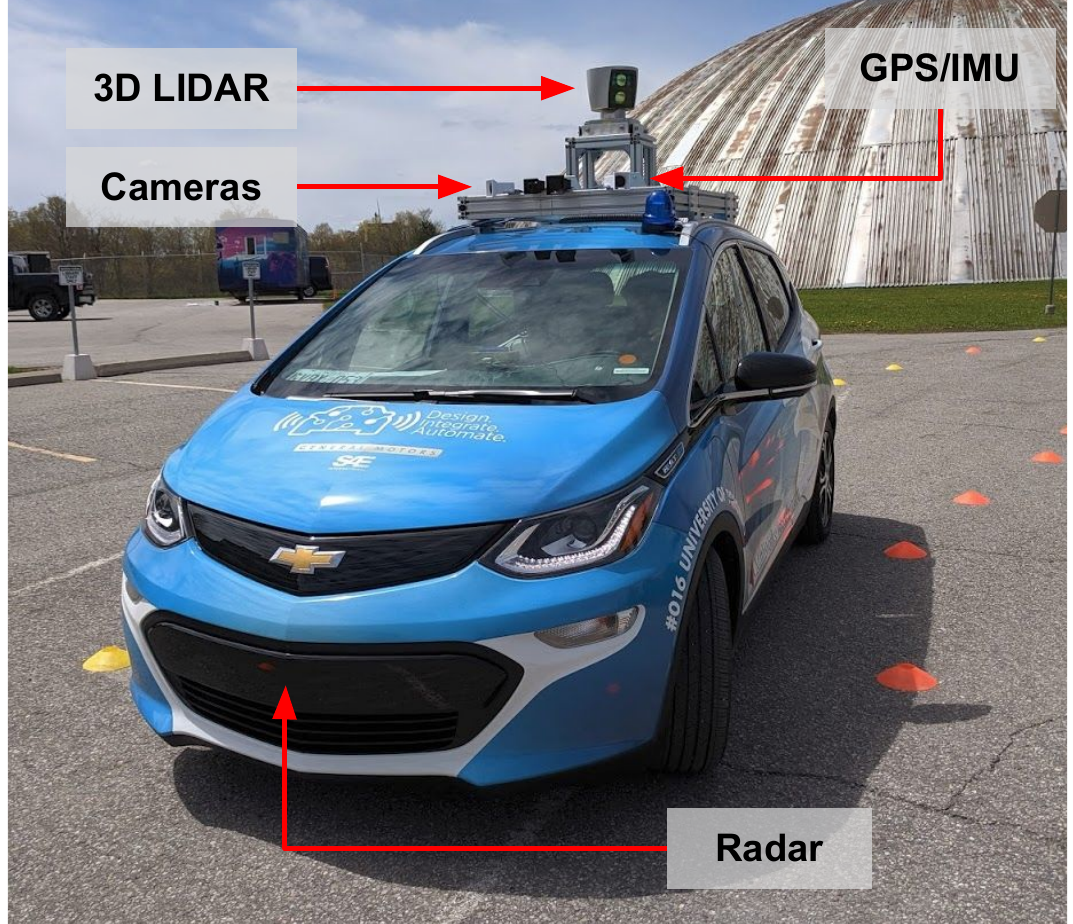}    
    \centering
    \caption{Our self-driving car \textit{Zeus} at the University of Toronto. \change{A video depicting our Year 2 competition performance can be found here: \url{https://youtu.be/2Z6mPKIv0TM}}}
    \label{fig:zeus}
\end{figure}

\section{Background}

\subsection{Related Work}

The DARPA Urban Challenge competitors were the first to demonstrate the feasibility of autonomous driving in an urban setting \citep{bossurbanchallenge}, \citep{stanfordjunior}. This challenge required competitors to sequence several intersections while obeying the rules of the road and interacting with other autonomous vehicles. These systems relied heavily on GPS for localization, and relied on LIDAR for detecting lanes and vehicles. Few teams made significant use of vision. In the decade since, vision has become a key sensing modality. This is in part due to the proliferation of parallel computing and advances in deep learning.

More recently, these competitors provided updates to the Urban Challenge systems. In \citep{Levinson2011}, one update involved the use of \change{high-resolution} maps for online localization with centimeter-level accuracy. In \citep{Wei2013}, the authors describe an updated self-driving platform constructed using close-to-market sensors.

In 2014, the Mercedes/Bertha project succeeded in navigating 100 km autonomously using only radar and vision \citep{Ziegler2014}. The route included a mix of urban and rural driving with traffic lights, pedestrian crossings, and intersections with traffic. In order to navigate the route successfully, they relied on a detailed digital map which encoded the road topology and lane locations. A stereo camera was used as the primary source of 3D \change{object} detections. In 2017, the same group published an update for the 2016 Grand Cooperative Driving Challenge \citep{berthacoop}. They added LIDAR sensors and provided several updates to their perception software. \change{However,} this perception software was not enabled during their competition.

There have also been efforts to open-source self-driving software, notably Autoware \citep{autoware15} and Baidu's Apollo \citep{Apollo}. Both groups aim to provide an open-source repository to enable Level 4 autonomous driving in an urban environment. The Apollo software suite is much more advanced, having undergone a more rigorous development and testing process. The Apollo repository is also used by numerous other groups as a basis for research and development. To date, a system-level summary of Apollo has not yet been published. The most recent update to the Autoware project focuses on implementing their software on embedded platforms \citep{autoware18}.

Other published systems include the V-Charge project \citep{closetomarket2013}, BMW's work on autonomous highway driving \citep{bmw2015}, an autonomous golf cart pilot in Singapore \citep{golfcars2015}, and the PROUD driverless car test \citep{broggi2015proud}. In the V-Charge project, the goal was to develop an autonomous valet parking system using close-to-market sensors. \citet{bmw2015} describe the system that was developed at BMW to perform autonomous highway driving. Their system uses a combination of LIDAR, vision and radar to detect lane markings and other vehicles. They also relied on high-precision maps of the lanes and road boundaries. 

\citet{golfcars2015} focused on demonstrating a low-speed autonomous shuttle to raise public awareness of autonomous vehicles. Since the golf carts operated continuously for long periods of time, reliability was critical. The vehicles were programmed to follow a predefined path and would stop and wait for any blockage to become clear. 

\citet{broggi2015proud} conducted an autonomous driving test on public roads in Parma, Italy. Their system used both vision and LIDAR to detect vehicles and relied on a highly-accurate GPS/IMU for localization. Although impressive at the time, their route included minimal interaction with other vehicles, only a single traffic light, and pedestrians were limited to a controlled situation at the end of the route. 

Since the DARPA Urban Challenge, there have been several other autonomous driving competitions. The first Grand Cooperative Driving Challenge (GCDC) was held in 2011. This competition focused on the interaction between autonomous vehicles. Each vehicle was able to communicate with the others via a predetermined V2V radio interface. The 2016 GCDC included cooperative lane changes and cooperative intersection handling \citep{berthacoop}.

From 2010-2014, three autonomous vehicle competitions were held in South Korea, organized by Hyundai. The 2012 competition included a simulated urban environment with traffic lights, moving vehicles, and static pedestrians. The winning system relied primarily on 2D LIDAR to detect obstacles and a GPS/IMU to localize \citep{jo2015developmentkorea}. 

In China, the Intelligent Vehicle Future Challenges (IVFCs) have been held every year since 2009 with increasing complexity. In 2012, the competition included lane keeping, traffic lights and signs, pedestrian avoidance, and merging into moving traffic. 

Another competition, but one that's dedicated to students, is the Formula Student Driverless competition which was first held in 2017. The competition required teams to complete 10 laps of a previously unknown track delimited by pylons. The vehicle employed was an electric 4WD car developed by AMZ for the Formula Student Electric challenge in 2015. The winning team, fl{\"u}ela driverless, developed a LIDAR SLAM system to localize within the track \citep{fluela}.

A new competition which began its initial development in 2016 is the Roborace competition. Eventually their goal is to have as many as 10 autonomous cars on a track racing simultaneously. The vehicular platforms themselves are standardized across teams and include the numerous sensors and compute power required for the competition. The idea will be that individual teams will develop the software that will run on these platforms \citep{roborace}.

\vspace{-2mm}

Dataset papers may serve as useful references of sensor configurations. Currently, the most popular benchmark for autonomous driving is the KITTI dataset \citep{Geiger2012Kitti}. Other self-driving datasets include the Oxford Robotcar dataset \citep{Maddern2017}, the ApolloScape dataset \citep{apolloscape}, and NuScenes \citep{caesar2019nuscenes}. Each of these datasets provides LIDAR, camera, and GPS/IMU data for different purposes.

\vspace{-2mm}

The SAE AutoDrive Challenge is unique in that it is a full self-driving competition aimed primarily at undergraduate students. With minimal resources and a constrained time frame, our team of students was able to develop \change{a functioning autonomous vehicle}. This work provides an up-to-date description of a self-driving car including information on team organization, sensor configuration, and deep learning acceleration. Such a complete picture of the self-driving development process has not been shown in literature before. Another factor that distinguishes this work is our use of deep learning for perception. Of the previously mentioned systems, only Autoware, Apollo, and Bertha (2016) claim to make use of deep learning. However Autoware has not been rigorously tested, Apollo has yet to publish results of their testing, and Bertha's updated perception suite was disabled for their competition. Furthermore, this work describes our closed-loop performance in detail with several figures, which is important but uncommon among existing works. Due to Intel being an exclusive computing sponsor of the AutoDrive competition, they have restricted teams from using NVIDIA GPUs for on-board computations. As a result, we have invested significant effort into leveraging CPUs and FPGAs for deep learning acceleration, which we discuss in this work.

\vspace{-2mm}

\begin{figure*} [ht]
    \centering
    \subfigure[Speed Zone Course]{\includegraphics[width=0.47\textwidth]{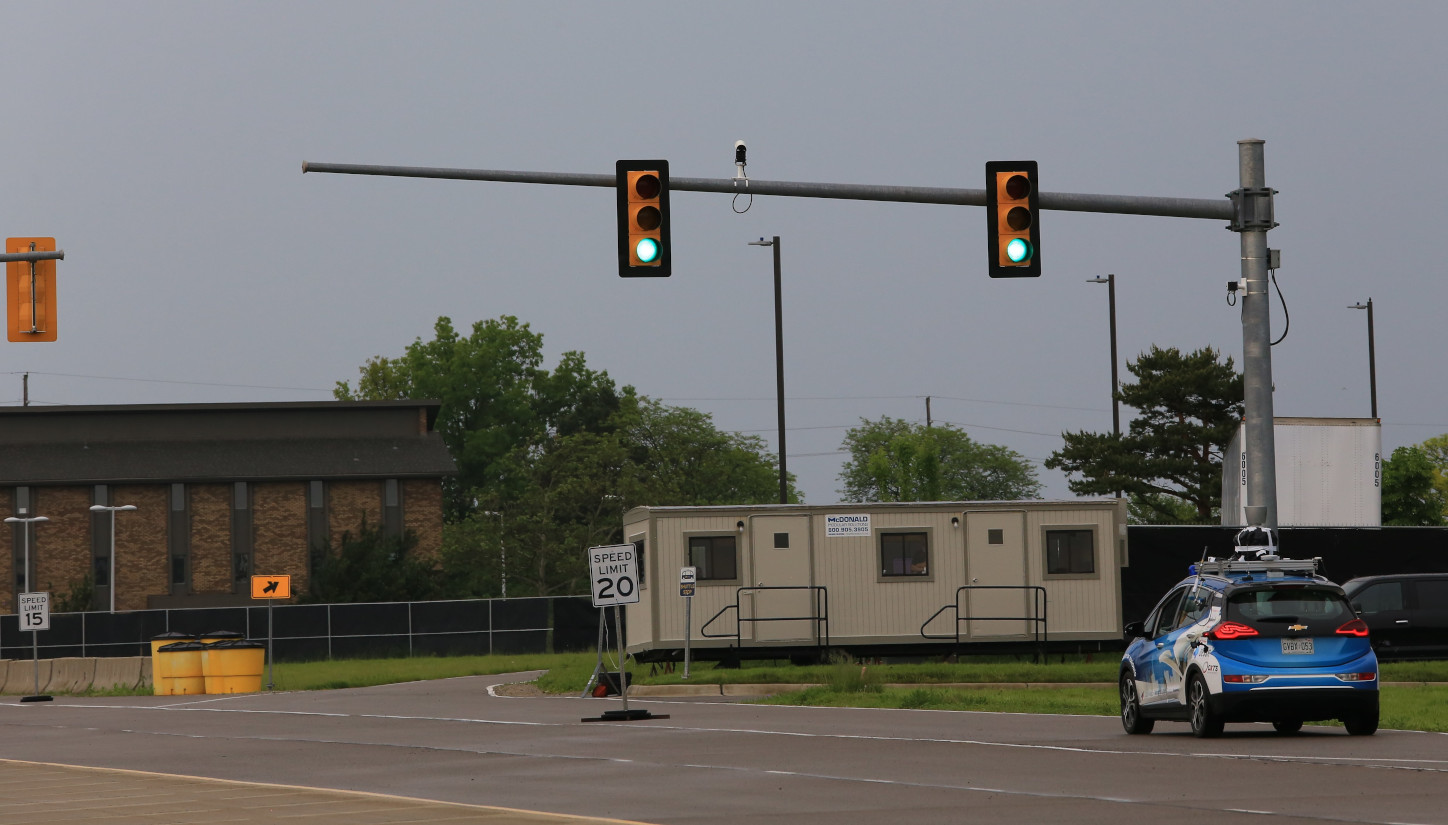}}
    \subfigure[Traffic Control Sign Challenge]{\includegraphics[width=0.47\textwidth]{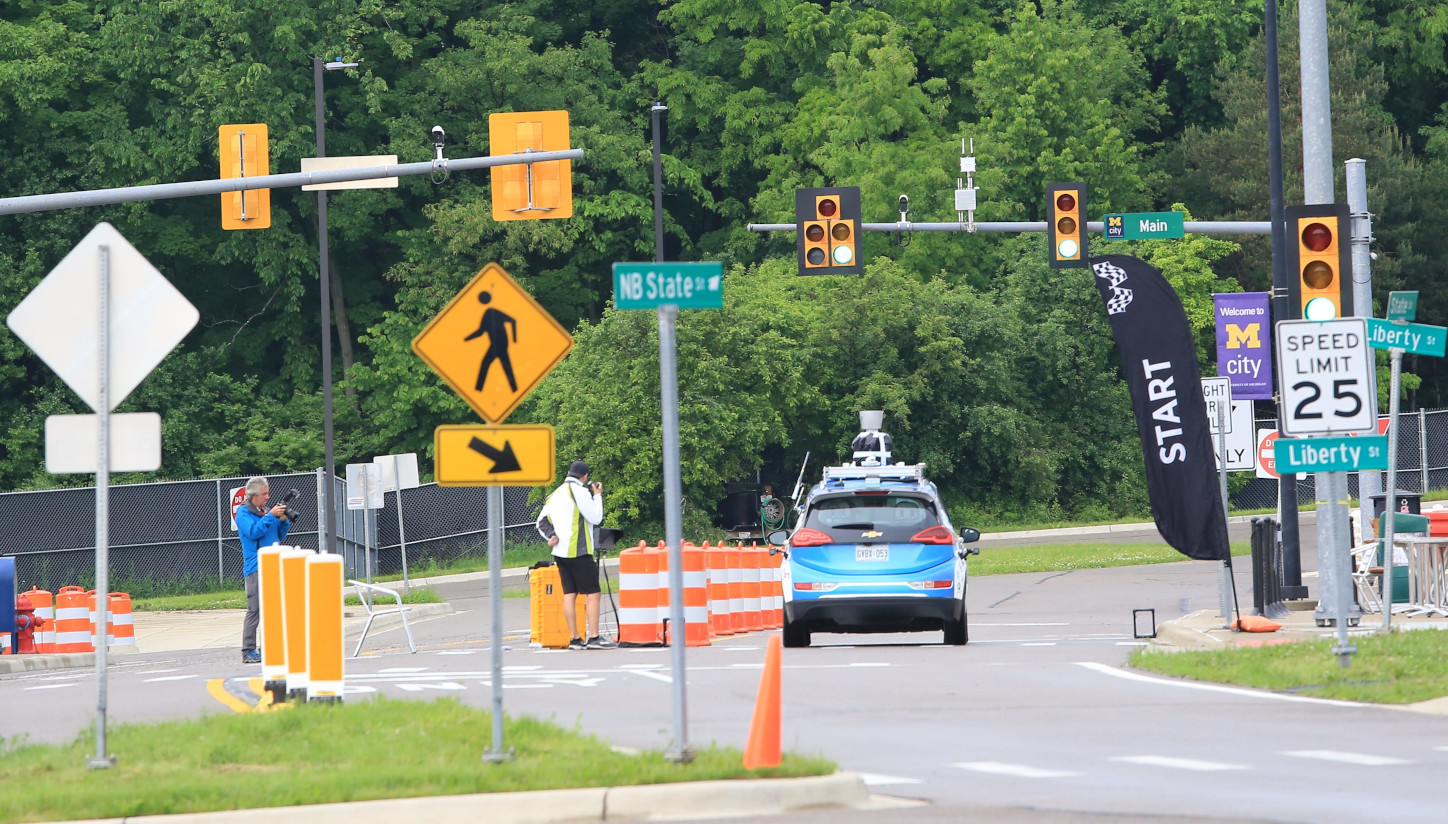}}
    \subfigure[Pedestrian Challenge]{\includegraphics[width=0.47\textwidth]{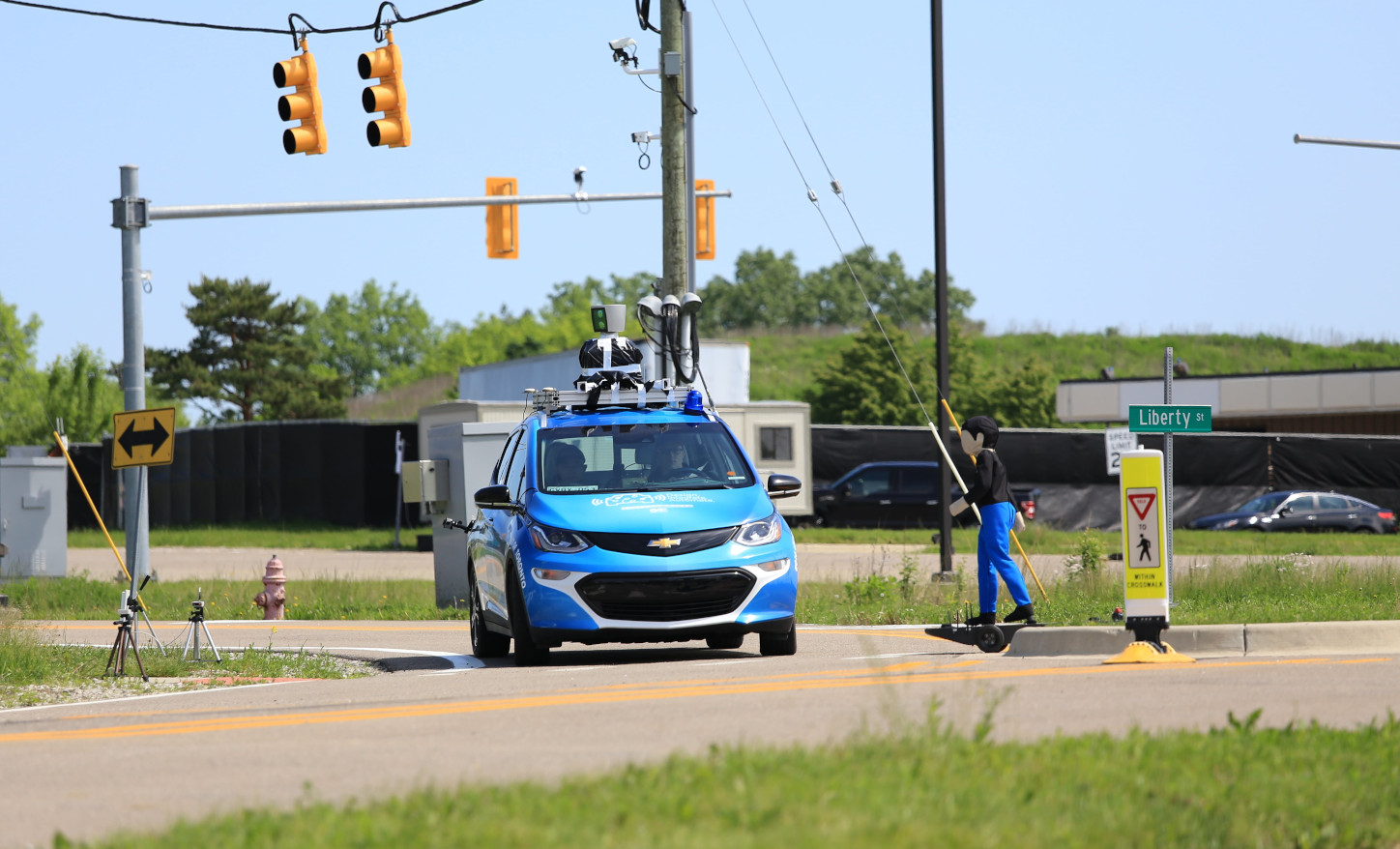}}
    \subfigure[MCity Challenge]{\includegraphics[width=0.47\textwidth]{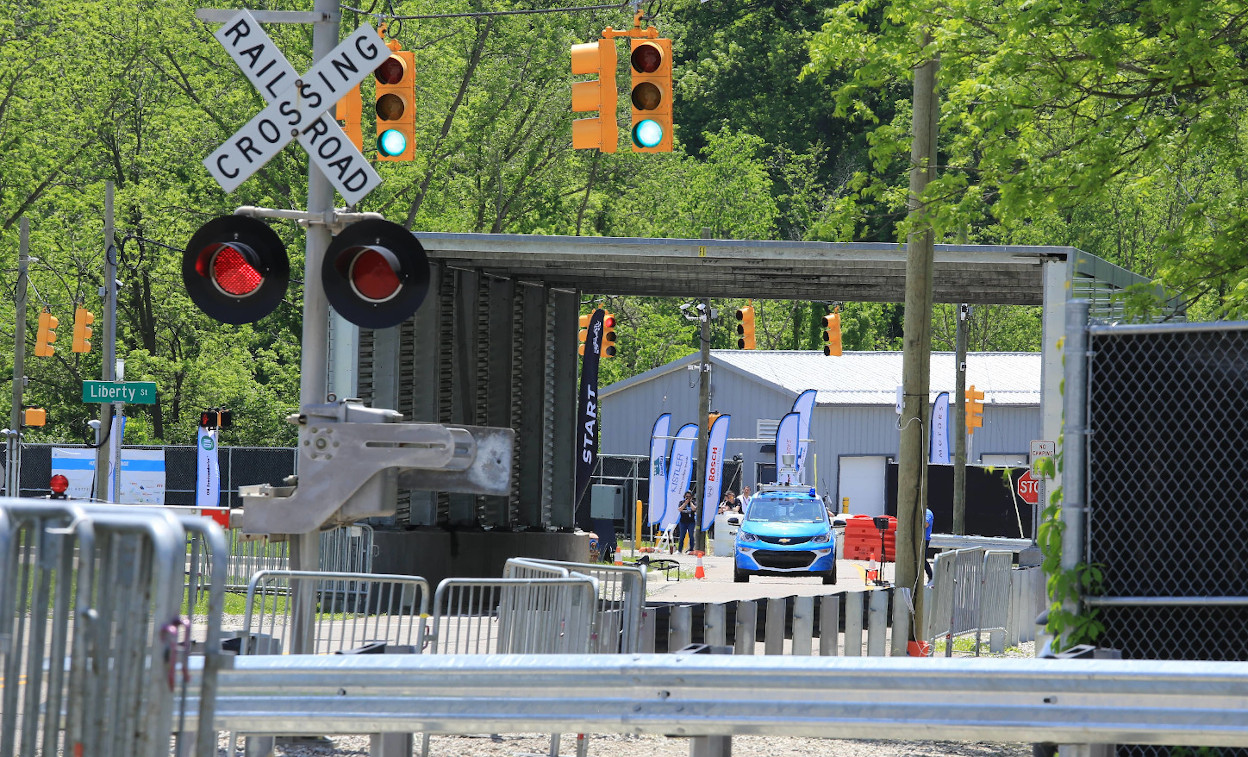}}
    \caption{Zeus during the Year 2 competition at MCity. The Year 2 competition was divided into four challenges: Traffic Control Sign Challenge (with Speed Zone Course), Pedestrian Challenge, MCity Challenge, and Intersection Challenge.}
    \label{fig:competition}
\end{figure*}

\subsection{Year 2 Competition}

The first year of the competition required teams to perform lane-keeping, stop at a series of stop signs and avoid static objects by performing lane-change maneuvers. This competition was held at GM's Desert Proving Grounds in Yuma, Arizona. No map of the course was given and multiple laps were not permitted. Lane detection was intended to be the sole source of localization information. Although the challenges were straightforward, only six months were given between the vehicle delivery date and the competition. In order to achieve a working system in time, we relied on simple, robust algorithms. A description of our Year 1 architecture can be found in \citep{burnett2018building}.

The second year was divided into four challenges: the Traffic Control Sign Challenge, Pedestrian Challenge, Intersection Challenge, and MCity Challenge. 

The Traffic Control Sign Challenge had two segments: a speed zone and a traffic sign course. The speed zone required vehicles to perform lane-keeping while abiding by posted speed limits. Traffic sign positions were not encoded in the map given to teams. The traffic sign course required vehicles to continue straight along the road unless directed otherwise by signs present on the road. There were 13 possible sign classes in total including Right-Turn-Only and Do-Not-Enter signs. The traffic sign course ended in a pull-in parking maneuver where some spots were occupied and some had handicap parking signs associated with them.

For the other three challenges, high-level GPS waypoints were given. Waypoints were placed at the center of each intersection where a turn needed to be made. These waypoints provided a global path through MCity but were insufficient for autonomous driving by themselves.

In the Pedestrian Challenge, pedestrian dummies were placed at the side of marked crosswalks and at intersections with flashing red lights. If a pedestrian is waiting at the side of the crosswalk, vehicles are required to come to a stop and wait for the pedestrian to completely cross the road. In the case that a pedestrian is stationary for longer than five seconds, the vehicle should continue driving. The most difficult scenario involves making a left-hand turn through an intersection with flashing red lights. In this case, it is necessary to check for pedestrians crossing the road both immediately in front of the vehicle and on the left-hand crosswalk.

In the Intersection Challenge, vehicles were required to react appropriately to traffic lights at a series of intersections. This challenge was the longest, requiring teams to sequence 13 intersections to complete the course.

\vspace{-1mm}

The MCity Challenge required vehicles to sequence several intersections while handling obstacles along the way. These obstacles included a tunnel, railroad crossing, static cyclist, and a dynamic animal. Several images of the Year 2 competition courses are shown in Figure~\ref{fig:competition}.

\begin{figure*} [ht]
    \includegraphics[width=0.95\textwidth]{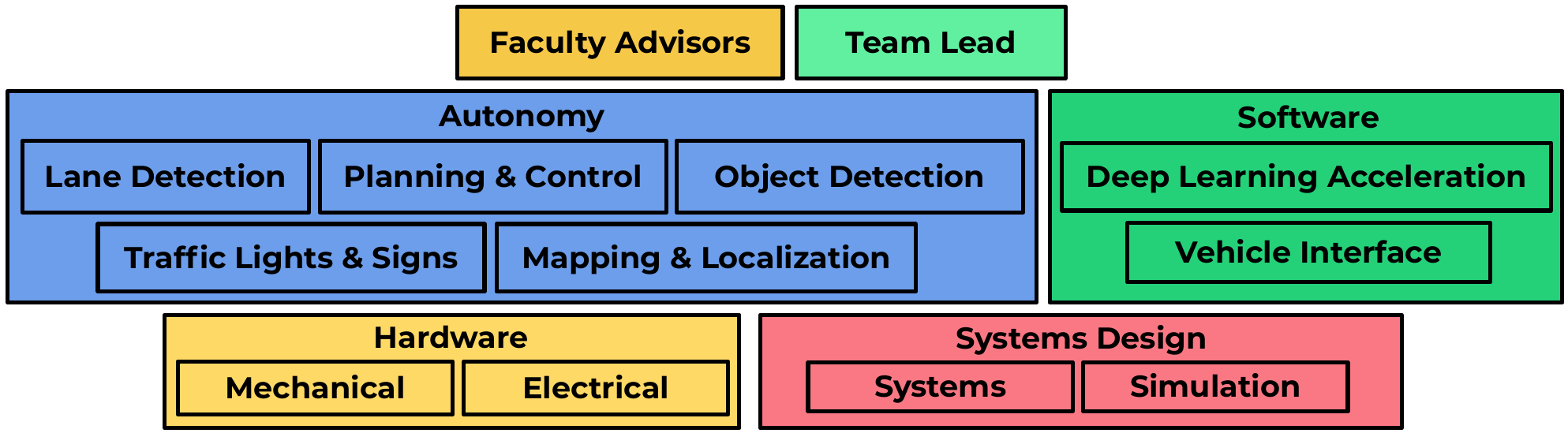}    
    \centering
    \caption{aUToronto's organizational chart. The team is roughly divided into the \change{areas} of autonomy, software, hardware, and systems design. This hierarchical structure was critical to avoid overwhelming the Team Lead.}
    \label{fig:teamorg}
\end{figure*}

\vspace{-1mm}

\section{Team Organization}
aUToronto consists of close to 100 students at the University of Toronto, the majority of which are undergraduates. Figure~\ref{fig:teamorg} depicts the team's organizational structure. The team has twelve sub-teams that cover the broader areas of autonomy, software, hardware, and systems design. Most sub-teams focus on a single component of the software stack. Each sub-team has 1-2 team leads who take responsibility for their sub-team's progress. The sub-teams themselves range in size between three and ten general members. The Team Lead makes high-level design decisions and interfaces directly with sub-team leads. This two-tiered management structure prevents the Team Lead from being overwhelmed.

\change{All team software was tracked using private Gitlab repositories. Where publicly available repositories were used, such as ROS sensor drivers, these repositories were cloned and kept in a frozen state to prevent regression. A single privileged repository was used to track the software that would run in realtime on Zeus. All third party software and adjacent projects were kept separate. A single ROS workspace is used for all aUToronto software, enabling a simple build process. aUToronto's continuous integration environment included a build script which would create a virtual machine, install all dependencies, and build the codebase from scratch. At the bare minimum, code was required to build before being able to merge into the main repository. aUToronto employed a plus-one system whereby all new features and bug fixes were submitted as a merge request and subsequently reviewed by a team lead. The team strived to follow the Google C++ coding guidelines but this was not strictly enforced. In general, we relied quite heavily on real-world tests conducted on Zeus. Individual components were tested in isolation using either data taken on Zeus for offline testing or online testing while Zeus was operating in open-loop. A set of regression tests was performed on Zeus on a weekly basis to verify that basic features were not deteriorating. Integration testing was conducted by testing the system in open-loop both offline and directly on Zeus. Simulation testing was limited to the validation of planning logic and semantic maps. The simulation environments we used included custom C++ ROS tests and the RightHook simulation environment \citep{RightHook}.}

\vspace{-1mm}

Weekly meetings are \change{held} by the Team Lead with all sub-team leads present. This serves to keep the entire team on track and to \change{encourage} communication between sub-teams. All work is carried out by the students with faculty advisors only providing occasional advice.

\begin{figure*} [hb]
    \includegraphics[width=\textwidth]{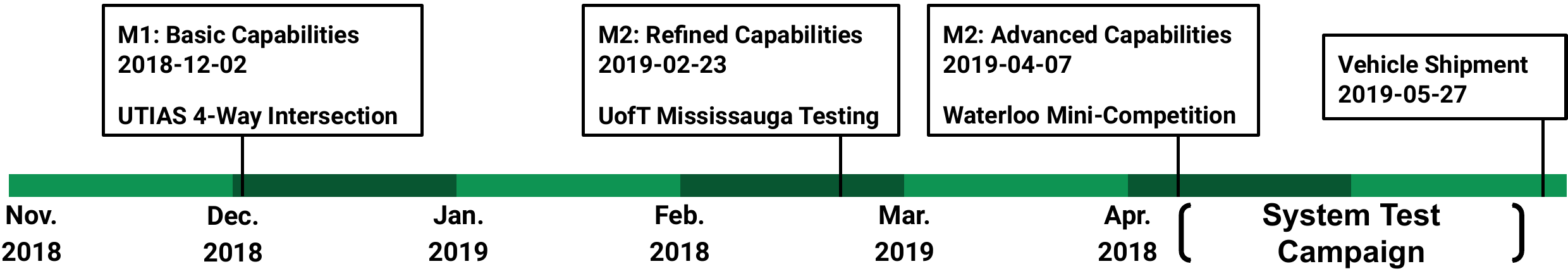}    
    \centering
    \caption{Year 2 development timeline. The timeline was set up so that major milestones incrementally added features to Zeus. Each milestone culminated in a series of real-world tests at a new location. Test locations included U of T's Mississauga campus, and the Clearpath Robotics office.}
    \label{fig:timeline}
\end{figure*}

The team consists primarily of undergraduates working in their spare time. For this reason, it is critical to have a timeline that is both aggressive and realistic. aUToronto's development timeline for Year 2 of the competition is depicted in Figure~\ref{fig:timeline}. The timeline depicts the seven months leading up to the competition. The four months prior to this point consisted of hiring, on-boarding, research, and ramping up development efforts. The main parts highlighted in the timeline are the \change{milestones} and system test campaign. 

The purpose of each milestone is to gradually improve on capabilities that can be demonstrated in real-world tests. Each milestone ends with a series of closed-loop tests at a new location. This serves to motivate the entire team to achieve a common objective in a timely manner. During the System Test Campaign, a small team worked almost daily on testing and bug fixes. This test campaign was vital for achieving good performance at the competition.

\begin{figure*} [ht]
    \centering
    \subfigure[Sensor placement side view]{\includegraphics[width=0.47\textwidth]{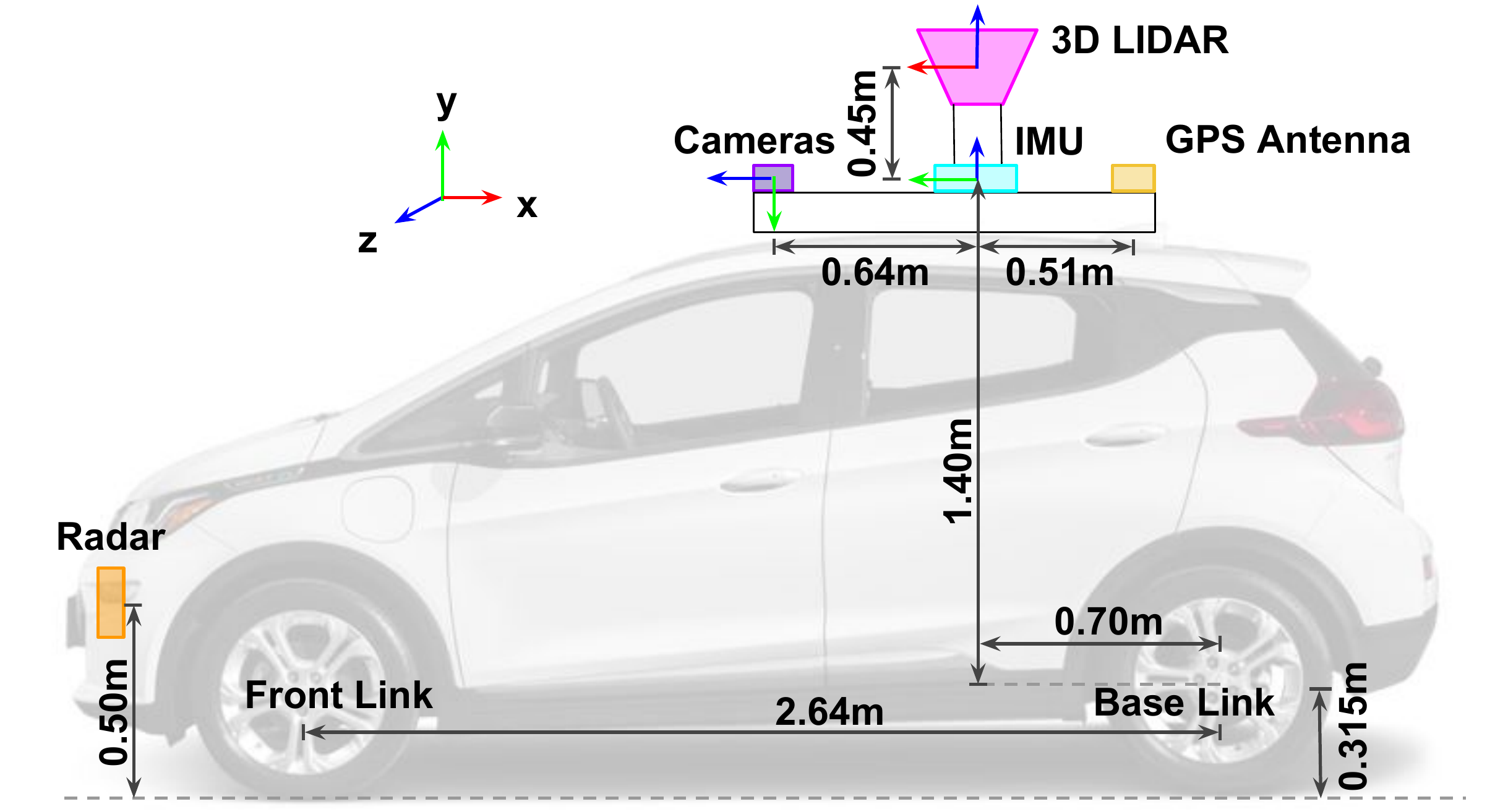}}
    \subfigure[Sensor placement top-down view]{\includegraphics[width=0.47\textwidth]{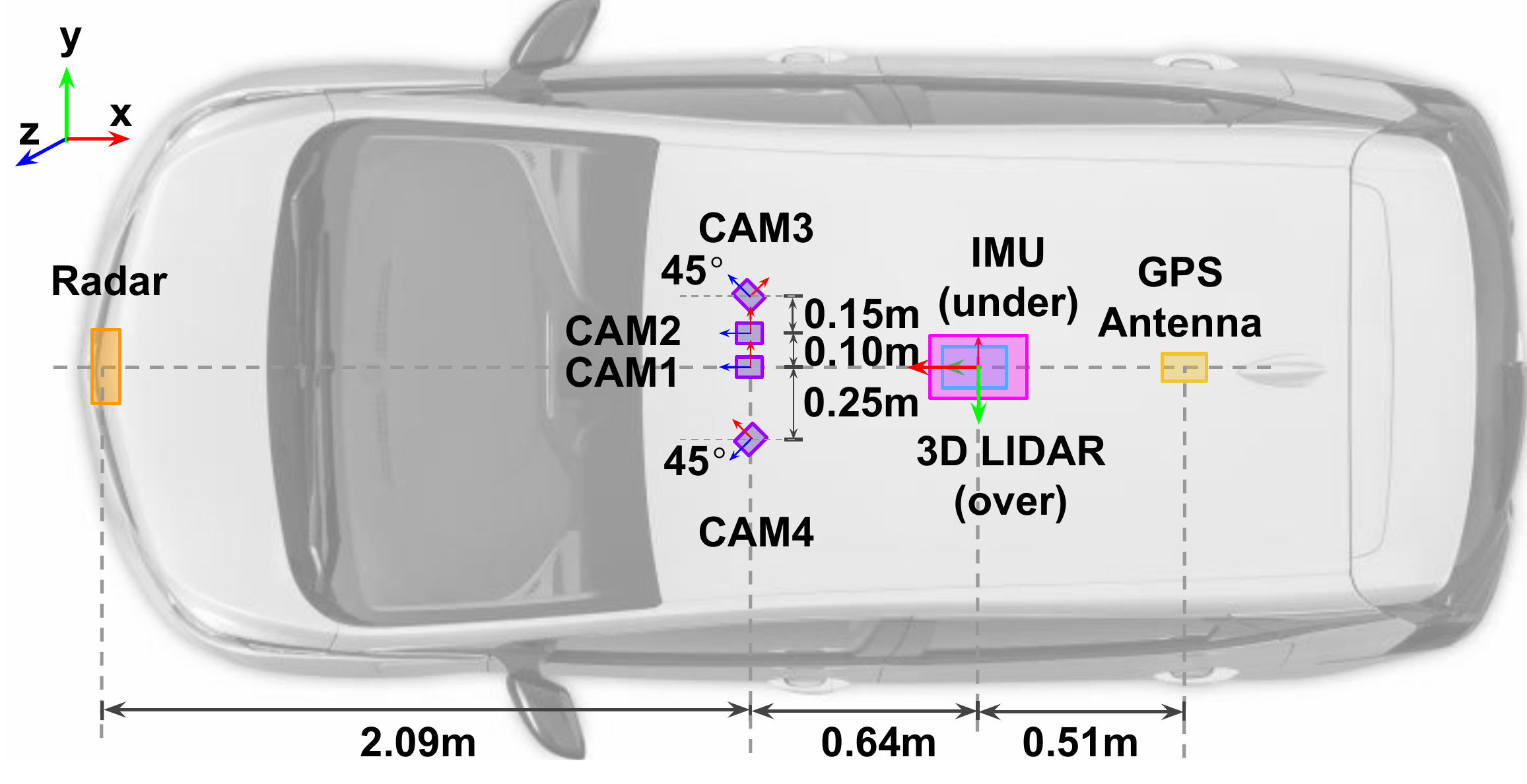}}
    \caption{Several sensors are placed along the longitudinal axis of the vehicle. Most sensors are placed so that rotation angles are multiples of 90 degrees. These two factors simplify calibration.}
    \label{fig:sensor_place}
\end{figure*}

\vspace{-1mm}

\section{System Overview}

\vspace{-2mm}

aUToronto's self-driving car, Zeus, is a 2017 Chevrolet Bolt electric vehicle. The sensor suite includes a Velodyne HDL-64 3D LIDAR, four Blackfly S monocular cameras (5.0 MP, 75 FPS), a Novatel PwrPak7 GPS/IMU, and a Continental ARS430 radar. The compute server has two Intel Xeon E5-2699 v4 processors that together contain 44 physical cores operating at 3.6 GHz. The server also contains an Intel Arria 10 FPGA for deep learning acceleration. The total cost of the hardware added to Zeus is \$175,000 CAD.

\vspace{-1mm}

Figure~\ref{fig:sensor_place} depicts the placement of sensors on Zeus. The primary factors that drove sensor placement include ease of calibration, field of view, and detection range. Camera 1, 3D LIDAR, IMU receiver, and GPS antenna are all placed along the longitudinal axis of the vehicle. Where possible, sensors are placed such that the rotation angles between them are a multiple of 90 degrees. These two factors make the calibration process simpler. Sensor mounts were designed so that the transformation obtained from the CAD model would be close to the calibrated value.

\vspace{-1mm}

\begin{figure} [t]
    \includegraphics[width=0.55\columnwidth]{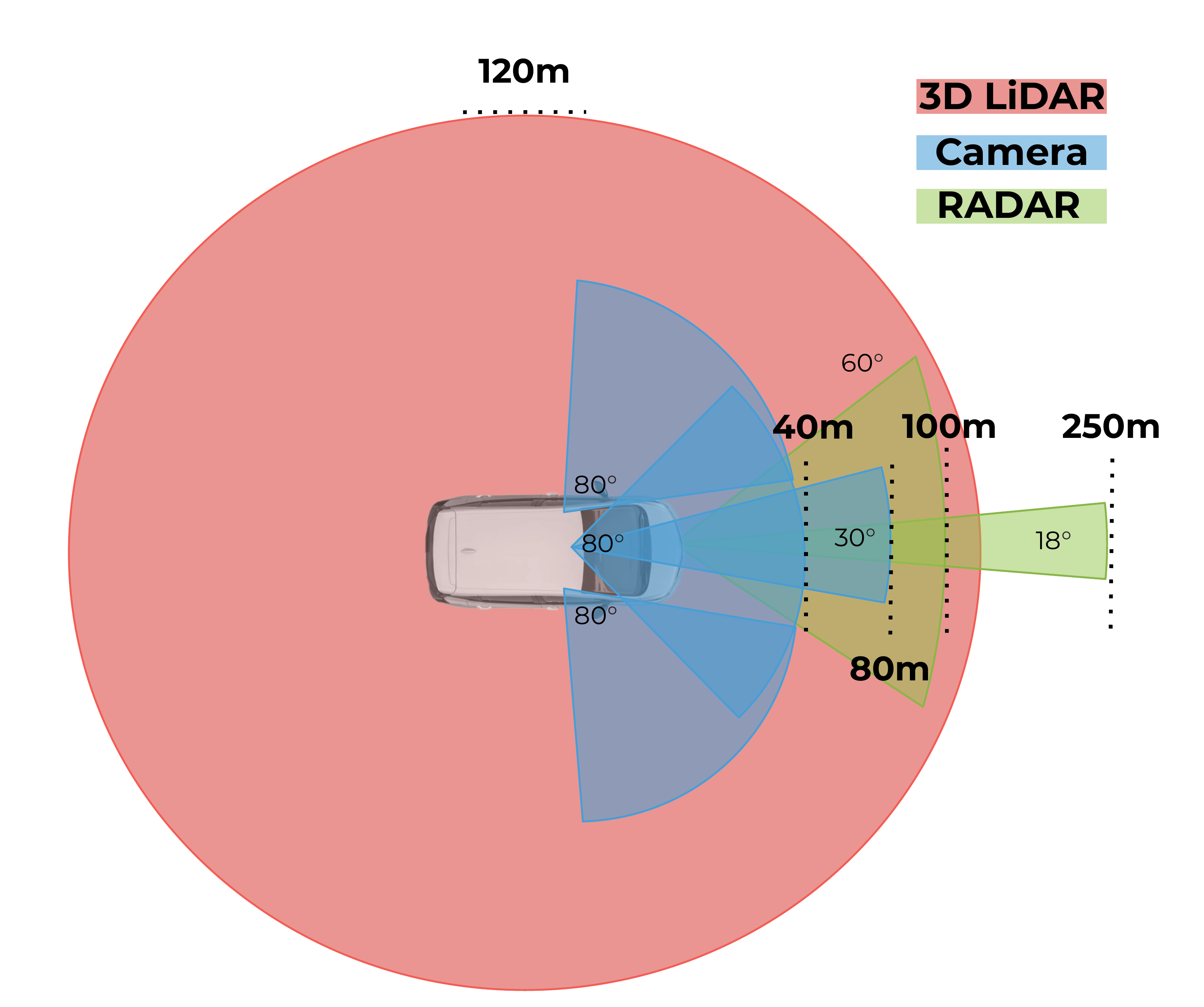}    
    \centering
    \caption{This figure depicts the fields of view of sensors on Zeus. Note that the front-facing cameras provide close to 170 degrees of horizontal field of view. A \change{narrow-field-of-view} camera boosts visual detection range.}
    \label{fig:fov}
\end{figure}

Figure~\ref{fig:fov} depicts sensor fields of view. Of the four monocular cameras, two point straight forwards, and two at \change{45-degree} angles. This configuration was chosen to maximize the horizontal field of view. Cameras 1, 3, 4 have a \change{5-mm} lens, which results in an \change{80-degree} field of view. Camera 2 uses a \change{16-mm} lens, which results in a \change{30-degree} field of view. This \change{narrow-field-of-view} camera doubles the visual detection range of objects from 50~m up to 100~m.

\vspace{-1mm}

The Velodyne HDL-64 has a rated range of 120 m. In our experiments, the effective range of pedestrian and vehicle detection is closer to 40 m and 80 m, respectively. An automotive radar sensor was installed behind the front bumper. This sensor was not used during the Year 2 competition, but is intended as an extra layer of safety in the future.

\vspace{-1mm}

To calibrate the extrinsic transformation between Camera 1 and the Velodyne, we use the open-source toolbox described in \citet{Unnikrishnan2005}. Calibration is performed by moving a checkerboard target in front of the vehicle and capturing image and LIDAR pointcloud pairs. For other extrinsic transformations, hand measurements and CAD model values are used. We also design our perception algorithms to be robust to minor calibration error.

\vspace{-1mm}

Mechanical sensor mounts are attached to a rectangular rack consisting of double-wide aluminum extrusion. The rack is mounted rigidly to the vehicle's sport rails via custom-machined interface components. This structure provides a base for modular and reconfigurable sensor attachments. A central tower structure is employed to mitigate occlusion of the Velodyne HDL-64.

\vspace{-1mm}

All electronics are powered via the Chevrolet Bolt's Auxiliary Power Supply, which can supply up to 1~kW at 12~V. A liquid cooling system was installed to regulate the temperature of the CPUs and FPGA.

Autonomous control of the vehicle is facilitated through the Vehicle Interface: a software API developed by \mbox{aUToronto} that enables communication between the ROS network and the vehicle. Communication is achieved using a serial data connection over the vehicle's CAN buses. This enables the autonomy software to control torque, steering, and transmission. Each team was given documentation from GM to enable the development of this interface.

\vspace{-1mm}

\begin{figure} [ht]
    \includegraphics[width=0.55\columnwidth]{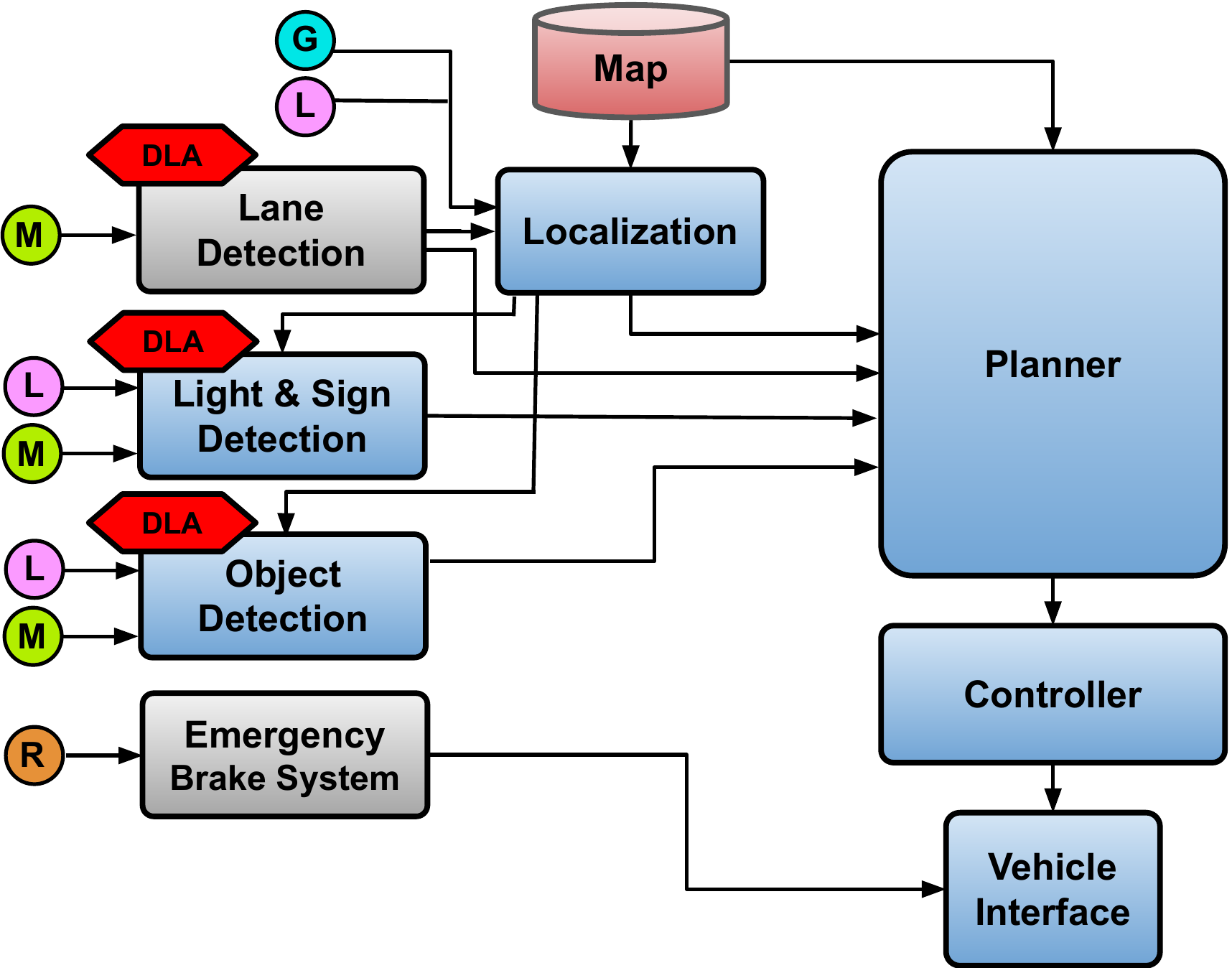}    
    \centering
    \caption{This figure depicts Zeus' software architecture. L: 3D LIDAR, M: Monocular Cameras, R: Radar, G: GPS/IMU, DLA: Deep Learning Acceleration. Localization is either Novatel GPS/IMU or Applanix LIDAR localization. \change{Lane Detection and Emergency Braking (Gray) were not used at the Year 2 competition.}}
    \label{fig:swarch}
\end{figure}

Figure~\ref{fig:swarch} depicts Zeus' software architecture. Sections~\ref{objectsection},\ref{lnssection},\ref{lanesection},\ref{locsection} provide detailed information on \change{Object Detection, Light and Sign Detection, Lane Detection, and Localization.} The perception nodes \change{generate} a high-level abstraction of the environment around Zeus, and pass this information on to the planner. \change{The planner first} generates a global path for the vehicle by searching through a road graph and reaching each high-level waypoint. \change{Internally, we convert third party semantic maps into our own format which is based on OpenStreetMaps. Our internal format has a graph-like structure which can easily be traversed using a graph search algorithm. The planner also} generates a local trajectory \change{by linking together the centerlines of the desired road segments and performing minor path smoothing}. The controller then outputs torque and steering commands which are enacted by the Vehicle Interface. The design of the planner and controller is described in sections~\ref{planningsection},\ref{controlsection}. All software was written in C++ and \change{runs} on ROS \change{Kinetic} \citep{ros} using the compute server described above.

\vspace{-1mm}

\section{Deep Learning Acceleration} \label{dlasection}

\vspace{-1mm}

Autonomous driving software presents a heavy computational load. The optimization of this software on our computing platform is driven by reducing end-to-end latency between sensor inputs and control actions. Zeus relies on multiple DNNs to process \change{high-resolution} camera images.

\vspace{-1mm}

Each software component is implemented as one or more nodes using ROS as the underlying infrastructure. Many nodes must run concurrently, and the end-to-end latency from sensor inputs to control actions should not exceed 100 ms. This requirement is based on the commonly used rule-of-thumb for what constitutes a real-time system. DNNs are the most significant challenge to minimizing latency.

\vspace{-1mm}

The Intel Crystal Rugged server has a theoretical peak floating-point operation throughput of 1.8 TFLOPS. At a glance, this appears sufficient for running inference of several DNNs in a timely fashion. However, in practice it is challenging to select a DNN architecture that is capable of running in \change{realtime}, on sufficiently large images, without a GPU. To address this constraint, our team configured an Intel Arria 10 FPGA to be used as a DNN inference accelerator.

\vspace{-1mm}

\begin{figure} [h]
    \includegraphics[width=0.85\columnwidth]{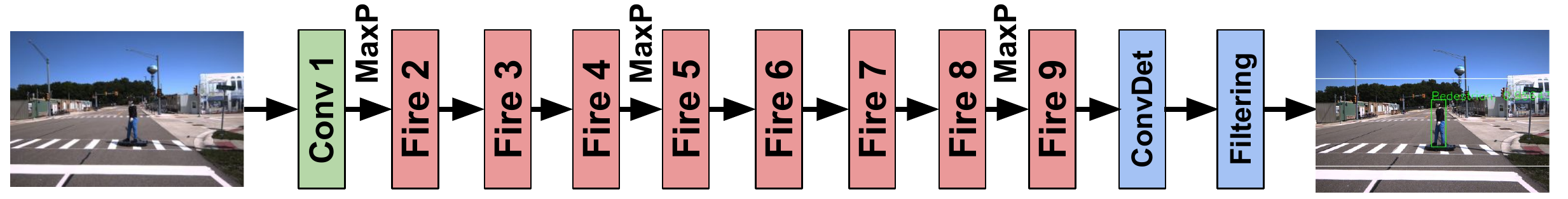}    
    \centering
    \caption{The SqueezeDet architecture. Fire: 1x1 convolutions followed by a RELU operation, 1x1 convolutions and 3x3 convolutions computed in parallel, and another RELU operation. ConvDet: bounding boxes are directly regressed from a dense anchor grid. Filtering: non-maximum suppression. MaxP: max pooling.}
    \label{fig:squeezearch}
\end{figure}

\begin{figure} [t]
    \includegraphics[trim={2.1cm 0 4.8cm 1cm}, clip , width=0.50\columnwidth]{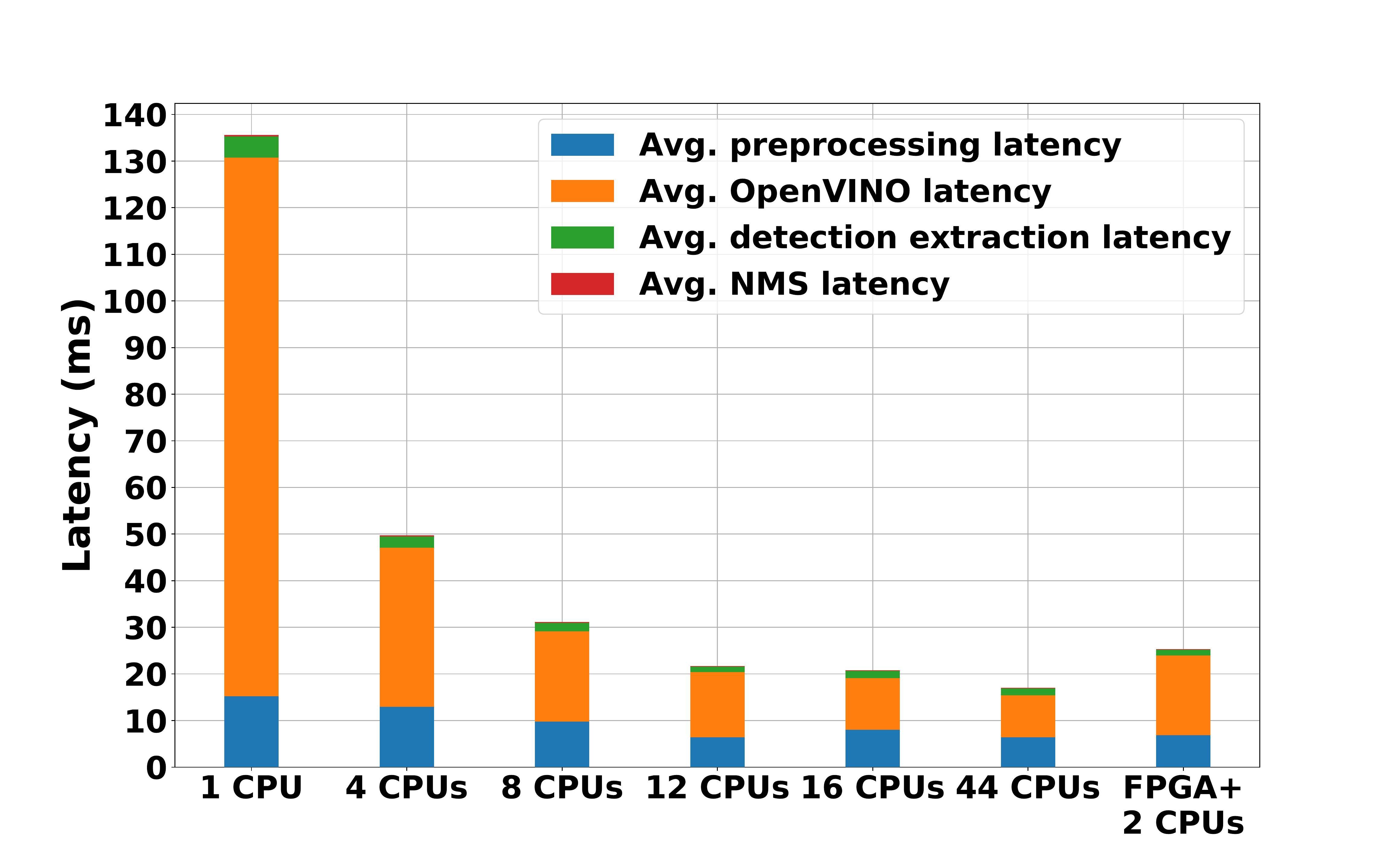}    
    \centering
    \caption{SqueezeDet inference on 2x Intel Xeon E5-2699 v4 processors with support of Intel OpenVINO. This figure demonstrates the diminishing returns of accelerating SqueezeDet with CPUs.}
    \label{fig:openvino-benchmark}
\end{figure}

 We observed that SqueezeDet exceeded \change{its} competitors in terms of speed for medium-size images \citep{wu2017squeezedet}. \change{A diagram of the SqueezeDet architecture is given in Figure~\ref{fig:squeezearch}. Given six CPU cores, SSD300 yielded a latency of 100 ms. A ResNet-50 backbone paired with SqueezeDet's ConvDet layer yielded a latency of 110 ms. Compared to SqueezeDet's latency of 40 ms, neither of these options were suitable. YoloV3 is a good alternative, promising improved performance and high inference speed \citep{redmon2018yolov3}. However, YoloV3 is not natively supported by OpenVINO.} Figure~\ref{fig:openvino-benchmark} \change{shows that} there are diminishing returns to accelerating DNNs on multiple CPUs. This places an upper limit on the complexity of a viable DNN architecture \change{and the size of the input image}.

\change{
There are couple of reasons why SqueezeDet runs so well on OpenVINO. First, it is comprised entirely of convolutional layers, making it straightforward to accelerate.  Second, SqueezeDet does not use multi-scale features as is done in SSD and YoloV3. In those other networks, higher-level features in the network are concatenated to the feature volume immediately preceding bounding box regression. This allows those networks to achieve better performance across multiple scales at the expense of computation time.}

\vspace{-1mm}

The key to leveraging our server's computing power is the Intel OpenVINO SDK. OpenVINO provides runtime libraries optimized for Intel CPUs. Operations for DNN inference can benefit from SIMD instructions and multithreading. Moreover, OpenVINO contains pre-compiled accelerator images that target the Arria 10 FPGA. We implemented a library in C++ called \textit{Zeus DLA} that uses these features. It preprocesses input images, \change{performs inference using} OpenVINO, and performs \change{unsupported operations} in C++.

\vspace{-1mm}

We benchmarked the performance of Zeus DLA by running SqueezeDet with a different number of CPU cores and single-batched images on the CPUs as well as on the Arria 10 FPGA. Using 8 out of the 44 available CPU cores, our inference library can accomplish inference in \change{32 ms,} as shown in Figure~\ref{fig:openvino-benchmark}. Using the FPGA accelerator for SqueezeDet, the inference time with Zeus DLA is 26 ms.

\vspace{-1mm}

During the Year 2 competition, the Pedestrian Challenge presented the most demanding computational load. In this case, three DNNs are used to detect pedestrians on three different \change{cameras;} in \change{addition,} a single DNN is used to detect traffic lights. Three of the DNNs are assigned 8 CPU cores each, and the fourth DNN is run on an FPGA. This leaves 20 CPU cores for the remainder of the software stack.

\section{Machine Learning and Data Collection}

\vspace{-2.5mm}

Zeus' perception algorithms rely on \change{DNNs} to localize objects within 2D images.  Early in development, it became apparent that using publicly available data was \change{insufficient} for \change{performing well} in real-world driving. The remainder of this section describes our experience building a dataset and fine-tuning our \change{DNNs}.

\vspace{-2.5mm}

\change{The Year 2 competition required us} to detect thirteen different sign classes, three different traffic light configurations, and pedestrians. A common starting point is to use publicly available training data. Bosch has a good dataset for traffic lights \citep{BehrendtNovak2017ICRA}. However, the training set was not sufficient by itself to obtain good performance in our experiments. KITTI is one \change{of} the most popular self-driving datasets, with over 7000 training images including pedestrians, cyclists, and cars \citep{Geiger2012Kitti}. Training on only \change{KITTI data} did not generalize well to \change{experiments on our vehicle}. In order to achieve better performance, a custom \change{training set} was required.

\vspace{-2.5mm}

Most of our dataset was collected on public roads in Toronto in various weather conditions. Images were extracted at one frame per second. In some cases, it was not possible to obtain all the required data on public roads. In this case, replicas of the competition equipment were purchased and set up on private roads at \change{U of T}. This included the same traffic lights used at MCity, American traffic signs, and a replica of the \change{competition} pedestrian dummy. By obtaining replicas of their equipment, it was possible to optimize our \change{DNNs for the competition environment}. 

\vspace{-2.5mm}

All of our \change{own} data \change{were} collected in Toronto. However, the competition was to be held in Ann Arbor, Michigan. For this reason, there was some concern that our dataset might be too optimized for Toronto. To combat this, images were extracted from 4K dashcam YouTube videos of New York, Pittsburgh, and Vancouver. These videos were used to increase the variation within our traffic light dataset.

\vspace{-2.5mm}

The final dataset consisted of 17000 images of traffic lights, 15000 images of pedestrians and cars, and 35000 images of traffic signs. \change{We} sought to outsource \change{the data labelling} task. Initially, we tried Amazon's Mechanical Turk but ran into several issues of label quality. \change{A common solution to improve label quality is to have each image labeled by at least three different workers.} Then, intersection-over-union (IoU) metrics can be used to \change{automatically flag discrepancies. Even with these measures in place, the quantity of false negatives and coarsely drawn boxes increases substantially in frames with more than ten objects.} 

\vspace{-2.5mm}

\begin{figure*} [h]
    \centering
    \subfigure[Pedestrians and Cars]{\includegraphics[width=0.47\textwidth]{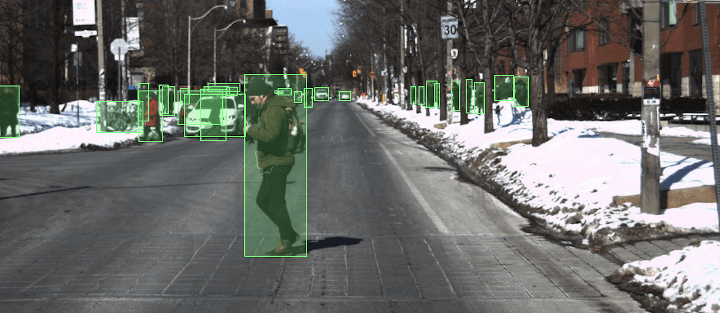}}
    \subfigure[Traffic Lights]{\includegraphics[width=0.47\textwidth]{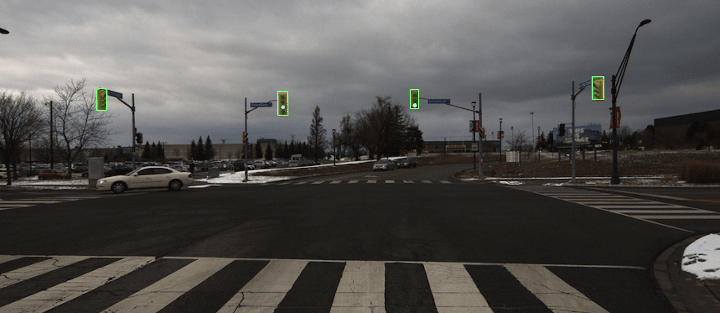}}
    \caption{This figure shows examples of the quality of Scale annotations for the tasks of object detection and traffic lights respectively. Scale annotations tend to be tight and have exceptional precision and recall.}
    \label{fig:scale}
\end{figure*}

To achieve better labels, we switched to using the Scale.ai data labelling service. At a slightly higher price, they guarantee precision, recall, and IoU targets on specified objects. Switching to Scale labels resulted in 5-10\% absolute improvement on our validation sets. The lesson here is that the quality of training labels can have a substantial impact on performance. \change{Examples of pedestrian, car, and traffic light labels are depicted in Figure~\ref{fig:scale}. This dataset will remain private while U of T is still competing in the AutoDrive Challenge.}

\vspace{-2.5mm}

aUToronto team members worked for several months to tune hyperparameters to reach at least 90\% precision and recall for each visual perception task. \change{These hyperparameters included the structure of the loss function, batch size, anchor sizes, pre-processing steps, data augmentation steps, and the probability threshold for positive detections. Our team's experimentation was bottle-necked by access to GPU resources.} When \change{object tracking was included}, the closed-loop performance \change{was qualitatively observed to output less false positives and less false negatives}. In general, collecting more training images resulted in better performance but with diminishing returns. Combining our dataset with other public datasets did not substantially improve performance. We observed that reducing the number of classes \change{led} to slightly better performance. Over 2000 images of our pedestrian dummy were added to the training set to ensure that this object would be detected. In experiments where pedestrian detection failed unexpectedly, we \change{added these images to} the training set.

\vspace{-2.5mm}

\change{The loss function used to train SqueezeDet is given in (\ref{squeeze_loss}) below. Bounding boxes were obtained by regressing four parameters: $(\delta x_{ijk}, \delta y_{ijk}, \delta w_{ijk}, \delta h_{ijk}) = R$ relative to K anchors at each grid location $(i,j)$ in the output. $I_{ijk} = 1$ if the k-th anchor at $(i,j)$ has the largest overlap with the ground truth bounding box. $\gamma_{ijk}$ is the predicted confidence score of the DNN. $\gamma^G_{ijk}$ is obtained by computing the IOU between the predicted bounding box and the ground truth. We found SmoothL1 loss on the bounding box parameters to achieve slightly better performance than simple squared error. The final term in the loss function corresponds to cross entropy over the box classes where $\ell_c^G \in \{0,1\}$, $p_c \in [0,1]$, $c \in [1,C]$. We used AdamOptimizer to train SqueezeDet using an initial learning rate of 0.001 for five epochs with a batch size of 20. We set $\lambda_{bbox} = 5$, $\lambda_{conf}^+ = 7.5$, $\lambda_{conf}^- = 2.5$. $W$ and $H$ correspond to the width and height of the input image. Our tuned threshold for positive detections was a confidence score of 0.4.}

\vspace{-5mm}

\begin{equation}
\begin{split}
\mathcal{L} &= \frac{1}{N_{obj}}  \lambda_{bbox} \sum_{i,j,k,r \in R} I_{ijk} \text{SmoothL1}(\delta r_{ijk} - \delta r^G_{ijk}) \\
 &+ \sum_{i,j,k} \Big[ \frac{\lambda_{conf}^+}{N_{obj}} I_{ijk} (\gamma_{ijk} - \gamma^G_{ijk})^2 + \frac{\lambda_{conf}^-}{WHK - N_{obj}} (1 - I_{ijk}) \gamma_{ijk}^2 \Big] + \frac{1}{N_{obj}} \sum_{i,j,k} I_{ijk} \ell_c^G \text{log}(p_c)
\end{split}
\label{squeeze_loss}
\end{equation}

\vspace{-5mm}

\begin{figure*} [ht]
    \includegraphics[width=\linewidth]{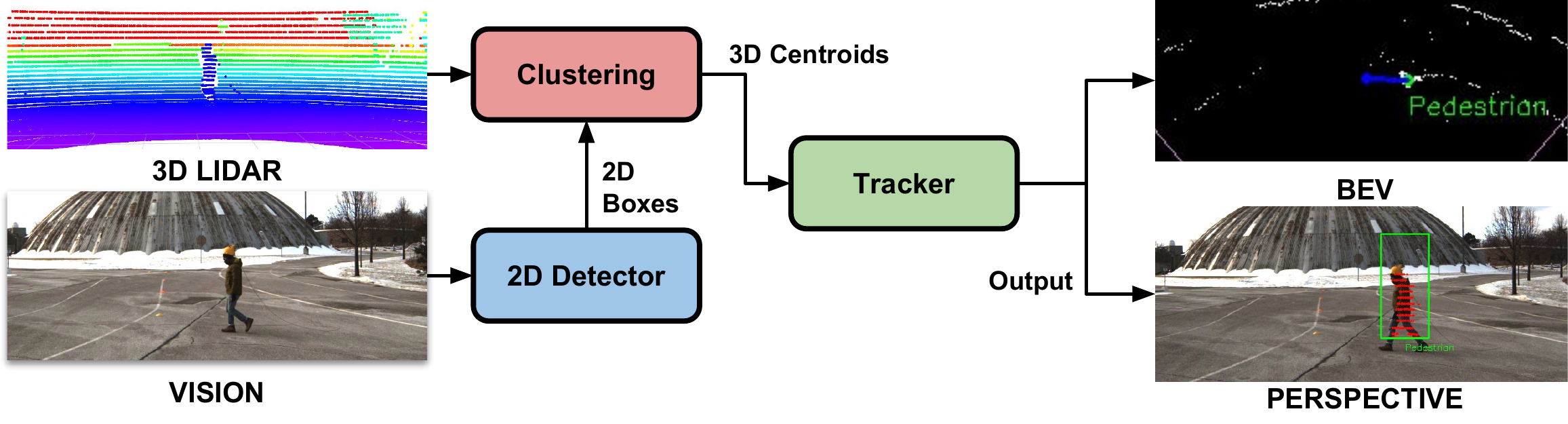}
    \centering
    \caption{\change{aUToTrack: our} pipeline for 3D object detection and tracking.}
    \label{fig:ros_diagram}
\end{figure*}

\section{Object Detection and Tracking} \label{objectsection}
aUToTrack consists of an off-the-shelf vision-based 2D object detector paired with a LIDAR clustering algorithm to extract a depth for each object. \change{Ego-vehicle} localization is then used to localize objects in a \change{static global} reference frame. Given these 3D measurements, we then used greedy data association and a linear Kalman filter to track the position and velocity of each object. Figure~\ref{fig:ros_diagram} illustrates the aUToTrack pipeline. In \citet{burnett2019aUToTrack}, we demonstrated that aUToTrack accurately estimates the position and velocity of pedestrians \change{on both the KITTI Object Tracking benchmark and our own dataset, UofTPed50.} \change{We have made this dataset publicly available, and it can be accessed using the link below} \footnote{\change{\url{www.autodrive.utoronto.ca/uoftped50}}}. \change{An updated version of our pipeline runs in less than 50ms on CPUs.}

\vspace{-1mm}

The SAE AutoDrive Challenge has so far restricted GPUs \change{from being used} for on-board computations. Thus, significant effort was invested into \change{designing a lightweight pipeline} capable of running on CPUs. With this in mind, SqueezeDet  was chosen \change{as our 2D detector} for the competition \citep{wu2017squeezedet}.

\vspace{-3mm}

\change{
\subsection{Related Work}

R-CNN was the work that first established CNNs as the state of the art for object detection in images \citep{girshick2013rcnn}. In their subsequent work on Faster R-CNN, the two-stage approach to object detection was introduced \citep{ren2015faster}. In a two-stage approach, a set of bounding box proposals is first generated by a region proposal network. In the second stage, these proposals are refined and classified. 

YOLO and SSD are credited with popularizing the notion of a single-stage detector \citep{redmon2016you} \citep{liu2016ssd}. SSD introduced the concept of anchor boxes in which each position in the output feature tensor has a set of predetermined box shapes associated with it. Detection then consists of determining a score for each anchor box and regressing offsets with respect to the anchor boxes. Single-stage detectors tend to be more computationally efficient but have historically achieved lower accuracy than two-stage approaches.

3D detectors estimate the centroid and volume of objects using a 3D bounding box. Before deep networks were applied to this problem, traditional robotics pipelines for object detection involved several common steps. First, the ground plane is extracted. Second, the remaining points are clustered. Third, the clusters are classified using a classical machine learning approach such as a support vector machine. Finally, the objects are tracked using Kalman filtering techniques \citep{himmelsbach2008} \citep{moosmann2009segmentation}.

As of this writing, the top approaches for 3D object detection fuse both vision and LIDAR data together to take advantage of both the accurate metric information that LIDAR provides and the semantic information from vision. Two such networks are Frustum PointNets \citep{qi2017frustum} and AVOD  \citep{avod}.

AVOD uses a feature pyramid network to extract features from an image and a top-down representation of a pointcloud. These features are then fused by cropping and resizing the sections of the input feature space that correspond to the projected 3D anchor boxes. The network then generates a set of 3D proposals, repeats the crop and fusion step, and finally regresses a set of 3D boxes using fully connected layers.

Frustum PointNets first obtains a 2D bounding box using a vision-based detector. They then use a PointNet to process the points which fall within the bounding box when projected onto the image plane. The PointNet is trained to regress a single 3D bounding box from this subset of the pointcloud \citep{qi2017pointnet}.

aUToTrack takes inspiration from Frustum PointNets by using a 2D detector to obtain initial object locations. This greatly reduces the dimensionality of the pointcloud and allows us to employ traditional robotics approaches in realtime. Our tracking approach takes inspiration from \citet{bewley2016simple}.
}

\vspace{-2mm}

\subsection{Clustering}

\vspace{-2mm}

\change{We} restrict our attention to points in front of the vehicle up to 40 m away, and 15 m to each side. \change{A range of 40 m was determined experimentally to work well while also providing enough time for a vehicle travelling at our maximum speed of 40 km/h to stop within our desired maximum acceleration of 2 m/s$^2$.} We subsequently segment and extract the ground plane \change{using RANSAC, followed by Linear Least Squares for refinement. We smooth the ground plane parameters temporally using a Kalman filter.} The remaining points are then transformed into the camera frame \change{using a pre-calibrated extrinsic transformation matrix $\textbf{T}_{cv}$}. A corresponding set of image locations \change{is} obtained by projecting the LIDAR points onto the image \change{using a perspective projection and the intrinsic camera matrix $\textbf{K}$.}

\vspace{-2mm}

We retrieve the LIDAR points that lie inside each 2D bounding box. \change{We then use PCL's Euclidean clustering on the corresponding 3D LIDAR points \citep{Rusu_ICRA2011_PCL}}. Several heuristics are used to choose the best cluster. These heuristics include comparing the detected distance to the expected size of the object and counting the number of points per cluster. \change{The position of the object is then computed as the centroid of the points in the cluster.} This algorithm is given in pseudo-code below.

\begin{algorithm}
	\caption{aUToTrack Clustering}
	\begin{algorithmic}[1]
		\INPUT pointcloud $\textbf{p} \in \mathcal{R}^{3 \times n}$ with n points, a set of 2D bounding box detections $\textbf{B} \in \mathcal{R}^{4 \times m}$
		\OUTPUT A list of K object centroids $\textbf{y}_k = [x~~y~~z]^T$
		\STATE $\textbf{p} \leftarrow \text{Passthrough}(\textbf{p}, W, L, H)$
		\STATE $\textbf{g} \leftarrow \text{Passthrough}(\textbf{p}, W_2, L_2, H_2)$
		\STATE $(a, b, c, d) \leftarrow \text{RANSAC}(\textbf{g})$
		\STATE $(\hat{a}, \hat{b}, \hat{c}, \hat{d})  \leftarrow \text{Kalman-Filter}(a, b, c, d) $
		\STATE $\textbf{g} \leftarrow \text{inliers}(\textbf{g}, \hat{a}, \hat{b}, \hat{c}, \hat{d})$
		\STATE $\{\textbf{p}\} \leftarrow \{\textbf{p}\} - \{\textbf{g}\}$
		\STATE $\bar{\textbf{p}} \leftarrow \textbf{T}_{cv} \bar{\textbf{p}}$
		\STATE $\textbf{u} \leftarrow \textbf{K} \bar{\textbf{p}} / \bar{\textbf{p}}_z$
		\FOR{$\textbf{b} = (c_x, c_y, w, h) \in \textbf{B}$}
		\STATE $\textbf{p}_b \leftarrow \text{Passthrough}(\bar{\textbf{p}}, \textbf{u}, \textbf{b})$
		\STATE $\text{clusters} \leftarrow \text{Euclidean-Clustering}(\textbf{p}_b)$
		\STATE $\text{best-cluster} \leftarrow $ Heuristics(clusters)
		\STATE $\textbf{y}_k = $centroid(best-cluster)
		\ENDFOR
	\end{algorithmic}

\end{algorithm}

\subsection{Tracker Setup} \label{sec:trackerSetup}

For each object, we keep a record of the state, $\hat{\mathbf{x}}$, covariance, $\hat{\mathbf{P}}$, class, object shape $(w,l,h)$, confidence level, and counters for track management. The state is defined in Equation~\eqref{obj-state}, where $(x,y,z)$ is the position, $(\dot{x},\dot{y})$ is the velocity within the ground plane. \change{The position and velocity are tracked in a static map frame external to the vehicle.} The confidence is obtained from the 2D detector and filtered temporally.

\vspace{-2mm}

A constant velocity motion model is used for all objects:
\begin{align}
    \mathbf{x} &= \begin{bmatrix} x & y & z & \dot{x} & \dot{y}\end{bmatrix}^T \label{obj-state}\\
    \mathbf{y} &= \begin{bmatrix} x & y & z \end{bmatrix}^T \\
    \mathbf{x}_k &= \mathbf{A}\mathbf{x}_{k-1} + \boldsymbol{\omega} \\
    \mathbf{y}_k &= \mathbf{C}\mathbf{x}_{k} + \boldsymbol{n} \\
    \boldsymbol{\omega} &\sim \mathcal{N}(\mathbf{0}, \mathbf{Q}) \\
    \boldsymbol{n} &\sim \mathcal{N}(\mathbf{0}, \mathbf{R})\\
    \mathbf{A} &=
        \begin{bmatrix}
            1 & 0 & 0 & T & 0 \\ 
            0 & 1 & 0 & 0 & T \\ 
            0 & 0 & 1 & 0 & 0 \\ 
            0 & 0 & 0 & 1 & 0 \\ 
            0 & 0 & 0 & 0 & 1
        \end{bmatrix} \\
    \mathbf{C} &= 
        \begin{bmatrix}
            1 & 0 & 0 & 0 & 0 \\ 
            0 & 1 & 0 & 0 & 0 \\ 
            0 & 0 & 1 & 0 & 0
        \end{bmatrix}
\end{align}
\vspace{-2mm}

{\setlength{\parindent}{0cm}
In this setup, $\mathbf{x}$ is the state of the object, $\mathbf{y}$ is the measurement, $\boldsymbol{\omega}$ is the system noise, $\boldsymbol{n}$ is the measurement noise, $\mathbf{A}$ is the state matrix, and $\mathbf{C}$ is the observation matrix. $\mathbf{Q}$ and $\mathbf{R}$ are the system noise and measurement noise covariances, which we assume to be diagonal. \change{The diagonal entries of $\mathbf{Q}$ and $\mathbf{R}$ are used to tune the tracker for responsiveness vs. smoothness.  The equations that describe the linear Kalman filter used here can be found in Section 3.3 in \citep{barfoot2017state} The remaining 1D variables including the width, length, height, and confidence are filtered using an Alpha-Beta filter for each.} 
}

\vspace{-2mm}
\begin{table}[h]
    \centering
    \caption{Runtime of each component in aUToTrack. The total runtime is sum of one instance of SqueezeDet running on 16 threads followed by the clustering and tracking run sequentially on one thread each. The runtime of SqueezeDet on a 1080Ti GPU and Arria 10 FPGA are provided for comparison.}
    \label{runtime}
    \begin{tabular}{|l||l|l|}
    \hline
    \textbf{Component}       & \textbf{Run Time} & \textbf{Hardware}              \\ \hline \hline
    SqueezeDet (Pedestrian Detection) & 18 ms              & NVIDIA GTX1080Ti GPU (For Comparison Only)                \\ \hline
    SqueezeDet (Pedestrian Detection) & 26 ms              & Arria 10 FPGA (For Comparison Only)                \\ \hline
    SqueezeDet (Pedestrian Detection) & \textbf{32 ms}             & Intel Xeon E5-2699R (16 threads) \\ \hline
    Clustering               & \textbf{15 ms}             & Intel Xeon E5-2699R (1 thread)   \\ \hline
    Tracker                  & \textbf{$<$1 ms}              & Intel Xeon E5-2699R (1 thread)   \\ \hline
    \textbf{Total Run Time:} & \textbf{47 ms}             & Intel Xeon E5-2699R Only       \\ \hline
    \end{tabular}
    \label{table:runtime}
\end{table}
\vspace{-2mm}
\subsection{Data Association and Track Management}
Our data association is based on \change{metric information} only. This works well for 3D \change{objects} such as \change{pedestrians and} cars \change{that} tend to be separated by significant distances. We use static gates to associate new detections to existing tracks. The gates are \change{designed} using the maximum possible inter-frame motion. Assuming a maximum speed of 5 m/s for pedestrians and a 0.1 s time step, we have a gating region of 0.5 m. 
\vspace{-2mm}

We used a greedy approach to associate measurements \change{with} tracked objects. For each tracked object, we evaluated the distance between the object \change{and} observations within its gate to find the nearest neighbor. The nearest neighbor is then assigned to the tracked object, \change{and} removed from the list.

\vspace{-2mm}

We employed a strategy of greedy track creation and lazy deletion for managing tracks. In greedy track creation, every observation becomes a new track. However, every track must go through a \change{trial period}.  While objects are in their trial period, they are removed from the list of objects being tracked if they miss a single frame. Once objects are promoted from their trial period, we count the number of consecutive frames that an object has been unobserved for. In order for a non-trial track to be removed from the list, there must be no associated measurements for \change{several} consecutive frames. \change{Due to the simplicity of the algorithms we employed, our pipeline runs exceptionally fast, as shown in Table~\ref{runtime}.}

\vspace{-2mm}

\begin{table}[t]
    \centering
    \caption{Position and Velocity Estimation Error vs. Target Distance}
    \label{distance-table}
    \begin{tabular}{|l||l|l|l|l|l|l|l|}
    \hline
    \textbf{\begin{tabular}[c]{@{}l@{}}Target \\ Distance(m)\end{tabular}} & \textbf{5}    & \textbf{10}   & \textbf{15}   & \textbf{20}   & \textbf{25}   & \textbf{30}   & \textbf{35}    \\ \hline \hline
    \textbf{\begin{tabular}[c]{@{}l@{}}Position \\ RMSE (m)\end{tabular}}   & 0.14 & 0.18 & 0.21 & 0.26 & 0.22 & 0.27 & 0.37 \\ \hline
    \textbf{\begin{tabular}[c]{@{}l@{}}Velocity \\ RMSE (m/s)\end{tabular}} & 0.20 & 0.19 & 0.18 & 0.23 & 0.32 & 0.29 & 0.55 \\ \hline
    \end{tabular}
\end{table}

\vspace{-3mm}

\subsection{Performance}

\vspace{-2mm}

In order to assess the performance of our approach prior to the competition, a dataset was collected with GPS-based ground truth for pedestrian motion. \change{This} new dataset, named UofTPed50, is \change{used} for benchmarking 3D object detection and tracking of a \change{single} pedestrian.  UofTPed50 consists of 50 sequences of varying distance, trajectory shape, pedestrian appearance, and \change{ego-vehicle} velocities. \change{Position and velocity information are reported in a static global reference frame. Hence, a stationary pedestrian corresponds to a velocity of 0 m/s.}

\vspace{-2mm}

In several sequences in UofTPed50, the pedestrian walks laterally from one side of the vehicle to the other at evenly spaced distances. Using these sequences, one can measure the impact of varying target distance on performance. We use Root Mean Squared Error (RMSE) as our error metric for both position and velocity estimation. As summarized in Table~\ref{distance-table}, our position and velocity estimation error tends to increase with distance. We achieve consistent velocity estimation accuracy up to 30 m\change{, but} the performance drops off around 35 m. This is potentially due to the \change{impact of a poor calibration} further from the LIDAR.

\vspace{-2mm}

\begin{figure*} [b]
    \centering
    \subfigure[Position estimation]{\includegraphics[width=0.47\textwidth]{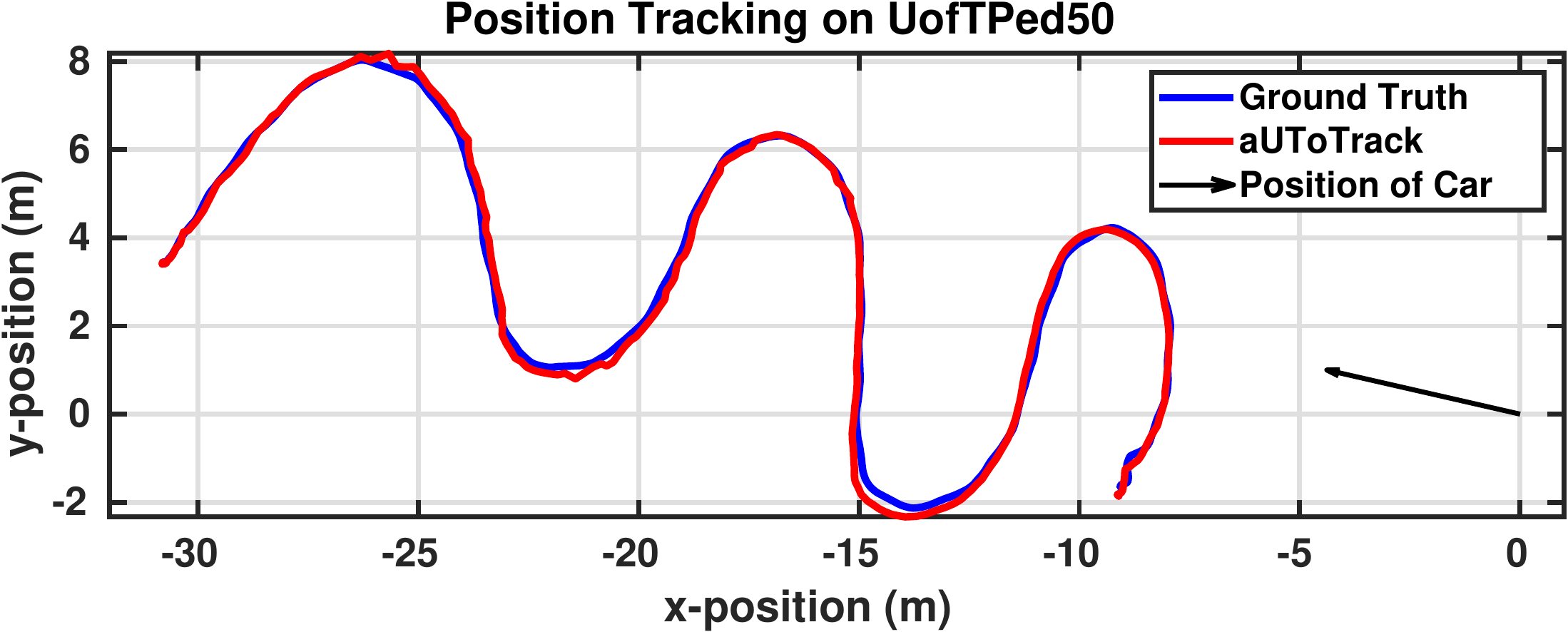}}
    \subfigure[Velocity estimation]{\includegraphics[width=0.47\textwidth]{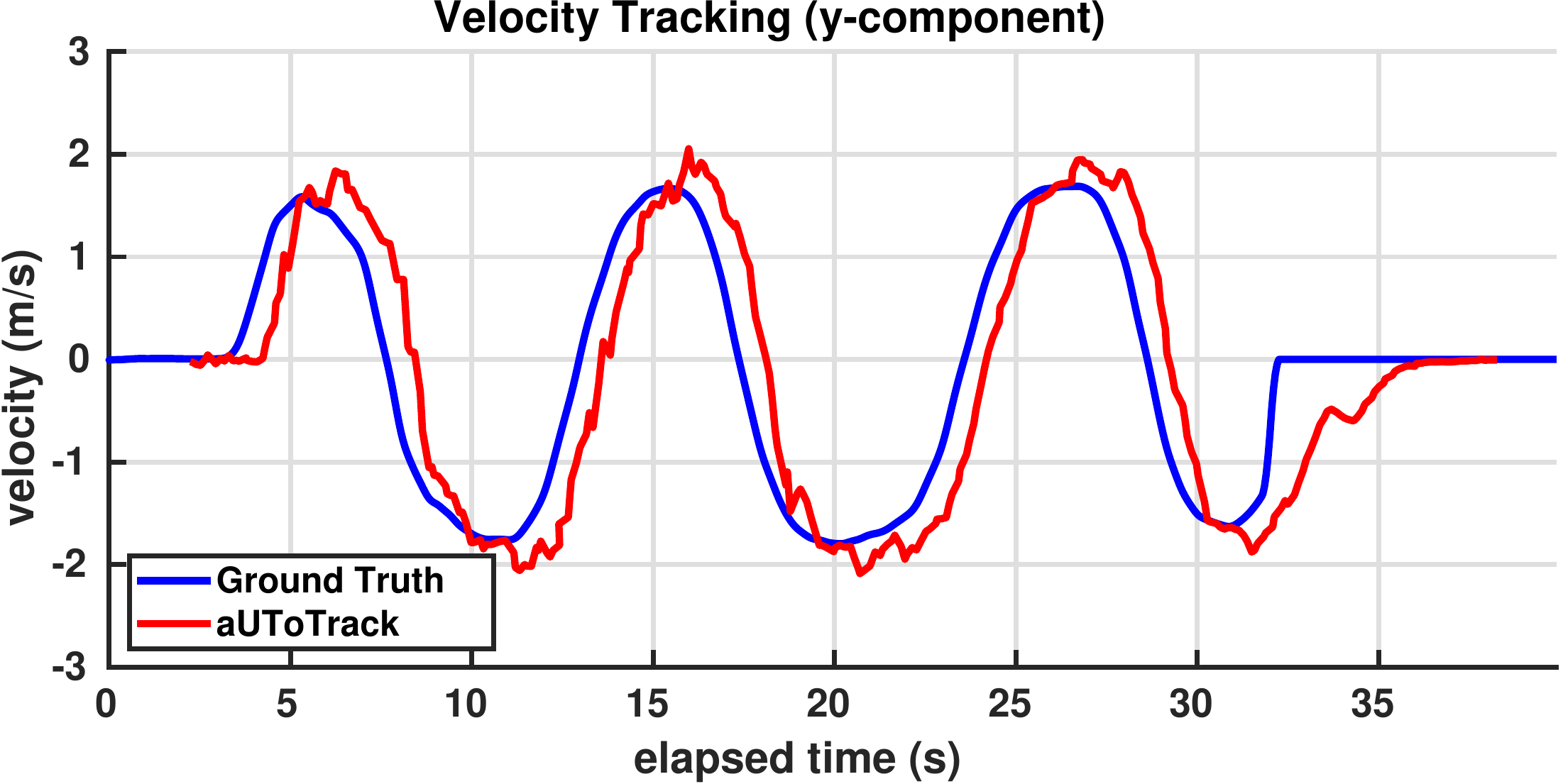}}
    \caption{Position and velocity estimation of the Zig-Zag scenario.}
    \label{fig:zigzag}
\end{figure*}

\change{Figure~\ref{fig:zigzag} demonstrates a challenging Zig-Zag trajectory where the heading and velocity of the pedestrian changes quickly over time.} aUToTrack is \change{able to track} the reference trajectory with reasonable accuracy. However, the position and velocity estimation appear to overshoot and lag behind the ground truth. This is likely due to estimator dynamics and could be addressed with parameter tuning.

\vspace{-2mm}

\change{It should be noted that the centroid estimation technique we use is not as accurate for cars.} This is likely because the majority of the points being clustered are from the \change{side} of the object facing the LIDAR. For this reason, our centroid estimation may be off by \change{up to} 1 m for cars.  For safe autonomous driving, a more accurate centroid estimate is likely required. \change{Candidate approaches for 3D detection of vehicles include recent LIDAR-only approaches such as PointPillars \citep{pointpillars} and PIXOR \citep{yang2018pixor}.} Nevertheless, our approach is still accurate \change{enough} for the AutoDrive Challenge.

\vspace{-2mm}

\vspace{-2mm}

\section{Traffic Light and Sign Detection} \label{lnssection}

\vspace{-1mm}

\begin{figure} [t]
    \includegraphics[width=0.7\columnwidth]{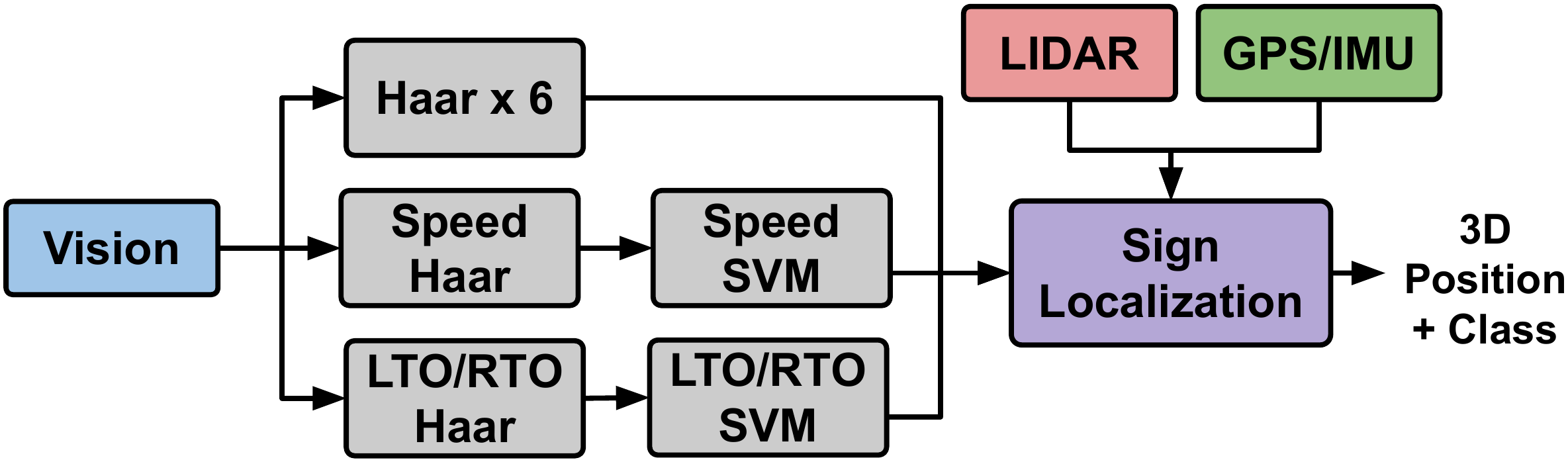}   
    \centering
    \caption{This figure depicts our traffic sign detection \change{pipeline}. A generic Haar cascade detector is used to identify speed limit signs and Left-Turn-Only vs. Right-Turn-Only signs. These detections are then classified by \change{a} support vector machine for each \change{set}. The other six sign classes are detected and classified by a single Haar cascade for each \change{sign}.}
    \label{fig:signarch}
\end{figure}

\vspace{-1mm}

SqueezeDet was used to detect and classify the state of traffic lights. In order to achieve good performance, SqueezeDet was trained on a custom dataset of over 17000 images. Replicas of the competition traffic lights were used to optimize our detector for MCity. \change{Only two traffic light states were trained on: red and green.}

\vspace{-1mm}

SqueezeDet is lightweight, but lacks the expressive power to accurately detect and classify 13 different traffic signs. For this reason, we opted to use Haar cascades instead \citep{ViolaJones2001}. Traffic signs are generally easier to detect than pedestrians as their appearance remains consistent across viewpoint and illumination. For this reason, simpler detectors that take advantage of template matching work well.

\vspace{-1mm}

In some cases, Haar cascades struggle to differentiate between signs with similar appearances. This included Left-Turn-Only (LTO) and Right-Turn-Only (RTO) text signs, and speed limit signs. To resolve this issue, sign detection was divided into two stages: an initial generic sign detector for LTO/RTO and speed \change{limit} signs followed by an SVM for classification. Figure~\ref{fig:signarch} depicts this architecture. This method allowed us to exceed 90\% precision and recall on our validation set. When tracking and pruning heuristics are included, performance improves 5-10\%.

\vspace{-1mm}

Once 2D bounding boxes for traffic lights and signs are obtained, they must be localized with respect to the vehicle so that higher-level decisions can be made by the planner. To localize signs, we use a pipeline that is very similar to our pedestrian detection. We project LIDAR points onto the image plane and extract the points that correspond to the sign's bounding box. We then run Euclidean clustering and return the cluster \change{that is} closest to the vehicle. By comparing the expected bounding box size with the detected bounding box, false positives can be pruned out. To smooth sign classes temporally, we treat signs as generic \change{sign} objects, and keep track of the previous 10 classes associated with that sign. When publishing the detected signs, we return the most common class observed within the last 10 detections of that sign. Temporal filtering is essential for smooth and reliable performance.

\vspace{-1mm}

\begin{figure} [t]
    \includegraphics[width=0.5\columnwidth]{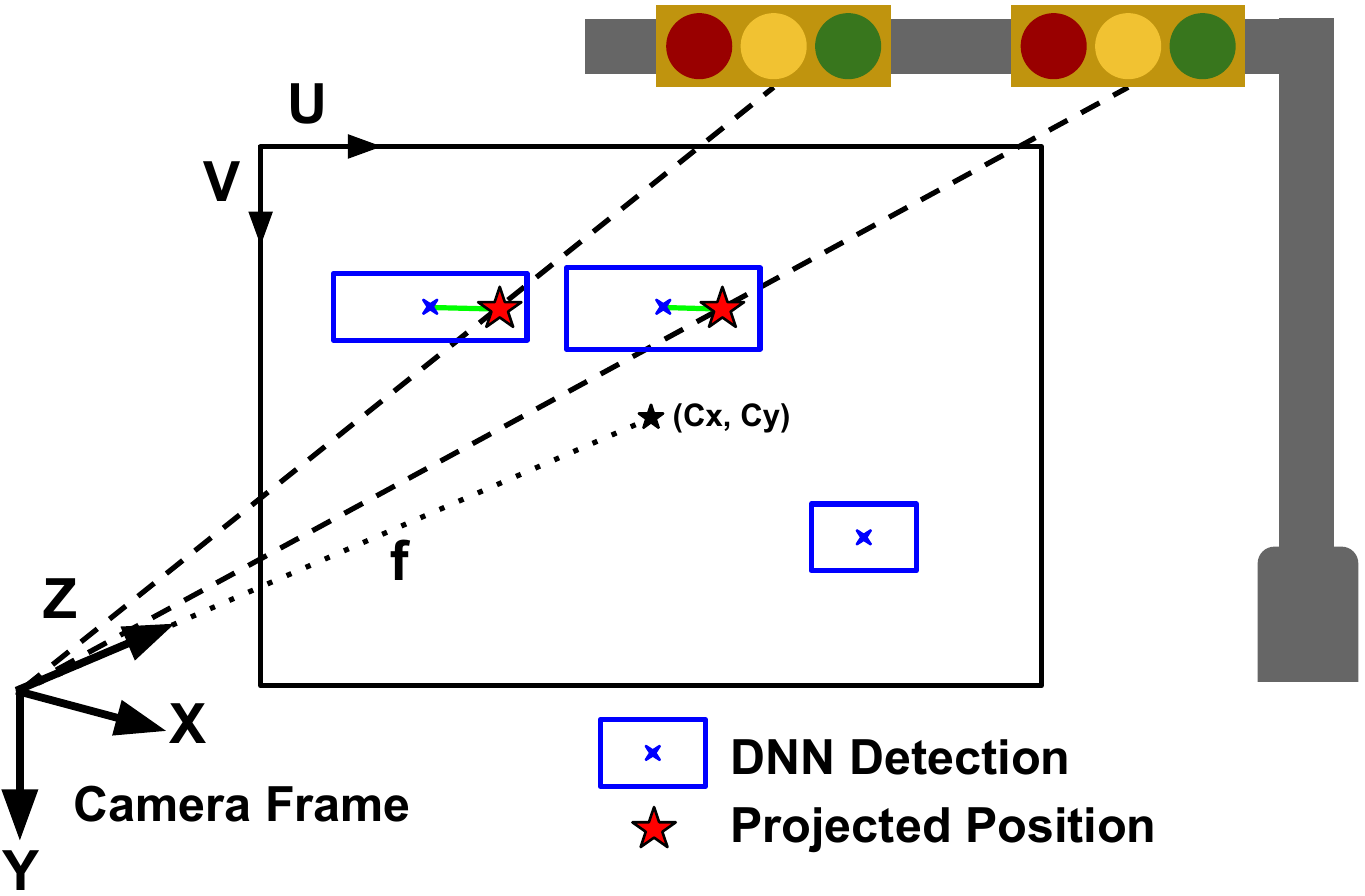}   
    \centering
    \caption{\change{In this figure there are two upcoming traffic lights. Using our position in the semantic map, the locations of these traffic lights are projected onto the image plane (denoted as the red stars). The DNN has output two true positive detections and one false positive in this frame (denoted as blue boxes). The green lines denote the association between detections and expected positions. }}
    \label{fig:lightproject}
\end{figure}

Traffic lights were included in the semantic map used at the competition. As the vehicle approaches an intersection, the expected \change{positions} of traffic lights in 3D space are projected onto the image plane. Figure~\ref{fig:lightproject} depicts this process. By associating raw detections with the expected positions, the exact position of traffic lights and their associated lane can be obtained easily from the map. To make these associations, we minimize a cost function based on Euclidean distance in the image plane and difference in heights. We also apply a maximum association distance of 3 m within the image plane. The corresponding pixel value will change depending on the expected distance to the upcoming lights.

\vspace{-1mm}

We assume that all traffic lights are red by default. If there are several relevant traffic lights at an upcoming intersection, they must all be detected as green before the vehicle will proceed. We also keep track of the 10 previous detections associated with each expected traffic light to smooth detections temporally. This is a very cautious approach, but ensured that \change{Zeus stopped} at each red light at the competition. 

\vspace{-1mm}

A timer was \change{included} to prevent the vehicle from waiting at a set of traffic lights forever. In our case, the vehicle was set to wait for a maximum of 60 seconds. This timeout prevented the vehicle from becoming stuck during the Intersection Challenge. Heuristics like this \change{were} key to \change{winning the} competition.

\vspace{-1mm}

Flashing red lights were a challenging aspect of the Year 2 competition. Initially, our approach was to keep track of the previous 20 detections, and to simply calculate the duty cycle of red vs. off to determine if the lights were flashing. In our case, if the duty cycle of red was between 35\% and 65\%, then a light was considered flashing. \change{Since our traffic light detector was only trained to recognize red and green traffic lights, the class output for off traffic lights was close to random. To overcome this problem, we designed a hand-crafted computer vision algorithm based on OpenCV libraries to determine whether a traffic light was truly off. This hand-crafted method underperformed in varying illumination conditions and ultimately was not enabled during the competition. The lesson learned here is that 'off' traffic lights should be an additional traffic light state trained on in order to achieve good flashing light performance.} At the competition, the Pedestrian Challenge was the only challenge to include flashing red lights. As such, we were able to simply treat flashing red lights as solid red lights with a 5 second timeout.

\vspace{-1mm}

\begin{figure}[h]
    \center{\includegraphics[width=0.55\columnwidth]{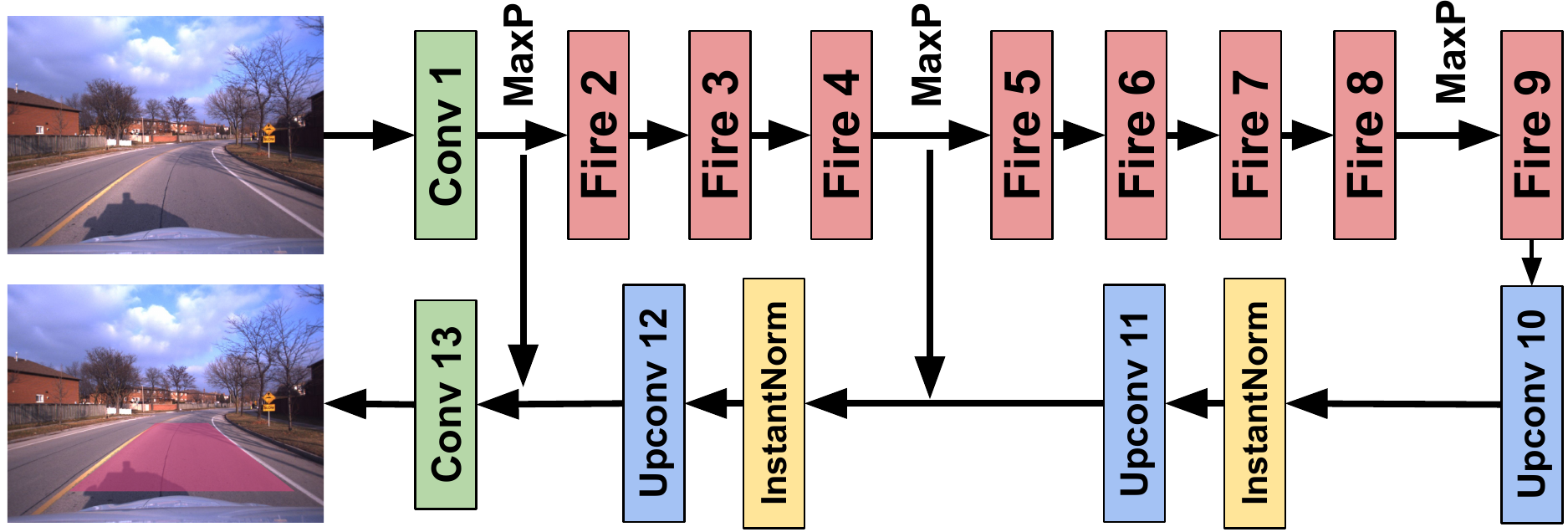}}
    \caption{\change{Our proposed semantic segmentation architecture.}}
    \label{fig:lane_dect1}
\end{figure}

\vspace{-2mm}

\section{Lane Detection} \label{lanesection}

\vspace{-2mm}

Zeus did not use lane detection at the Year 2 competition. However, due to the importance of this component in Year 1 and it's potential usage in future years of the competition, it continues to be an active area of research for aUToronto.

\vspace{-2mm}

In the first year of the competition, no map was \change{given;} as \change{such,} lane detection was the sole source of localization information. This system needed to be reliable and robust to changing illumination conditions. aUToronto developed three different lane detection approaches that each ran in \change{realtime}. These included steerable filters, a convolutional neural network, and a LIDAR-based approach \citep{burnett2018building}.

\vspace{-2mm}

\vspace{-2mm}

The Year 1 system was optimized for tight corners and lanes with quickly varying curvature. At this time, lane lines were always present where the vehicle was expected to drive. At the Year 2 competition, that system was no longer appropriate. Lanes can be faded in some locations and may not be present on some roads. Further, bad weather and illumination can make it difficult to detect lane lines consistently.

\vspace{-2mm}

For Year 2, we originally planned to use GPS/IMU positioning and to use lane detection for lateral corrections. However, to use lane detection this way, it must be shown to be as reliable as the GPS/IMU. Otherwise, these corrections could add error to the position estimates. \change{An altnerative way to use lane detection is to compare the output against the expected lane positions reported by the semantic map. This approach allows for the bias between the GPS coordinate system and the map coordinate system to be calibrated. In this case, lane detection does not need to be constantly running with a high reliability, but rather only needs to run occasionally with a few discrete measurements to obtain this bias correction. This means that the lane detection module can use a large, accurate, and potentially slow DNN for semantic segmentation. This project was moved to future work for the Year 3 competition and beyond.}

\vspace{-2mm}

\change{In an attempt to achieve} real-time performance on CPUs, we designed a lightweight segmentation network based on Squeeze-SegNet \citep{lane2}. Our modified structure is shown in Figure~\ref{fig:lane_dect1}. The contracting layers in the network are initialized using SqueezeNet weights pre-trained on ImageNet. Due to its limited number of layers, SqueezeNet has a very small receptive field, making it difficult to distinguish the ego-lane from adjacent lanes. To combat this, some convolution layers have been replaced with dilated convolution layers. Further, all batch normalization layers were replaced with instance normalization layers to improve the robustness to intensity change and color shift.
\vspace{-2mm}

We trained our model on the large BDD100k dataset and tested on BDD's evaluation set \citep{lane5}. We achieve an accuracy of 0.95 and an mIOU of 0.79 on BDD100k whereas our baseline Squeeze-SegNet \change{resulted} in an accuracy of 0.83 and an mIOU of 0.69. Furthermore, our approach is capable of running at up to 10 FPS on CPUs. Evaluation was conducted at 256 x 256 image resolution, which we found to be a suitable balance between accuracy and performance.

\vspace{-2mm}

\begin{figure}[hb!]
    \center{\includegraphics[width=0.45\columnwidth]{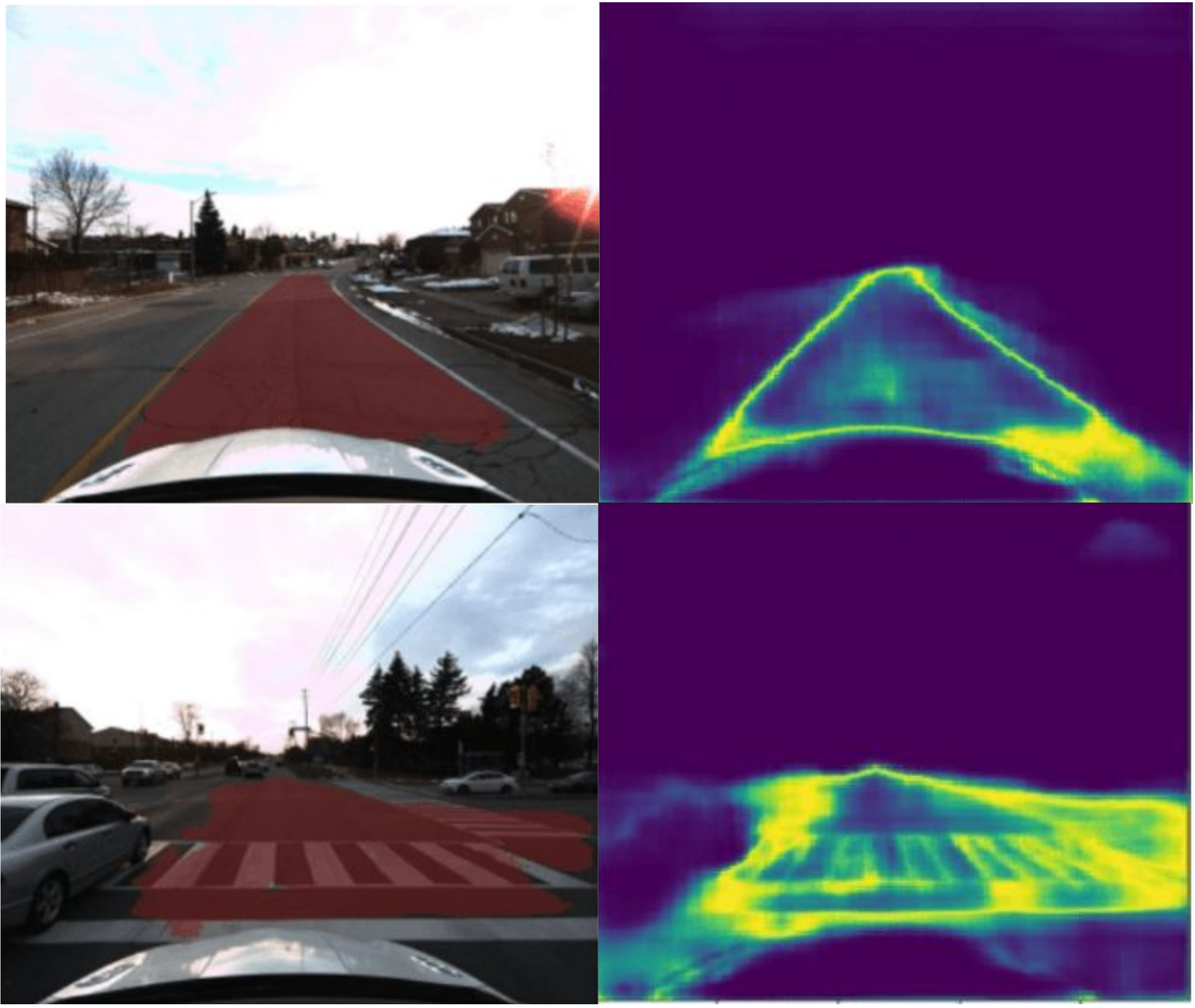}}
    \caption{Top: the driving lane is clearly visible. The associated entropy is low. Bottom: the vehicle approaches an intersection. The entropy is high, indicating the lane detection output should be ignored.}
    \label{fig:lane_uncertainty}
\end{figure}

In many scenarios, the ego-lane is ill-defined such as at an intersection. These scenarios present difficulties for deep learning systems. Thus, we experimented with extending our model with an error prediction component based on Bayesian Deep Learning. Our model uses Monte Carlo dropout to calculate entropy as is done in \cite{lane4}. Using this approach, entropy is high when the input sample is significantly different from the training samples. In these cases, the ego-lane tends to be ill-defined. Hence, a threshold on entropy can \change{be} used to identify where lane detection should be ignored, as shown in Figure~\ref{fig:lane_uncertainty}.

\vspace{-2mm}

\vspace{-2mm}

\begin{figure}[ht]
  \centering
  \subfigure[Post-processed LIDAR map of UTIAS by Applanix]{\includegraphics[width=0.45\columnwidth]{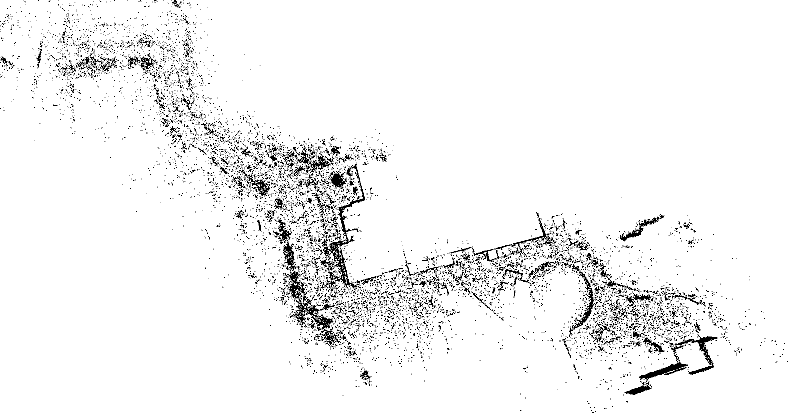}} 
  \subfigure[Trajectory output by LIDAR localization]{\includegraphics[width=0.45\columnwidth]{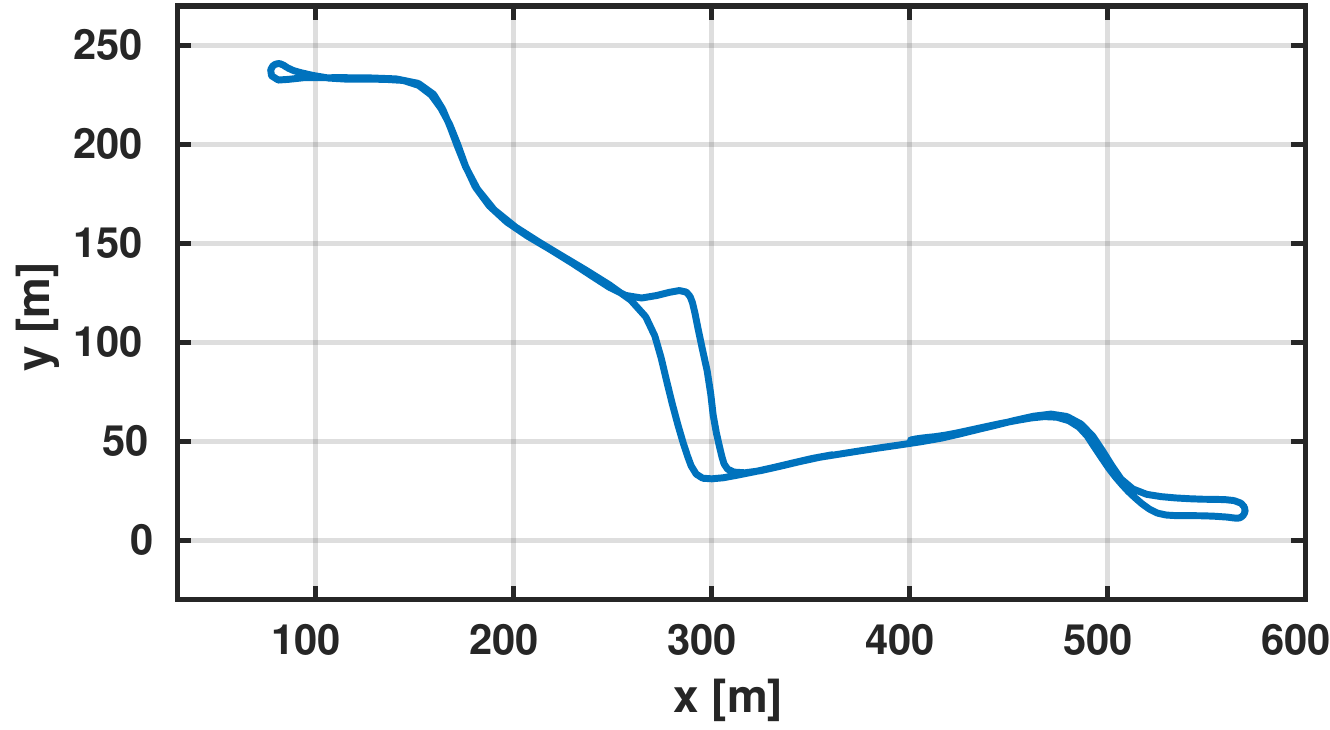}}
  \subfigure[Satellite Image of UTIAS (Google Earth)]{\includegraphics[width=0.45\columnwidth]{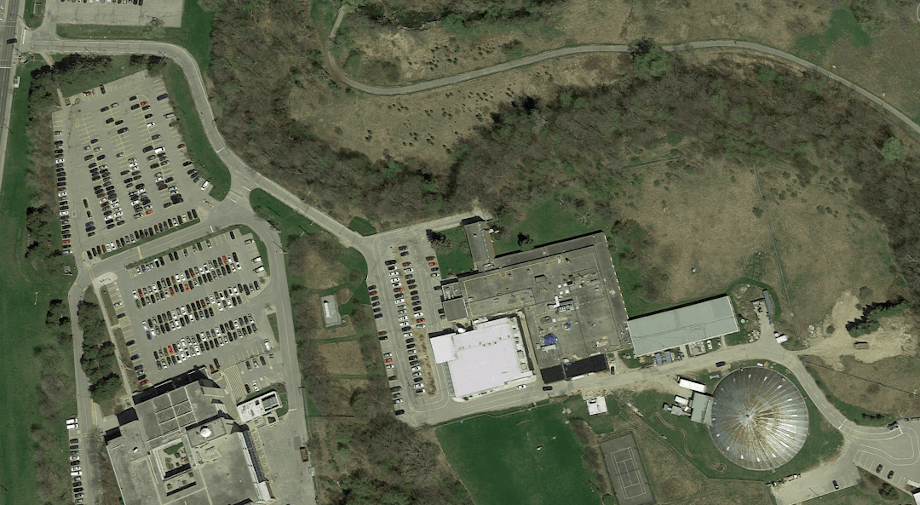}}
  \caption{(a) \change{key-frame-based} LIDAR map built using Applanix software. (b) trajectory estimated using only LIDAR localization. (c) Satellite image of UTIAS.}
  \label{fig:map_and_traj}
  \vspace{-3mm}
\end{figure}

\section{Mapping}

There are two types of maps we are concerned with: semantic maps and LIDAR maps. Semantic maps encode the location of lanes, traffic lights, and more. LIDAR maps may consist of a single aligned pointcloud for an entire area or a set of pointclouds with GPS coordinates. \change{An example of a LIDAR map is shown in Figure~\ref{fig:map_and_traj} alongside the resulting trajectory estimate and Google Earth image for reference}.

\vspace{-2mm}

Semantic maps can be used to off-load a significant fraction of active perception to an initial map-building phase that can be done offline. This reduces the problem of self-driving from needing to perceive all features in \change{realtime} to only needing to perceive features that can't be encoded in a static \change{map,} such as the location of other traffic participants and the state of traffic lights.

\vspace{-2mm}

In Year 2 of the competition, we made the assumption that our semantic map contained no major mistakes. Thus, we simply needed to localize ourselves within that map to take advantage of all the information it provided. In cases where the operational area is small, or a repeated route is used, this assumption generally holds. However, if the map that is being used encompasses an entire city, it becomes critical to relax this assumption and actively look for inconsistencies in the map such as a new construction zone. 

\vspace{-2mm}

Semantic maps are often represented using geometry such as points, lines, and polygons. The data formats provided to our team were not immediately conducive to running planning algorithms such as A*. It was helpful to convert these maps into a graph-based format to simplify the planning software. For this purpose, we developed a pipeline which allowed us to convert semantic maps into our own internal format which is based on \citep{OpenStreetMap}.

\vspace{-2mm}

\change{For the competition, we chose Carmera to be our map supplier}. They report a low relative error between points in their map. Further, they supplied us with a LIDAR map of MCity which enabled \change{us to use} LIDAR localization.

\begin{figure}[t]
  \centering
  \includegraphics[width=0.7\columnwidth]{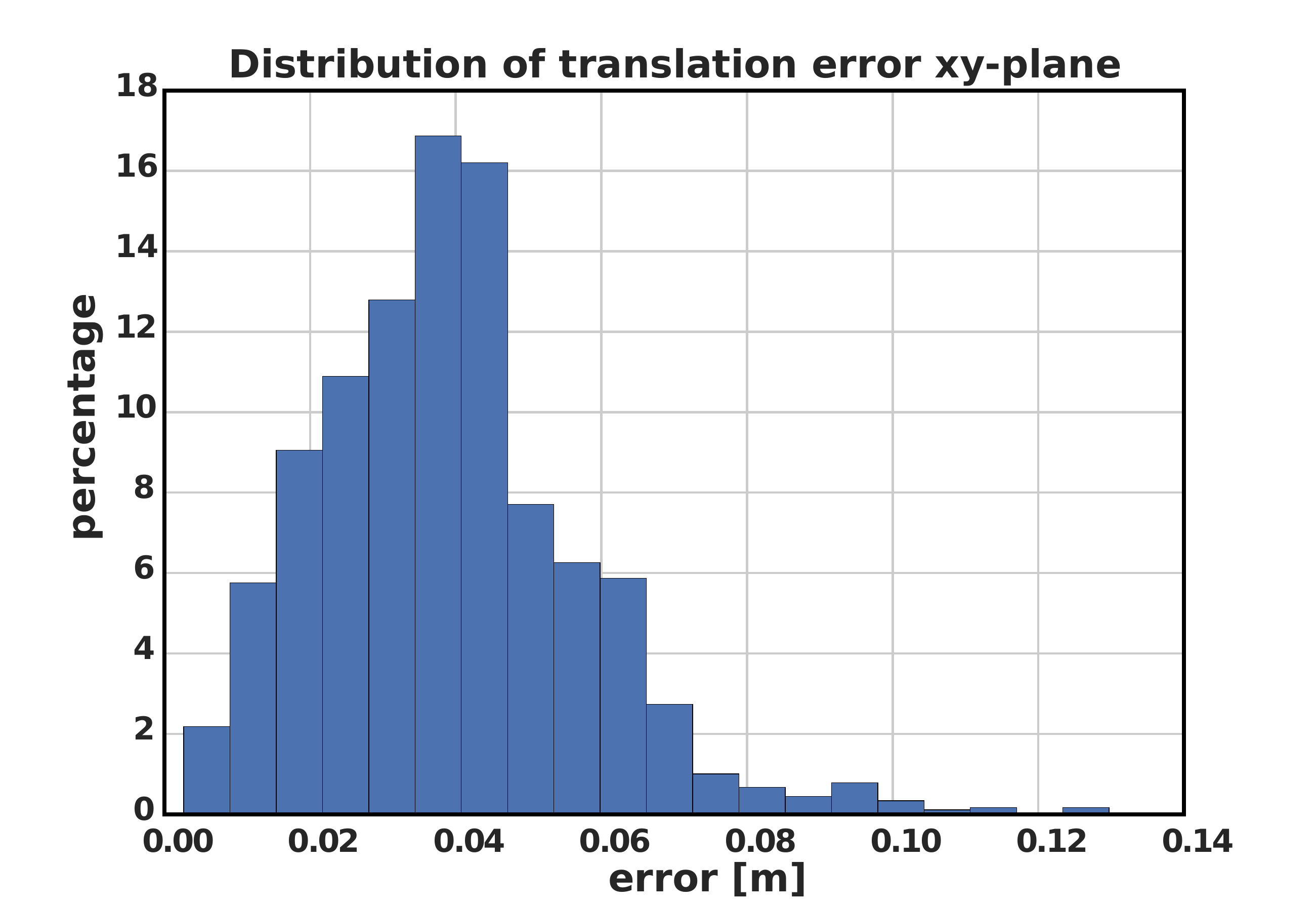}
  \caption{Position error histogram ($xy$-plane) of LIDAR-only localization along a trajectory (see Figure~\ref{fig:map_and_traj}) compared to ground truth obtained from post-processed Applanix POSLV GPS/IMU output.}
  \label{fig:lidar_loc_hist}
  \vspace{-3mm}
\end{figure}

\section{Localization} \label{locsection}

As mentioned in Section~\ref{lanesection}, lane detection was not used at the Year 2 competition. This was mainly due to concerns about reliability and the fact that lanes were not present everywhere at the Year 2 competition.

\vspace{-2mm}

A key design problem in Year 2 required dealing with a potentially GPS-denied tunnel. Further, over the course of a year of testing we experienced several position jumps greater than 50 cm. In order to meet competition requirements, these problems needed to be addressed.

\vspace{-2mm}

The two options we compared for localization at MCity were: LIDAR localization provided by Applanix, and GPS/IMU localization provided by Novatel (with a Terrastar satellite subscription).

\vspace{-2mm}

One drawback to GNSS systems is that they require a clear view of the sky. During milestone testing in a heavily wooded area in Mississauga, our precision frequently dropped and took an excessive period of time to converge.

\vspace{-2mm}

On \change{the} other hand, LIDAR localization presents numerous advantages. LIDAR is robust to ambient light change. In addition, Applanix’s LIDAR localization is robust to minor changes in scene geometry such as moving vehicles. As long as large structures like buildings remain fixed, their LIDAR localization will work. Since the sensor data \change{are} relative to the vehicle, it does not suffer the same shortcomings as GNSS.

\vspace{-2mm}

Figure~\ref{fig:lidar_loc_hist} shows an error histogram plot for LIDAR localization. In this case, the ground truth is Applanix POSLV data post-processed with their POSPac suite. It should be noted that post-processing mitigates the shortcomings of GNSS by performing a batch optimization over the entire trajectory after a data-taking run has ended. Since the batch optimization incorporates both future and past data, it cannot be used in \change{realtime}. In this experiment, the majority of errors are under 10 cm and there were no errors over 14 cm. This is comparable to the accuracy reported by the Novatel GPS/IMU. The resulting LIDAR-based localization is not susceptible to GPS dropout.

\begin{figure*}[ht]
\centering
\includegraphics[width=0.85\textwidth]{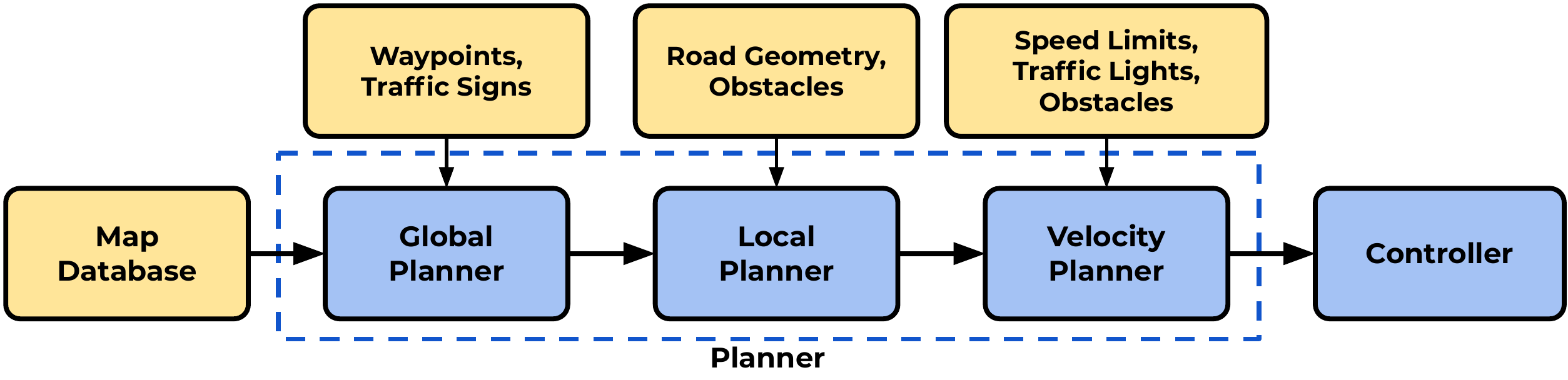}
\caption{The hierarchical structure of our motion planner. The Global Planner outputs a high-level route, the Local Planner produces a path, and the Velocity Planner generates a velocity profile.}
\label{fig:plannerhierarchical}
\end{figure*}

\section{Planning} \label{planningsection}

\vspace{-2mm}

The Year 2 challenges included point-to-point navigation in an urban environment and complex driving scenarios. \change{The} planner \change{needed to} respond to external stimuli \change{such as} traffic lights, traffic signs and pedestrians. At the same time, the planner \change{needed to} generate maneuvers that are safe and comfortable in \change{realtime}. Our hierarchical planner consisted of a Global Planner, Local Planner and a Velocity Planner\change{, as is shown in Figure~\ref{fig:plannerhierarchical}}. Similar hierarchical designs were deployed by the top-ranked teams in the DARPA Urban Challenge \citep{bossurbanchallenge}. 

\vspace{-2mm}

\subsection{Global Planner}
\vspace{-2mm}
The Global Planner operates on a connectivity-graph and selects future destinations depending on the current challenge. We classify the challenges into two groups: \textit{Sign Following} and \textit{Intersection Traversal}. Since the two groups require different map traversal strategies, two different global planning algorithms are implemented. Nevertheless, both algorithms aim to extract a subset of connected road segments from the semantic map to form a high-level mission plan.

\vspace{-2mm}
\textit{Sign Following} requires the vehicle to follow the direction of traffic signs: Left-Turn-Only, Right-Turn-Only, and Do-Not-Enter. The Global Planner first retrieves the vehicle's current road segment and the desired action from any detected traffic signs. The planner then looks up the successors of the current road segment. If a successor's direction and location coincides with a detected sign, the successor will be added to the plan. Otherwise, a successor is chosen based on a predetermined priority where straight has the highest priority. This process is repeated until the number of road segments in the plan reaches a predetermined limit or a terminating node in the map is reached. 
\vspace{-2mm}

\change{The pull-in parking maneuver required its own specialized logic and maneuvers. The course was set up to end by driving through a region with diagonal parking spots on the right hand side. Some spots were occupied by vehicles and one spot had a handicap parking sign. The location and shape of the spots were contained in the semantic map. Perception nodes monitored the occupancy of each spot and the 3D location of handicap signs. Once an available spot was found, a pull-in parking maneuver was generated.}

\vspace{-2mm}

\begin{figure*}[ht]
  \centering
  \subfigure[Toy example of planner operation]{\includegraphics[width=0.49\columnwidth]{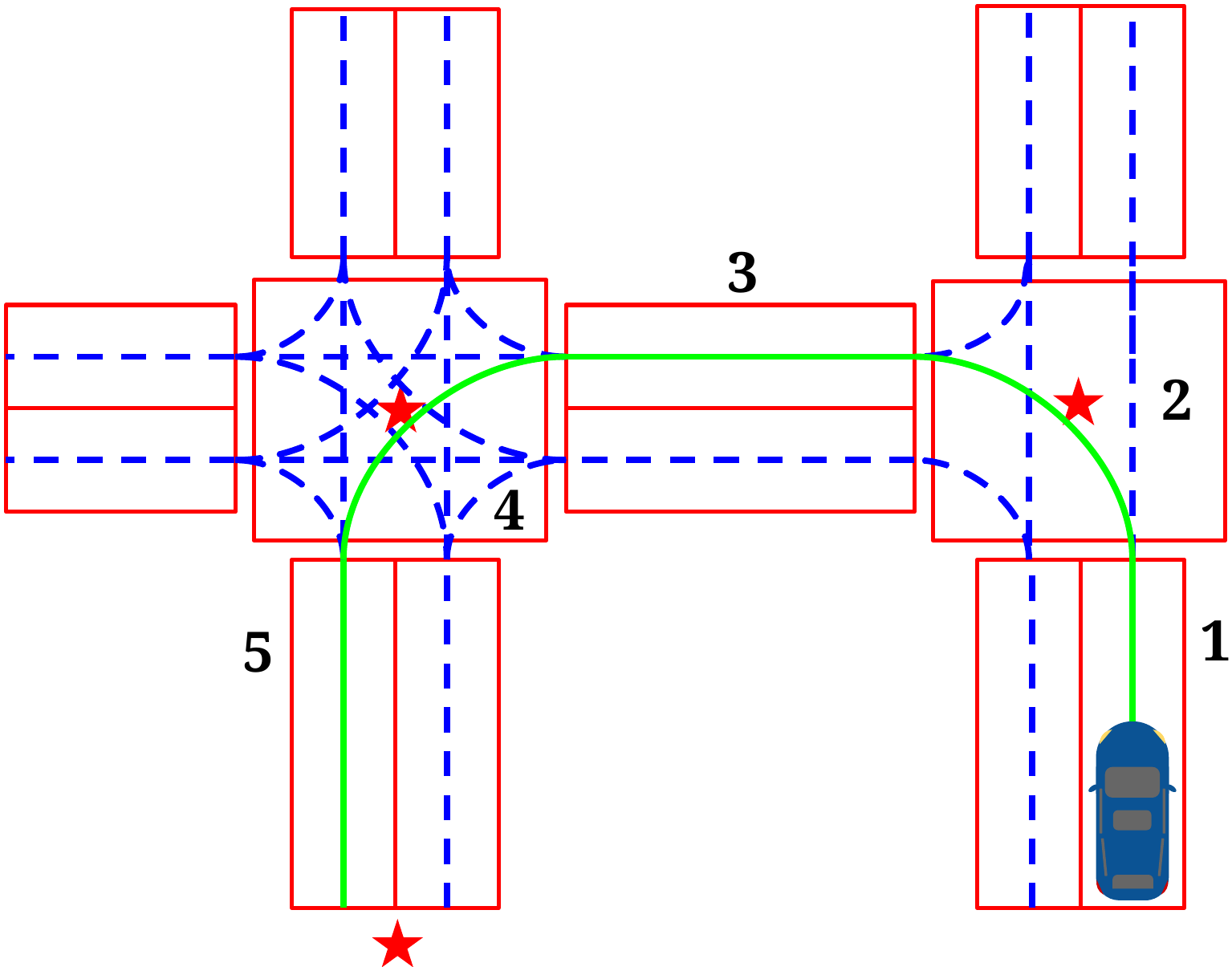}}
  \subfigure[Snapshot of the planner at MCity]{\includegraphics[width=0.49\columnwidth]{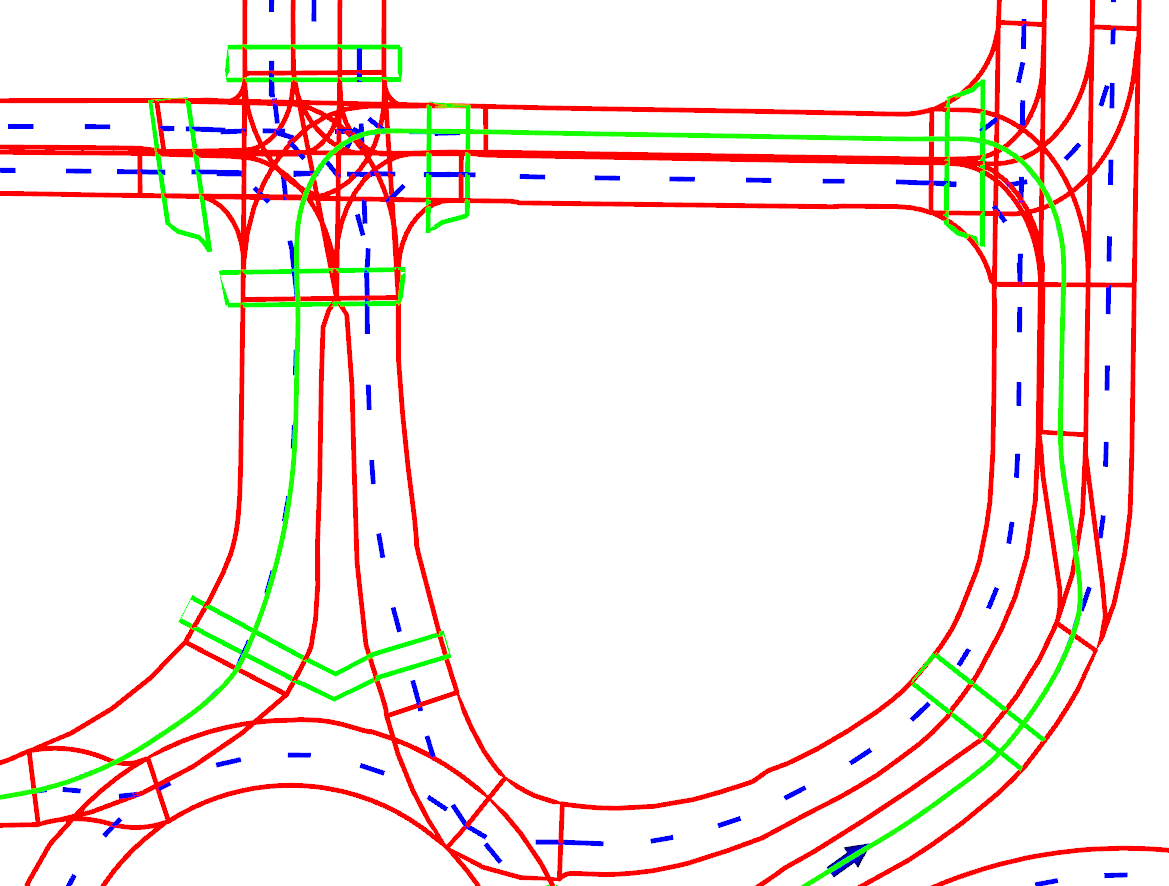}}
  \caption{\change{(a) Our semantic maps consist of road segments (red polygons) and centerlines (blue dashed lines). The red stars represent high-level waypoints centered at the intersections that need to be visited. The global planner starts at the current vehicle's position and analyzes all the road segments between successive pairs of waypoints. The global planner uses dynamic programing and A* to determine the best sequence of road segments to take in order to reach each waypoint. The local planner then stitches together the centerlines and turning arcs of those road segments and performs minor smoothing to output the desired path (given as the green line). (b) using the same color scheme, this figure depicts the actual path planned by Zeus during the first pedestrian challenge starting at the bottom right (blue arrow).}}
  \label{fig:globalexample}
\end{figure*}

\textit{Intersection Traversal} requires the vehicle to visit a list of waypoints at intersections. The planner aims to find an optimal (shortest) route that will guide the vehicle to visit all the waypoints in the correct order. To achieve an optimal path, we use a dynamic programming approach that utilizes A* as the backbone. The planner first finds the intersections enclosing the desired way-points. These intersections contain several road segments that enter and exit it. An algorithm based on A* that incorporates length, curvature, and lane change penalties is run between all exiting-entering road segment pairs between consecutive intersections. The length of each route is recorded to construct  a new graph. Value iteration is executed on the new graph to find the optimal route.  The Global Planner only needs to run once whenever the high-level mission is changed or the topological structure of the map is altered (e.g. a road is blocked). \change{The purpose and output of the global planner is explained further in Figure~\ref{fig:globalexample}.}
\vspace{-2mm}

\change{Stop signs and railroad crossing both require the vehicle to stop and wait for a fixed period of time before proceeding. A pseudo-obstacle is used to force the vehicle to stop at the desired location. Pedestrians waiting to cross or actively crossing require the vehicle to stop. If pedestrians are waiting at the curb, we place a pseudo-obstacle before the crosswalk with a timeout of 5 seconds. If the pedestrian begins crossing within this time, a permanent pseudo-obstacle is placed until the pedestrian has reached the other side.}

\vspace{-2mm}

\subsection{Local Planner}
The Local Planner receives route information from the Global Planner. To achieve a safe and comfortable path, the Local Planner aims to generate a collision-free and low-curvature path. We use an optimization algorithm based on \cite{stanfordjunior} that minimizes a nonlinear cost function $E$. The goal is to generate a path $\mathbf{P}$ = [$\mathbf{x}_1$ ... $\mathbf{x}_N$] that minimizes the cost function. Each pose in the path is defined as $\mathbf{x}_i$ = $\begin{bmatrix} x_i & y_i & \theta_i \end{bmatrix}^T$ The cost function, which has three terms for each pose, is defined below:

\vspace{-2mm}

\begin{equation}
E = \sum_{i=1}^{N} E_{i,cur} + E_{i,dev} + E_{i,obs}     
\end{equation}
\\
\vspace{-2mm}

The curvature term $E_{cur}$  minimizes abrupt steering actions by penalizing heading change at each index of the path. The curvature term is defined below:

\vspace{-2mm}
\begin{equation}
E_{i,cur} = w_{cur}\Big(\frac{\Delta \theta_i}{|\Delta \mathbf{x}_i|}\Big)  
\end{equation}

\vspace{-2mm}

{\setlength{\parindent}{0cm}
where $w_{cur}$ is the weight of the curvature term, $\Delta \theta_i = \cos^{-1} \frac{\mathbf{v}_{i-1,i} \cdot \mathbf{v}_{i,i+1}}{|\mathbf{v}_{i-1,i}||\mathbf{v}_{i,i+1}|}$ is the heading change at $\mathbf{x}_i$, and $\mathbf{v}_{i-1,i}=\mathbf{x}_i-\mathbf{x}_{i-1}$, $\mathbf{v}_{i,i+1}=\mathbf{x}_{i+1}-\mathbf{x}_{i}$ are the position changes between $\mathbf{x}_{i-1}, \mathbf{x}_{i}$ and $\mathbf{x}_{i+1}$. The deviation term $E_{dev}$ ensures the final path is close to the center of road by penalizing path deviation. The deviation term is defined below: }

\vspace{-2mm}

\begin{equation}
E_{i,dev} = w_{dev}|\mathbf{x}_{i}-\mathbf{c}_{i}|  
\end{equation}

\vspace{-2mm}

\begin{figure*}[t]
\centering
\includegraphics[width=0.85\textwidth]{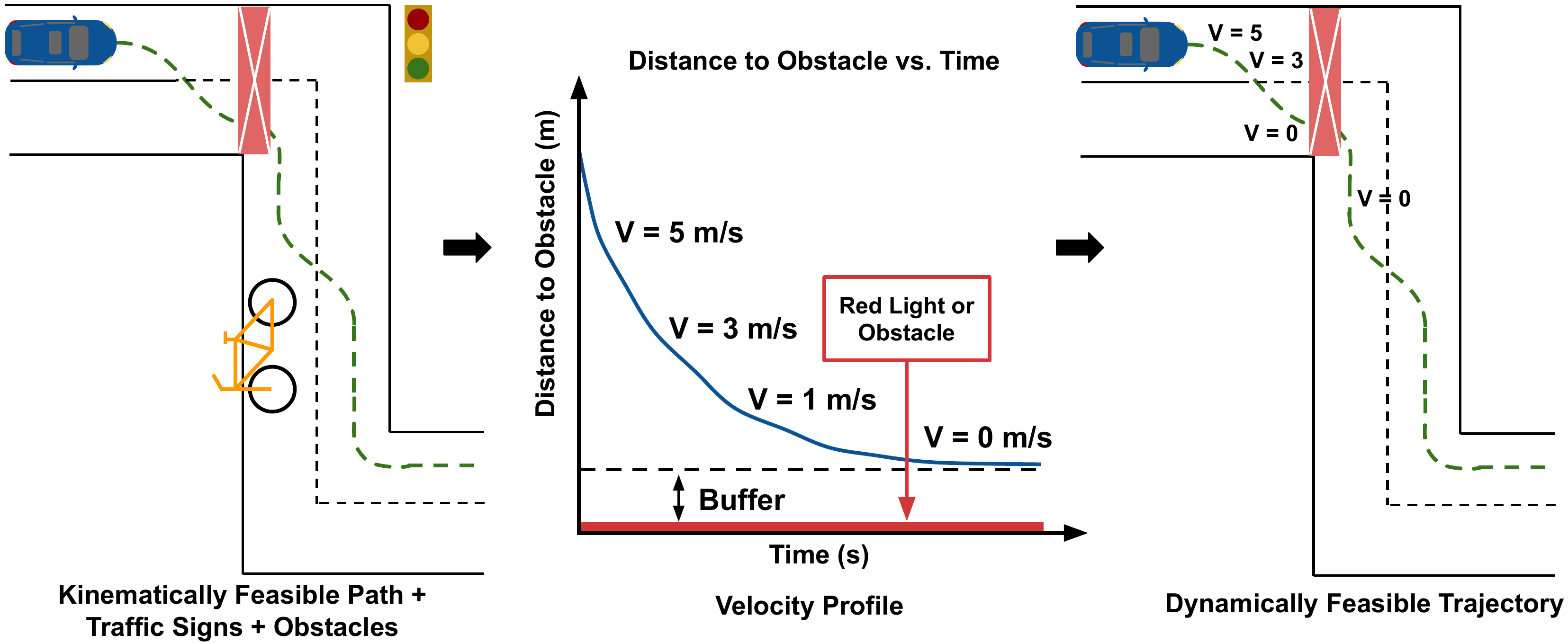}
\caption{The Velocity Planner uses control points to manage velocity profile along the path.}
\label{fig:vel_planner}
\end{figure*}

{\setlength{\parindent}{0cm}
where $w_{dev}$ is the weight of the deviation term and $\mathbf{c}_{i}$ is the closest point to $\mathbf{x}_{i}$ on the road center line. The obstacle term keeps the path far from obstacles by penalizing components of the path that are within some threshold $d$ from obstacles. The obstacle term is defined below:
}
\begin{equation}
    \begin{aligned}
        E_{i,obs} &= 0 \quad \text{for} \quad |\mathbf{x}_{i}-\mathbf{o}_{i}| \geq d \\
        E_{i,obs} &= w_{obs}\Big(\frac{d_{obs}}{|\mathbf{x}_{i}-\mathbf{o}_{i}|} -1 \Big)^{2} \quad \text{for} \quad |\mathbf{x}_{i}-\mathbf{o}_{i}| < d
    \end{aligned}
\end{equation}{}

{\setlength{\parindent}{0cm}
where $w_{obs}$ is the weight of the obstacle term and $\mathbf{o}_{i}$ is the \change{distance to the} closest obstacle to $\mathbf{x}_{i}$. Since we did not need to avoid obstacles in Year 2, this term is ignored for faster computation. Finally, the optimization problem can be solved iteratively using gradient descent. The initial guess $\mathbf{P}^{(0)}$ can be generated by connecting the center-lines of the road segments. The gradient descent formulation is given below: }

\begin{equation}
\mathbf{P}^{(t+1)} = \mathbf{P}^{(t)} - \eta \frac{\partial E(\mathbf{P}^{(t)})}{\partial \mathbf{P}^{(t)}}   
\end{equation}

The gradient is approximated using the equation below:

\begin{equation}
\frac{\partial E\left(\mathbf{P}^{(t)}\right)}{\partial \mathbf{x}_{i}^{(t)}} \approx \frac{E\left(\mathbf{P}^{(t)}+\Delta \mathbf{P}_{i}\right)-E\left(\mathbf{P}^{(t)}-\Delta \mathbf{P}_{i}\right)}{2 \Delta \mathbf{x}_{i}^{(t)}}   
\end{equation}{}

{\setlength{\parindent}{0cm}
where  $\Delta \mathbf{P}_i$ is a small disturbance to $\mathbf{x}_i$. The Local Planner only needs to run once when a new route is generated by the Global Planner or when an object detection is received.
}

\subsection{Velocity Planner}
The Velocity Planner assigns a speed to each pose in the path generated by \change{the} Local Planner. Efficient online velocity profiling is achieved by placing control points along the path and connecting them with constant-acceleration motion primitives. For example, when the vehicle needs to stop for a red light, the Velocity Planner places a 0-speed control point at the stop line. An illustration of this process is shown in Figure \ref{fig:vel_planner}.

\section{Control} \label{controlsection}

%

\begin{figure}[t]
\centering
\includegraphics[width=0.7\linewidth]{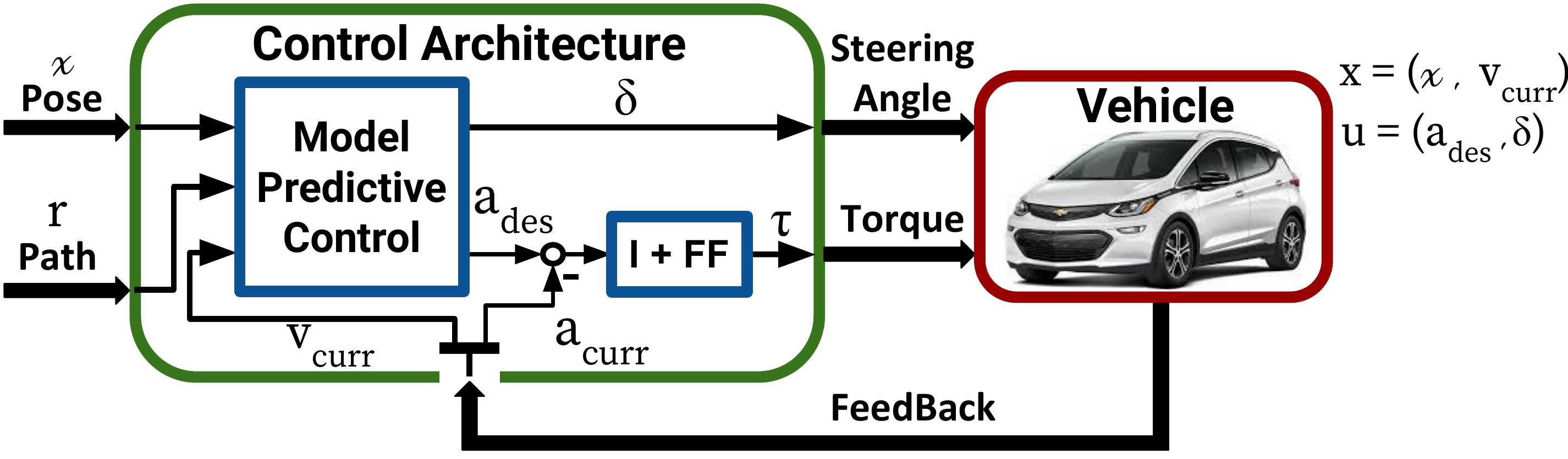}
\caption{Control architecture for Zeus. \change{x: current vehicle state. u: control input.}}
\label{fig:control-arch}
\end{figure}

The high-level architecture of our controller is shown in Figure~\ref{fig:control-arch}. The inputs to the controller are a desired path, velocity profile, and the current vehicle state. The controller then outputs steering wheel angle $\delta$ and torque $\tau$ for the vehicle. We designed a Nonlinear Model Predictive Controller (NMPC) for this purpose. To reject disturbances such as slopes, we added an integral controller on acceleration error in addition to a feed-forward term that directly maps the desired acceleration to torque. 

Compared with other control approaches, such as feedback linearization \citep{FBL}, MPC offers several advantages for autonomous driving. First, MPC only makes mild assumptions about the motion model which allows it to be generalized to more complex models required in some challenging driving scenarios \citep{Survey}. Second, by decoupling controller design from vehicle modelling, less effort is required to improve upon a design. Lastly, MPC makes it easier to incorporate motion constraints. This allows it to avoid the common pitfalls that impact other control approaches. 

MPC also presents a couple \change{of} drawbacks. Stability is not guaranteed unless stronger assumptions about the motion model are made. However, this turns out to be less of an issue in practice. MPC also presents a higher computational cost than other control approaches, but this was not an issue for our team due to our ample compute resources on-board.



\subsection{NMPC Formulation}
NMPC aims to track a timed reference signal $\mathbf{r}$ derived from the desired path given by the planner. NMPC predicts how vehicle states $\mathbf{x}$ will evolve over a finite time window of size $N$ given control commands $\mathbf{u}$, nonlinear motion model $\mathbf{F}$ and current state $\mathbf{x}_o$. Using the predicted states, NPMC evaluates the cost of the control effort $\mathbf{J}_u$ and the tracking errors resulting from the predicted states $\mathbf{J}_e$. It obtains the best control input sequence $\mathbf{u}^{*}$ by minimizing the cost function while respecting the vehicle state and control command constraints $\mathbf{H}$. Following the approach in \cite{Survey}, we formulate this problem as a nonlinear constrained optimization given as:
 
\begin{mini}
	{\mathbf{x}_n, \mathbf{u}_n}{\hspace{0ex} \sum_{n=k+1}^{k+N} \mathbf{J}_e(\mathbf{x}_n, \mathbf{r}_n) +  \sum_{n=k}^{k+N-1} \mathbf{J}_u(\mathbf{u}_n, \mathbf{u}_{n-1}) }
	{\label{eq:GenForm}}{}
	\addConstraint{\mathbf{x}_{n+1}}{= \mathbf{F}(\mathbf{x}_n, \mathbf{u}_n), \hspace{3ex} n = k, k+1 \dots k+N-1}
	\addConstraint{\; \mathbf{H}(\mathbf{x}}{_{k+1:k+N}, \mathbf{u}_{k:k+N-1})\leq 0}
	\addConstraint{\mathbf{x}_{k} = }{\mathbf{x}_o, \;\mathbf{u}_{k-1} = \mathbf{u}^{*}_{k-1}}
\end{mini}
where the scalar subscripts are time indices. In a fashion known as \textit{receding horizon}, only the first optimal command of the sequence (i.e. $\mathbf{u}^*_k$) is applied and the process is repeated at the next control time step.


A simple 2D kinematic bicycle model was used, whose state vector $\mathbf{x} := [x \; y  \; v \; \theta]^T$ consists of planar position, speed and heading. The control input vector $\mathbf{u}:=[a \; \delta]^T$ consists of acceleration and steering angle. The cost functions for tracking error and control effort are given below:
\begin{equation} \label{eq:CostFunctions}
\begin{split}
 \mathbf{J}_e(\mathbf{x}_n, \mathbf{r}_n) &= (\mathbf{r}_n-\mathbf{x}_n)^T \mathbf{Q}_n (\mathbf{r}_n -\mathbf{x}_n) \;,\\
\mathbf{J}_u(\mathbf{u}_n,\mathbf{u}_{n-1}) &=  (\mathbf{u}_n - \mathbf{u}_{n-1})^T \mathbf{S}_n (\mathbf{u}_n - \mathbf{u}_{n-1})\;
\end{split}
\end{equation}

{\setlength{\parindent}{0cm}
where tracking error is defined as the Euclidean distance between vehicle state $\mathbf{x}$ and its desired state $\mathbf{r}$. The rate of change of control inputs is interpreted as the control effort. $\mathbf{Q}_n$ and $\mathbf{S}_n$ are positive semi-definite diagonal matrices trading off the relative importance of each element in their associated vectors. $\mathbf{H}$ includes the linear constraints on velocity, longitudinal acceleration, jerk, and steering angle at each predicted time step. $\mathbf{H}$ also includes nonlinear constraints on lateral acceleration.
}

\subsection{Sequential Quadratic Programming}
We use Sequential Quadratic Programming (SQP) to solve a nonlinear optimization problem as in~\citet{MPCSQP}. SQP is an iterative approach where a sequence of quadratic programming problems \change{is} constructed and solved until \change{the} solution converges. To put \eqref{eq:GenForm} into quadratic form, we linearize the motion model and nonlinear constraints in $\mathbf{H}$ about the optimal solution $(\mathbf{x}_n^*,\mathbf{u}_n^*)$ found in the previous iteration. In the first iteration of SQP, we use the solution from the previous control time step. The resulting quadratic problems can be solved efficiently using an off-the-shelf solver.

\subsection{Timed Reference Generation}
Our NMPC formulation \eqref{eq:GenForm} is a trajectory tracking controller that requires a timed reference signal $\{\mathbf{r}_n\}_{n=k+1}^{k+N}$, which can be derived from the reference path and its associated velocity profile. We set $\mathbf{r}_{k+1}$ to be the closest waypoint on the desired path and simulate forward using a kinematic bicycle model to obtain subsequent desired states.
 


Our formulation provides a straightforward way to tune parameters, \change{but} it can still be tricky in practice. A bad choice of parameters can lead to jerky commands, poor tracking performance, or instability. A good starting point is to make sure that the structure of the cost function is well-understood.

\newpage

\section{Year 2 Competition Performance} \label{performancesection}

Zeus placed first in each dynamic challenge by a significant margin. U of T also placed first in all but two static event categories. This included social responsibility, concept design presentation, a simulation challenge, and a mapping challenge. The static non-driving events constituted 60\% of the 1000 total possible points for Year 2. The total points resulting from the competition are shown in Table~\ref{tab:scorecard}.

The first two days of the competition consisted of unpacking, safety inspections, and the installation of OXTS GNSS systems for scoring. The OXTS systems were mounted in each vehicle to monitor kinematic variables, lane boundary crossings, and whether the vehicle stopped appropriately. \change{These} data \change{were} analyzed by judges in order to assign points. In general, points were awarded for performing the expected \change{behavior,} such as attaining a speed limit or waiting for a pedestrian to cross the road. Points were subtracted for breaking Ann Arbor driving code or exceeding the prescribed kinematic envelope.

The third day of the competition included one hour of practice time \change{in} MCity. This was the only time that teams were given to tune algorithms. Zeus did not work perfectly during this practice run and several last-minute bug fixes were required. The most notable bug was an unforeseen error in LIDAR localization. This was addressed by switching to GPS/IMU localization and using the output of LIDAR localization to calibrate the offset between the Novatel GPS output and the desired position in the Carmera map frame. This issue is outlined in more detail in Section~\ref{subsecdiscuss}.

The last three days of the competition included the Traffic Control Sign Challenge, MCity and Pedestrian Challenge, and the Intersection \change{Challenge,} respectively. 

The remainder of this section provides details on Zeus' performance in each dynamic challenge. In doing so, we will describe some of the \change{noteworthy} errors our system experienced and the subsequent lessons learned. We will also discuss what did and did not go well, \change{last-minute} bug fixes, and our perspective on why we think we won.

\begin{table}[t]
\centering
\caption{Year 2 Competition Results}
\label{tab:scorecard}
\begin{tabular}{|l||c|}
\hline
\multicolumn{1}{|c|}{\textbf{University}} & \textbf{Total Points} \\ \hline \hline
\textbf{University of Toronto}            & \textbf{885}          \\ \hline
North Carolina A\&T State University      & 523                   \\ \hline
Texas A\&M University                     & 515                   \\ \hline
Michigan Technological University         & 471                   \\ \hline
Kettering University                      & 437                   \\ \hline
Virginia Tech                             & 430                   \\ \hline
Michigan State University                 & 352                   \\ \hline
University of Waterloo                    & 330                   \\ \hline
\end{tabular}
\end{table}

\subsection{Speed Zone Challenge}
The speed zone challenge required vehicles to drive straight along a section of highway and abide by the posted speed limits while staying within the lane lines. The speed limit positions were not encoded in the semantic map and as such had to be actively perceived and localized. At the competition, there were only two speed limit signs: a 20 mph \change{(9 m/s)} sign followed by a 15 mph \change{(7 m/s)} sign. \change{Note that our default driving speed at the start of the course was 5 m/s.} In order to receive full points, vehicles were required to reach the posted speed limit within 30 ft of the sign's position along the road.

\begin{figure}[t]
\centering
\includegraphics[width=0.6\columnwidth]{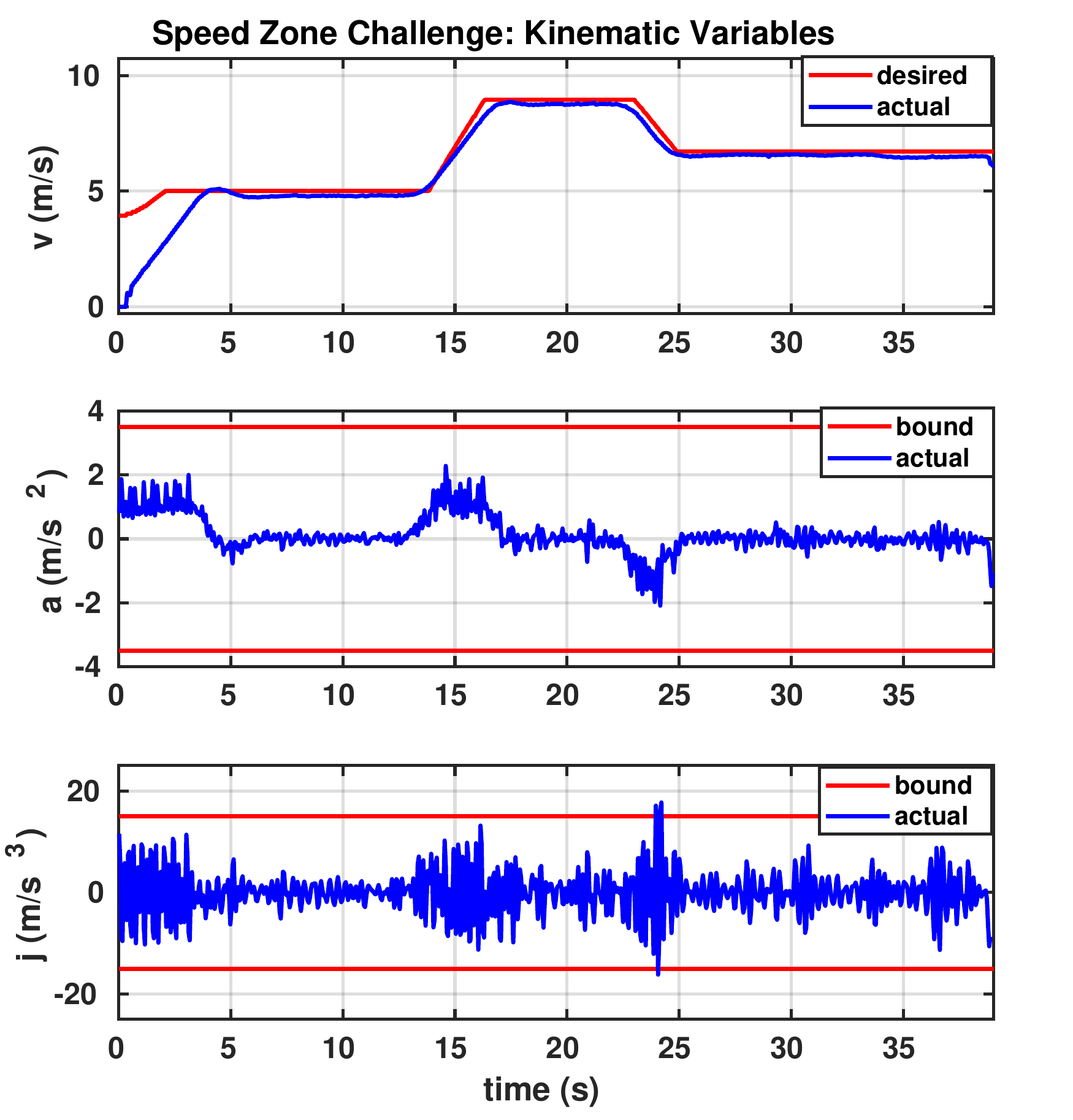}
\caption{This figure depicts Zeus' performance during the Speed Zone Challenge. The top figure shows the desired velocity alongside the actual velocity of the vehicle. The middle and bottom figures show that we stayed within our acceleration and jerk constraints for the entire course. This data \change{were} extracted from Zeus' own sensors during this challenge.}
\label{fig:speed}
\end{figure}

Figure~\ref{fig:speed} depicts Zeus' performance during the Speed Zone Challenge. At the start of the challenge, Zeus was not aligned properly in the lane and had to correct itself. Then, Zeus accelerated up to a default speed of 5 m/s. Zeus followed the center of the lane quite closely during the challenge but with a constant bias. We determined that this constant was due to both an error in our GPS/IMU offset calibration as well as a bug in the controller. Fortunately, the vehicle was still able to drive within 10 cm of the centerline for the entire challenge. Figure~\ref{fig:speed} shows that Zeus reached both speed limits correctly: 19.7 mph for the first speed limit and 14.8 mph at the second speed limit.
 
In order to detect speed limit signs sufficiently far in advance, we relied on an additional forward-facing camera with a \change{16-mm} lens and \change{30-degree} field of view. This camera effectively doubled the range of our sign detector by simply having a higher resolution further from the vehicle. At the competition, speed limit signs were detected over 80 m away, although we limited the detection range to 50 m due to the sparsity of LIDAR points beyond this distance.

Zeus received the maximum number of points for this challenge by staying within the prescribed acceleration and jerk envelope and attaining the posted limits within the required distance. Zeus stayed within the starting lane for the entire challenge.

\begin{figure}[ht]
    \centering
    \subfigure[Perspective View of Sign Detection]{
    \includegraphics[width=0.47\columnwidth]{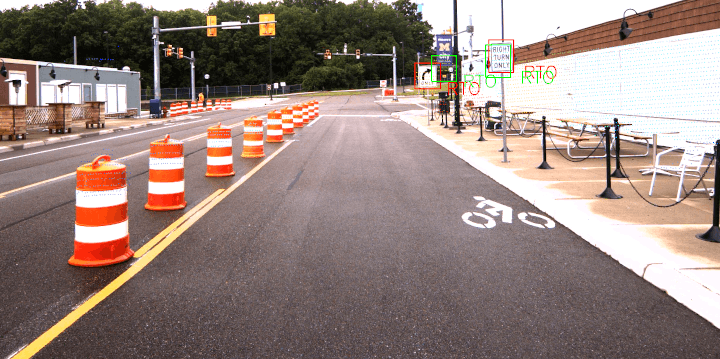}}
    \subfigure[Perspective View of Parking Spots]{
    \includegraphics[width=0.47\columnwidth]{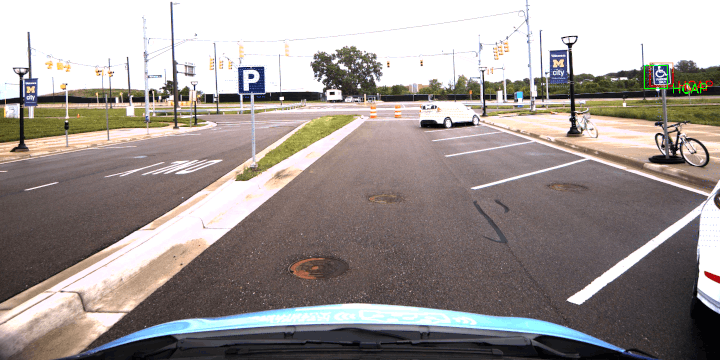}}
    \subfigure[Bird's Eye View of Sign Localization]{
    \includegraphics[width=0.47\columnwidth]{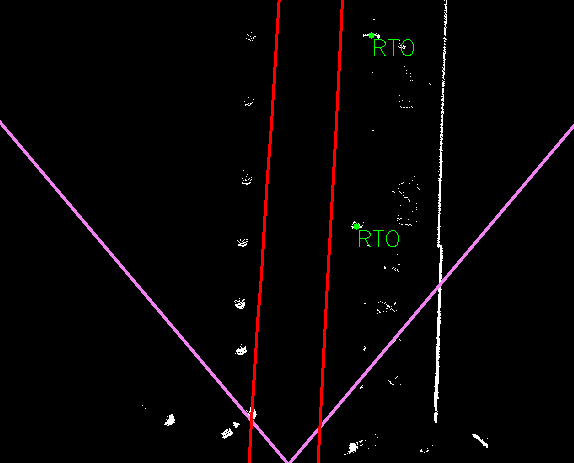}}
    \subfigure[Bird's Eye View of Parking Occupancy]{
    \includegraphics[width=0.47\columnwidth]{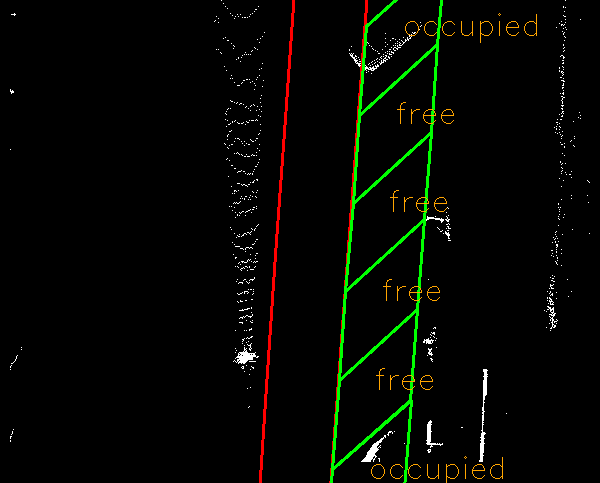}}
\caption{(a) Perspective view of signs detected in image. (c) Position of signs relative to the vehicle. (b) A handicap parking sign was detected and associated with a parking spot. (d) The bird's eye visualization of the parking occupancy node.}
    \label{fig:traffic}
\end{figure}

\vspace{-1mm}

\subsection{Traffic Control Sign Challenge}

\vspace{-1mm}

The Traffic Control Sign Challenge required the vehicle to continue straight unless directed otherwise by signs present on the road. At the competition, this challenge started with two semantically equivalent but visually different Right-Turn-Only (RTO) signs. Zeus correctly detected and localized these signs from the start line. After seeing these signs, our global planner triggered a re-plan to turn right at the upcoming intersection and then continue driving straight. Figure~\ref{fig:traffic} ~(a,c) depicts the sign detection at the start of the challenge from both the perspective view and bird's eye view. Figure~\ref{fig:traffic}~(a) depicts the raw detections in red and the tracked detections in green. For the second sign, the \change{pointcloud-to-detection} association is a little \change{off,} causing it to appear misaligned in the perspective image. The bottom image depicts the localization of both signs relative to the vehicle. The red lines correspond to the ego-lane and the purple lines are the field of view of the main camera.

\vspace{-1mm}

Due to some careful analysis of the rules and the layout of MCity, we determined that if Zeus drove through the parking section during this challenge, it should automatically start looking for a valid parking spot. Figure~\ref{fig:traffic}~(b) depicts the sign detection and parking occupancy once Zeus rounded the first corner. Two of the parking spots were occupied with dummy cars. A third parking spot was blocked by the presence of a handicap parking sign. 

\vspace{-1mm}

Upon turning the corner, our occupancy detection node began counting the number of points in each spot. Due to our temporal smoothing setup, each spot starts in an \textit{unknown} state and requires several \textit{free} detections before being marked as unoccupied. Because of this, Zeus almost did not detect a free parking spot in time. Figure~\ref{fig:traffic}~(d) depicts the output of the occupancy node after enough detections had been received to correctly determine which spots were occupied. At this point, a specialized planner generated a trajectory that would place Zeus in the center of a spot.

\vspace{-1mm}

During our one-hour practice, the parking maneuver did not initiate due to a bug in our planner. Hence, this maneuver had not yet been tested at MCity before this challenge. Several tuning parameters that had been validated at the University of Toronto were modified slightly given the shape of the spots at MCity and the results of simulation tests. \change{These simulation tests were conducted using our own internal simulation tools (using C++ and ROS) which simply create a model of the vehicle for the controller to interact with and allow the vehicle to move around within the semantic map. This simple simulator allows us to test planning logic and validate that the paths output by the planner are smooth and correct. Our} desired velocity was lowered to 1.5 m/s during the parking maneuver to ensure that our controller would perform with high accuracy. Given these last-minute changes and a little luck, we were able to park almost perfectly during the competition run. Zeus received the maximum possible points for this challenge. 

\vspace{-1mm}

\begin{figure}[t]
\centering
\includegraphics[width=0.6\columnwidth]{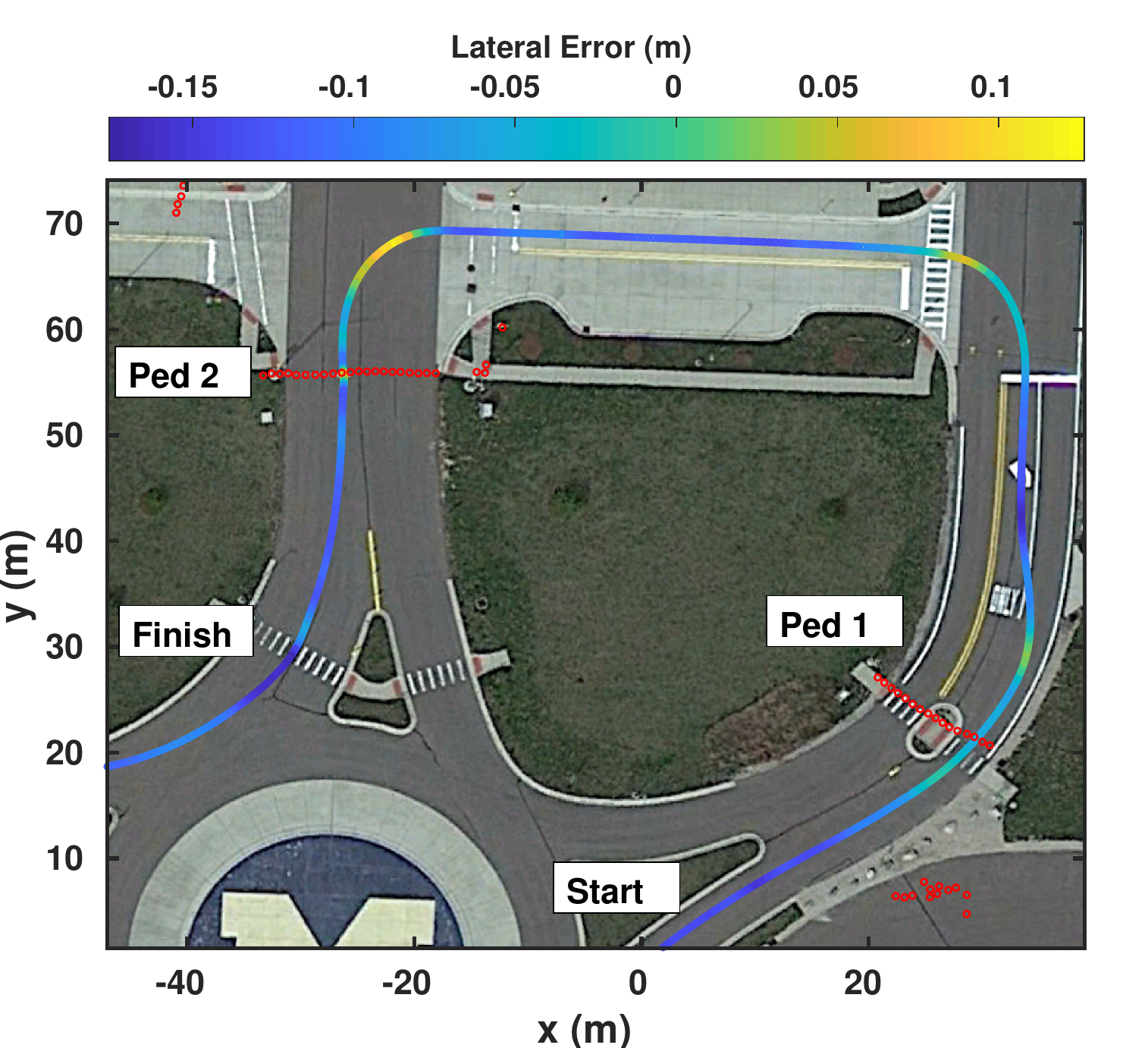}
\caption{This figure depicts Zeus' trajectory during Pedestrian Challenge course 1. The color of the trajectory corresponds to lateral error. The red dots correspond to detected pedestrian locations. Note that the satellite image is meant to provide context only as it does not align perfectly with the map. \change{Also note that the error does not have a mean of zero. This was due to a bug in our controller at MCity.}}
\label{fig:ped1}
\end{figure}

\vspace{-1mm}

\subsection{Pedestrian Challenge}

\vspace{-1mm}

The Pedestrian Challenge required vehicles to sequence several intersections while abiding by traffic lights and reacting appropriately to pedestrians attempting to cross the road. This challenge was divided into two separate courses. Figure~\ref{fig:ped1} depicts the path Zeus took during the first course. The path is colored by the lateral error at each point along the path. Note that Zeus stays quite close to the centerline. Also shown in Figure~\ref{fig:ped1} \change{are} the detected pedestrian positions in red. This minimal number of detections outside the crosswalk regions shows that our pedestrian detection has a very low false positive rate. On the bottom-right and top-left of Figure~\ref{fig:ped1}, extraneous detections correspond to by-standers and test crew. There were only a handful of false positives in the two courses.

\vspace{-1mm}

The first pedestrian encountered was a child-sized dummy. Although we did not explicitly train on this target, it was visually quite similar to the adult-sized dummy that we had trained on. Hence, our 2D DNN was still able to detect it. \change{However,} due to its size and position on the road, the child-sized dummy was not detected until it was within 15 m of the vehicle. In order to give Zeus ample time to detect and react to these pedestrians, our desired speed for this challenge was lowered to 3 m/s.

\vspace{-1mm}

\begin{figure}[t]
    \centering
    \subfigure[Perspective View]{
    \includegraphics[width=0.45\columnwidth]{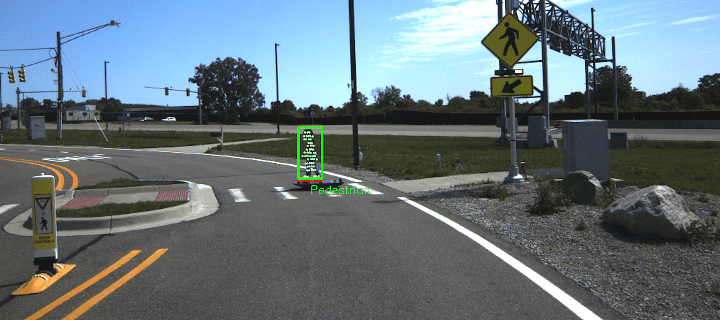}}
    \subfigure[Bird's Eye View]{
    \includegraphics[width=0.45\columnwidth]{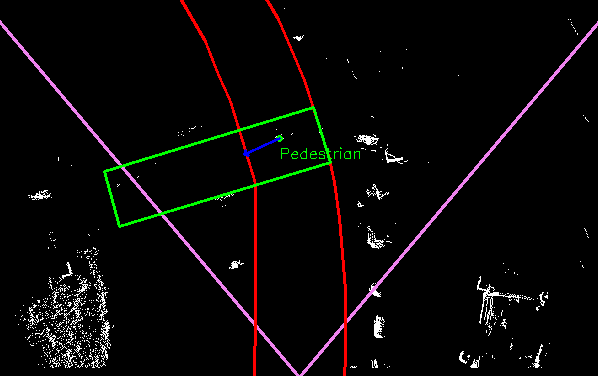}}
\caption{This figure depicts a pedestrian detection from both the perspective and bird's eye view.}
    \label{fig:ped2}
\end{figure}

Zeus' pedestrian detection and tracking worked largely as expected during this challenge. Figure~\ref{fig:ped2} depicts the child-sized dummy being tracked in the perspective and bird's eye view. In the perspective view, the tracked bounding box is green and the green dots are the LIDAR points associated with the pedestrian. In the bird's eye view, the position of the pedestrian relative to the vehicle is shown as a green dot. The red lines are the ego-lane, the crosswalk is highlighted in green, and the field of view of the main camera is shown in purple. The blue line attached to the pedestrian indicates its velocity vector.

\vspace{-1mm}

The child-sized dummy simply walked from one side of the road to the other in front of the vehicle. The second pedestrian involved a more complex scenario. In this case, the pedestrian was set up to walk towards Zeus on the left-hand crosswalk while Zeus was expected to turn left at the intersection. In this case, more than one crosswalk needed to be scanned and a greater detection range was required. The second pedestrian was initially detected over 40 m away.

\vspace{-1mm}

As mentioned in section~\ref{lnssection}, our traffic light detector had undefined behavior for traffic lights in the 'off' state. In order to handle flashing red lights, we simply treated them as solid red lights with a short \change{five-second} timeout. This worked quite well during the pedestrian challenge but caused us to perform a rolling stop at the first intersection and lose five points accordingly.

\vspace{-1mm}

After passing the second pedestrian, a pedestrian detection was reacquired by a corner camera halfway through the intersection. Due to a bug in our planner, this caused us to stop unexpectedly before the next crosswalk. This error caused us to violate our jerk constraints and lose a point.

\vspace{-1mm}

The second course used the same intersections and pedestrian locations but in the reverse order. Zeus tracked both pedestrians without error and experienced only a single false positive. Zeus received the maximum possible points for the second course.

\vspace{-1mm}

\begin{figure}[h]
    \centering
    \subfigure[Traffic Light Detection Error]{
    \includegraphics[width=0.45\columnwidth]{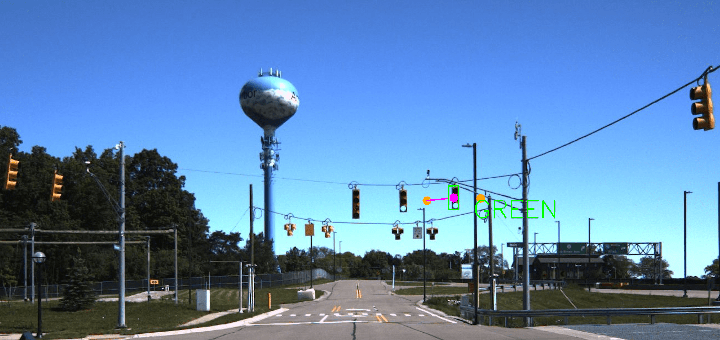}}
    \subfigure[Before Tunnel - Traffic Light Detection]{
    \includegraphics[width=0.45\columnwidth]{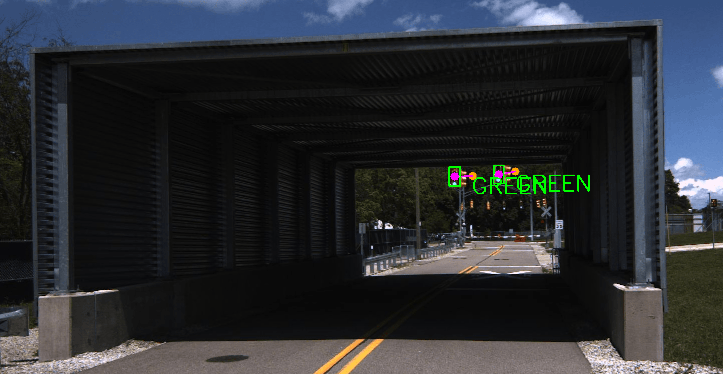}}
    \subfigure[Associated Raw Detections]{
    \includegraphics[width=0.45\columnwidth]{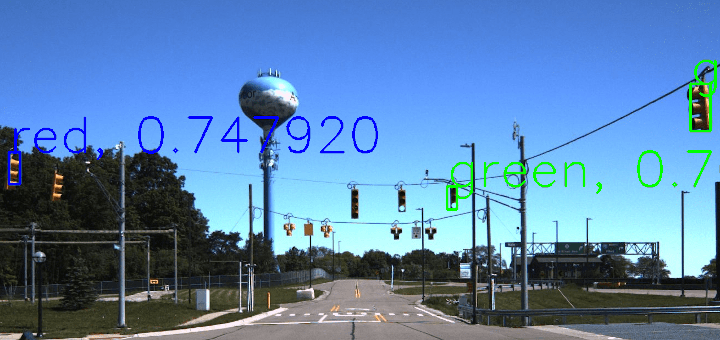}}
    \subfigure[Inside Tunnel - Traffic Light Detection]{
    \includegraphics[width=0.45\columnwidth]{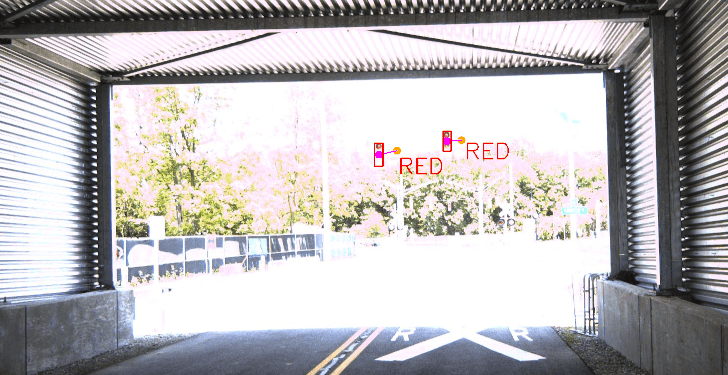}}
\caption{Left: Zeus was stuck at this intersection for two minutes. (a) There are three traffic lights the upcoming intersection. Only one has been properly detected but it has been associated with the wrong expected light position. (c) This image shows the root of the problem: the non-maximum suppression threshold has been set too high. This high threshold, coupled with the large number of traffic lights in the frame, caused some potential detections to be discarded. Right: Traffic Light Detection experiences an error when entering a tunnel. \change{(b) Before entering the tunnel, the light is correctly classified as green. (d) The area outside the tunnel becomes overexposed and causes the traffic light to be incorrectly classified as red.}}
\label{fig:lightbad}
\end{figure}

\vspace{-1mm}

\subsection{Intersection Challenge}

\vspace{-1mm}

During the Intersection Challenge, vehicles were required to sequence 13 intersections while abiding by traffic lights. Zeus' traffic light detection performed largely as expected during this challenge. We believe the success of this system is due to the massive custom dataset that was collected as well as the large amount of testing leading up to the competition. Despite this success, there was an interesting failure case that is worth discussion.

\vspace{-1mm}

At the second intersection, Zeus became stuck for two minutes before proceeding. The two factors that contributed to this error were the non-maximum suppression (NMS) threshold for SqueezeDet and an offset in our GPS-based positioning. Figure~\ref{fig:lightbad} depicts the traffic light detections at this intersection once the lights turned green. By setting the NMS threshold too high, detections that would have otherwise been returned were suppressed. Further, it can be seen in Figure~\ref{fig:lightbad}(a) that the there is a significant offset between the expected traffic light position (orange dot) and the detected position (green box). Both of these factors contributed to only one of the traffic lights being detected as green.

\vspace{-1mm}

Ordinarily, we require all traffic lights at an upcoming intersection to be detected as green before proceeding.
Figure~\ref{fig:lightbad}(a) shows that only one of the traffic lights has been detected and classified as green. Because of this, the other traffic light is assumed to be \change{red,} which prevents the vehicle from proceeding. Normally, a \change{60-second} timeout would prevent Zeus from becoming stuck at an intersection. In this case, SqueezeDet occasionally output the correct \change{detections,} allowing Zeus to start moving forward. However, the pitching of the vehicle caused the image to change sufficiently for our DNN to incorrectly classify the lights again and cause Zeus to hit the brakes. This process repeated until the vehicle crossed the stop-line, after which traffic light detections were ignored and the vehicle proceeded through the intersection.

\vspace{-1mm}

The lesson here is that although increasing an NMS threshold can reduce noise, it leads to more false negatives. For several perception tasks in self-driving such as object detection, false negatives can be more detrimental than false positives. As such, a good approach should bias towards more false positives and filter these out using tracking and semantic information. Another lesson is that requiring all traffic lights to be detected as green may be too cautious for real-world driving. There are many scenarios in which several redundant traffic lights are present. In these cases, it is sufficient to detect a subset of the lights as green.

\vspace{-1mm}

Zeus lost two points for exceeding jerk constraints and another eight points for not stopping behind the stop-line at four different intersections. In one of these cases, Zeus stopped after the stop-line due to an error in our semantic map. Zeus received the most points during this challenge by a significant margin.

\vspace{-1mm}

\begin{figure}[t]
    \centering
    \subfigure[Railroad Crossing]{
    \includegraphics[width=0.45\columnwidth]{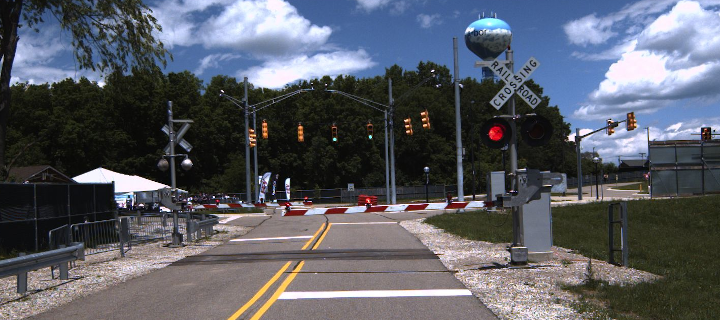}}
    \subfigure[Cyclist]{
    \includegraphics[width=0.45\columnwidth]{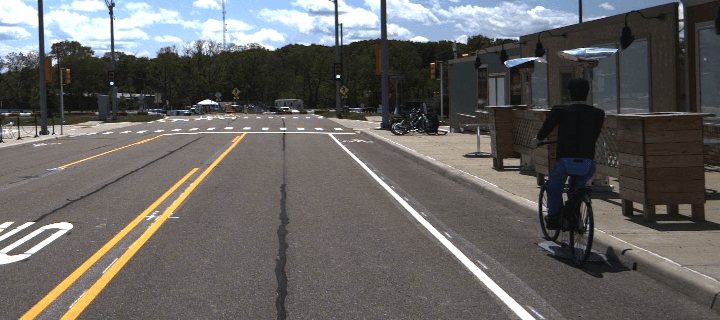}}
    \subfigure[Deer Crash]{
    \includegraphics[width=0.45\columnwidth]{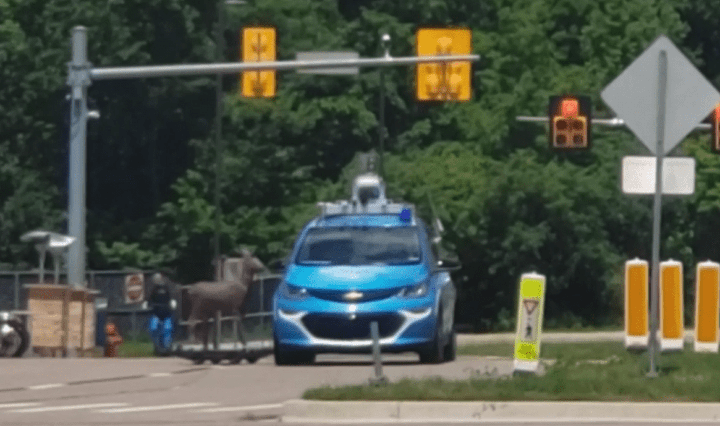}}
    \caption{MCity Challenge Obstacles included a tunnel, a railroad crossing, a static cyclist, and a dynamic deer. (c) depicts the moment right before the dummy deer collided with the side of Zeus.}
    \label{fig:mcitystuff}
\end{figure}

\subsection{MCity Challenge}

The MCity Challenge required vehicles to sequence several intersections while handling various obstacles along the way. These obstacles included a tunnel, railroad crossing, static cyclist, and dynamic animal. Figure~\ref{fig:mcitystuff} depicts the railroad crossing and static cyclist.

Zeus drove through the \change{10 m long and 6 m wide} tunnel mostly without issues. The Novatel GPS/IMU positioning did not jump in the tunnel as originally expected and Zeus stayed close to the centerline throughout. However, the traffic light detector did experience a noteworthy error. This error is depicted in Figure~\ref{fig:lightbad}~(b,d). Before entering the tunnel, we detected the oncoming traffic lights as green. \change{However,} once we entered the tunnel, the auto-exposure of the camera caused the region outside the tunnel to become over-exposed. This caused the traffic light detector to erroneously report the traffic lights as red. This in turn caused Zeus to slow down unnecessarily in the tunnel.

The lesson here is that handling changing illumination conditions while going through tunnels is a challenging problem. One potential solution to this problem could include having cameras with better dynamic range or using multiple cameras with different exposure settings. Another solution could take advantage of semantic information to schedule camera exposure values to handle tunnels better.

The second obstacle encountered was a railroad crossing. MCity has a fully-functional railroad crossing with moving arms and flashing lights. aUToronto ran out of time before the competition to develop perception software to handle railroads. It was not possible to test railroad handling during the 1-hour practice period. Thus, we did not know whether our secondary obstacle detection would be sufficient to detect the railroad arm.
As such, we made a last-minute decision to circumvent railroad detection completely by relying on a simple timed stop instead. This strategy allowed us to stop at the railroad but caused us to wait much longer than necessary after the arms lifted. None of the other AutoDrive teams made it past the railroad during the MCity Challenge.

The third obstacle was a static cyclist in a bike lane. Since there are only two bike lanes at MCity, we simply shifted our centerline over to the left for those regions. This centerline shift made it so that at least 1 m of clearance would be provided to a cyclist in the bike \change{lane,} thus satisfying competition requirements.

The fourth and final obstacle was a dynamic deer. The deer was positioned on the grass at the side of the road. The deer started moving when it was within 10 m of Zeus. When the deer entered the road, Zeus was travelling 4 m/s and the deer was 6 m away laterally and 5 m away longitudinally. The deer then collided with the side of \change{Zeus,} forcing the safety driver to perform a manual takeover. Figure~\ref{fig:mcitystuff}~(c) depicts the moment immediately before the collision.

Inspection of recorded data showed that our secondary obstacle detector did not detect the deer at all. There are several reasons for this. Early on in development, we made the incorrect assumption that the dynamic obstacles we would be required to handle would be in front of Zeus and in the ego-lane. Because of this assumption, we ignored LIDAR points outside the ego-lane or points behind the bumper longitudinally. The deer's trajectory was such that it was outside the ego-lane until the last moment, where it was already behind the front bumper longitudinally.

There are several lessons that can be drawn from this collision. The first is that our secondary obstacle detection had too many assumptions built into it. In the future, it will be important to track objects over a much larger region around the vehicle even if the competition does not explicitly require it. The second lesson revolves around prediction. Even if we had detected the deer, the collision would likely still have occurred. This is because our perception system currently does not predict the future motion of objects. Thus, a key component that we will develop in the future will enable Zeus to extrapolate the future positions of objects in order to predict potential collisions and react accordingly.

\subsection{Discussion} \label{subsecdiscuss}

Zeus completed three out of the four challenges without requiring a manual takeover. Even though the perception components such as object detection and traffic light detection experienced some faults, they operated as expected during the competition. False positives in object detection were filtered out using the semantic information of lane and crosswalk locations. Errors in traffic light classification were overcome using timeouts, ensuring Zeus would complete each challenge to maximize points.

A significant portion of the dynamic challenge points \change{was allocated} to remaining within \change{lane} lines, stopping accurately, and staying within a kinematic envelope. Our MPC controller kept us close to the reference trajectory and within these kinematic constraints for most of the competition. Even driving at higher speeds (40 km/h), the vehicle drove smoothly with no sudden movements.

In the previous sections, we identified some of the faults encountered by Zeus during each challenge. In the Traffic Control Sign Challenge, our parking occupancy software took too long to determine the occupancy of each \change{spot,} which almost prevented a parking maneuver. During the Pedestrian Challenge, Zeus committed a rolling stop and stopped unexpectedly during the first course. During the Intersection Challenge, Zeus became \change{temporarily} stuck at the second intersection due to a localization offset and a non-maximum suppression threshold. In the MCity Challenge, Zeus slowed down in the tunnel, relied on a hard-coded timeout to pass the railroad, and was hit by the dynamic deer.

By highlighting these faults in our system, our goal was to outline some of the weaknesses of the approaches that we employed. Although Zeus performed the best out of all the AutoDrive Challenge teams, these faults demonstrate that there is still a significant amount of work left in order to bring Zeus to Level 4 autonomy.

It is also important to note that Zeus required several last-minute bug fixes in order to perform well during the competition runs. During our 1-hour practice time, several bugs in the planner and LIDAR localization were uncovered. The planner bugs included issues with runtime performance, red light timeouts, railroad crossings, and pedestrian handling. Most notably, Zeus did not autonomously park during our 1-hour \change{practice,} which prevented us from tuning parameters before the competition run. Possibly the most significant bug uncovered during this 1-hour practice was that LIDAR localization did not work as expected. In some regions of MCity, the localization output experienced jumps.

As mentioned in section~\ref{locsection}, we had initially planned to use Applanix's LIDAR localization at the competition. Since access to MCity prior to the competition was prohibited, we acquired a LIDAR map from Carmera. This LIDAR map was in the form of a large aligned pointcloud for all of MCity. This LIDAR map needed to be converted to Applanix's map format in order to run their localization software. 

Our assessment is that our process of converting Carmera's map into Applanix's format resulted in the errors that were observed during the practice run. Ideally, LIDAR maps are custom-built using Applanix's own software. This is what we do at the University of Toronto and the resulting position estimates are exceptionally reliable. Unfortunately, there was not adequate time during the one hour of practice to build a LIDAR map and test autonomous functions.

Due to this issue with LIDAR localization, we were forced to use Novatel's GPS/IMU positioning instead. Our primary concern was that the Novatel positioning was biased with respect to \change{the} semantic map provided by Carmera. In order to calibrate for this bias, we compared our GPS/IMU positioning against LIDAR localization during the 1-hour practice run. It is important to note that LIDAR localization returns a relative measurement of one's position in a LIDAR \change{map,} whereas GPS/IMU positioning returns a global measurement of one's position on the Earth. For this reason, it was possible to treat LIDAR localization as a 'ground truth' to calculate the bias in our GPS/IMU positioning. The bias we calculated was quite large: 70 cm in Easting, 90 cm in Northing, and 120 cm in altitude. Without this bias correction, we may not have been able to complete any of the challenges.

As a team, aUToronto benefited from being well-organized and performing frequent testing on Zeus. Frequent testing was enabled by having access to private roads at U of T's Institute for Aerospace Studies (UTIAS) where Zeus' garage is located. The timeline leading up to the competition was broken up into milestones which each culminated in a series of real-world tests. Two of these milestones tests were held at new locations: first at U of T's Mississauga campus and second at the Clearpath Robotics office in Waterloo. By testing Zeus in new locations, we were forced to encounter the shortcomings in our system that might otherwise have gone unnoticed. Lastly, for the six weeks leading up to the competition, a small subset of aUToronto was dedicated to testing Zeus and fixing bugs on a daily basis. We believe that this testing prior to competition was one of the major elements that set Zeus apart in terms of reliability.

There were also several design choices that may have given aUToronto an edge. First, the decision was made early on to avoid active lane detection and to simply focus on driving using a semantic map.  We credit some of the dynamic challenge points to the performance of our MPC controller. Due to the competition restriction to use CPUs and FPGAs, significant effort was invested into designing perception components that would run in \change{realtime} on CPUs. By keeping this restriction in mind, aUToronto was able to develop perception components with low latency. Using multiple cameras was also a critical component to completing several challenges. Our \change{narrow-field-of-view} camera boosted visual detection range, and the \change{45-degree} cameras were essential to tracking a pedestrian from one side of the crosswalk to the other. Finally, we believe that training our DNNs on our own custom dataset allowed us to achieve high precision and recall numbers that we might not have otherwise.

\change{How could this system have been made more general? The reader will note that the system described in this work is very optimized for the AutoDrive competition. Here, we note several aspects of our design that make it generalizable to more complex environments. First, the entire software stack is agnostic to the actual test location. We simply create or acquire a semantic map for a new test location and Zeus is able to drive autonomously. However, the semantic map must be accurate. We believe that our traffic light detection system is generalizable to more complex driving. With a more powerful DNN than SqueezeDet, and a larger training set, it should be possible to detect and classify a greater variety of traffic light states with higher reliability. Further, our method of projecting the expected traffic light positions onto the image should be generalizable to any new intersection. Our object detection and tracking pipeline, aUToTrack, works well for pedestrians but not as well for cars. This does not necessarily mean that it needs to be replaced entirely for Level 4 autonomy, but rather that there should be several object detectors working in parallel to detect cars, static obstacles, and other traffic participants. In addition, our object tracking pipeline can easily be extended to track a large number of objects and types.

Here we summarize some of the major improvements that should be made before taking Zeus onto public roads. In Section~\ref{objectsection}, we mentioned that our object detection and tracking pipeline is not immediately extensible to vehicles. This shortcoming is primarily due to our method of clustering LIDAR points to obtain object centroids. In order to detect vehicles and retrieve an accurate 3D bounding box, using a modern 3D object detector is required. These types of detectors work directly on LIDAR data or on a fusion or LIDAR and vision. Examples of these types of detectors include PointPillars \citep{pointpillars} and AVOD \citep{avod}. Additional components that should be developed inclue a dynamic cluster detection node such as the method described in \citep{yoon2019mapless} and an occupancy-grid-style static obstacle detection system to support the primary dynamic object detection. 

We believe that our planner needs to be revamped while still maintaining a hierarchical structure. Currently, we stitch together the centerlines of road segments to form a path and perform minor adjustments for smoothness. This approach works well for the constrained environments of the AutoDrive competition but is unlikely to work reliably in real-world driving. For example, nudging around static obstacles that protrude onto the road would be very difficult to achieve with our current planner. Two candidate solutions to replace our current approach include a sampling-based approach as described in \citep{werling2010optimal} and a lattice-grid-based approach as described in \citep{pivtoraiko2009differentially}.
}

\change{How did the competition environment compare to real-world driving conditions? MCity is a small mock-town built for self-driving testing at the University of Michigan. The test site is a great analog for a small town urban environment. We did not get the impression that this site was designed to facilitate self-driving testing, but rather to be as realistic of a driving analog as possible. Nevertheless, the way in which MCity was used in the competition was far from a realistic driving test. Encounters with pedestrians were limited to four crosswalk situations with familiar pedestrian dummies on remote-control platforms. There were no pedestrians outside of the expected crossing locations and no jaywalkers, both of which are common occurrences in real driving. We believe that the traffic light testing performed at MCity was a realistic test, but simply lacked enough variety to be confident that our system would work on public roads. The most glaring omission was the lack of dynamic vehicles and other traffic participants which a self-driving car must be capable of handling on public roads. Furthermore, the tests were conducted in sunny or overcast conditions so inclement weather and night-time driving were not a factor.

How do these lessons learned translate to real-world driving? First, having a clear understanding of the strengths and weaknesses of a chosen DNN architecture is critical. To reiterate, we chose to use SqueezeDet because it was the best approach that fit within our computational constraint of using CPUs and FPGAs. There are many alternatives to this architecture which can run in real-time on a moderately powerful GPU while achieving significantly greater performance. Second, the quantity and quality of the training data can have a large impact on DNN performance. Great care must be taken to collect a sufficient quantity of data and for it to be representative of the test distribution while being mindful of class imbalances and the long tail of infrequent objects. Creating a dataset, creating a training pipeline, and managing a hyperparameter search is a significant engineering effort.  Most of the errors Zeus encountered at the Year 2 competition were the result of insufficient testing or simple software bugs. Clearly, following software engineering best practices and performing exhaustive simulation testing is critical to minimizing these preventable bugs. 
}

\section{Conclusions and Future Work}

\begin{figure}[t]
\centering
\includegraphics[width=0.6\columnwidth]{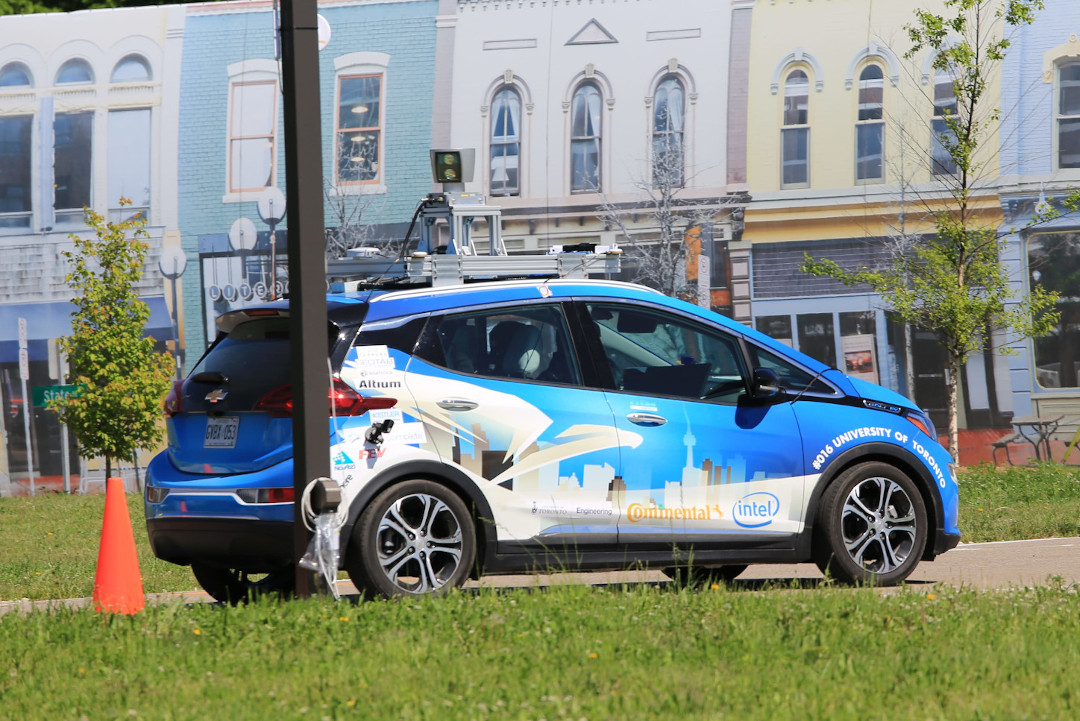}
\caption{This figure depicts Zeus at the start of the Intersection Challenge, in front of the movable building facades at MCity.}
\label{fig:zeusconc}
\end{figure}

This article provided a system description of Zeus, aUToronto's winning entry in the SAE AutoDrive Challenge. \change{A final picture of Zeus at the start of the Intersection Challenge is shown in Figure~\ref{fig:zeusconc}.} We described the team's organizational structure and the development timeline. In the system overview, we described the layout of sensors on Zeus and the resulting fields of view. We also described Zeus' software architecture \change{at} a \change{high level}. In the subsequent sections, we described the design of each major component of Zeus' software stack. In the final section, we described Zeus' performance on each dynamic challenge.

In order to achieve high performance from our DNNs for object detection, we collected our own dataset using a mixture of private and public road driving in Toronto. Our object detection and tracking pipeline, dubbed aUToTrack, relied on a 2D DNN and LIDAR clustering to identify pedestrians. We employed a Model Predictive Controller which enabled high tracking performance at the competition. For the majority of each challenge, we stayed within 15 cm of the desired centerline.

Year 3 of the AutoDrive Challenge will be held at the Ohio Transportation Research Center in October 2020. The purpose of the challenge is to simulate an autonomous ride-sharing vehicle. The challenge will include dynamic pedestrians, traffic lights and traffic signs. The most notable additions are the introduction of road closures in the form of construction zones. The routes are intended to be much longer and varied.

Our future work will include adding a prediction component to object detection and tracking. In order to properly handle jaywalkers \change{and deer}, it will be critical to predict the future positions of traffic participants. \change{Further, we are developing a LIDAR-based 3D object detector to detect vehicles in future years of the competition.} We will research using the output of the perception nodes as a localization correction. For the purpose of safety, we will investigate adding sensors to cover the blind spots of the HDL-64. Our planning software will be updated to dynamically plan paths around obstacles. We will investigate using a dynamic vehicle model with our model predictive controller. We will implement a dynamic cluster detector and an occupancy grid generator. We will be reopening our DNN architecture search in an attempt to find a better architecture that still runs in realtime.

\section{Acknowledgements}
This project would not have been possible without the generous contributions of our sponsors: SAE, General Motors, Intel, Continental, Velodyne, Novatel, Here maps, Mathworks, Applanix, Fleet Complete, Geotab, Dessa, RightHook, and Carmera.

Since the team's inception in the Spring of 2017, many individuals have contributed to aUToronto and to developing Zeus. 

Members that contributed heavily to the design and development of the Year 1 system include Zachary Kroeze, Andreas Schimpe, Mona Gridseth, Chengzhi Winston Liu, Qiyang Li, Venugopalan Krishnaswamy, Tianhao Hu, Stewart Jamieson, Kevin Jen, and Chris Lucasius.

Members that contributed to the Year 2 design but were not listed as an author include: Robert Adragna, Adam El-Masri, Zane Huang, Alvin Huang, Jacob Nazarenko, Chelsea Rosic, Jerry Li, Siyun Li, Cindy Ding, Jenny Bao, Ziyad Edher, Zheng Yao, Wenda Zhao, Ted Huang, Brian Liu, Richard Hu, Haowei Zhang, Lilly Yu, Kevin Hua, and QiYu Li.

The mechanical and electrical components of Zeus were largely the work of Anston Emmanuel, Liam Horrigan, Tharindu Silva, Mithun Jothiravi, Omar Rasheed, and Pranshu Malik.

Also, a special thanks is given to UofT Engineering for funding UofT's entry in the AutoDrive Challenge.

\bibliography{references}
\bibliographystyle{apalike}

\end{document}